\documentclass{article}

\usepackage[preprint,nonatbib]{neurips_2022}
\usepackage{etoolbox}
\usepackage{kvoptions}
\SetupKeyvalOptions{family=writeup, prefix=writeup@}
\usepackage{xcolor}
\usepackage{mathptmx}
\usepackage{caption}

\usepackage{xspace}

\usepackage[T1]{fontenc}
\usepackage[utf8]{inputenc}
\usepackage{wrapfig}

\usepackage[numbered]{bookmark}

\makeatletter \renewcommand\Hy@numberline[1]{#1. } \makeatother

\newcommand{\mynewbool}[2]{\newbool{#1}\setbool{#1}{#2}}
\newcommand{\myif}[2]{\ifbool{#1}{#2}{}}

\title{Robust Procrustes}
\author{
    Tal Amir
    \\ Faculty of Mathematics
    \\ Technion --- Israel Institue of Technology
    \\ \texttt{talamir@campus.technion.ac.il}
    \And Shahar Kovalsky
    \\ Department of Mathematics
    \\ University of North Carolina at Chapel Hill
    \\ \texttt{shaharko@unc.edu}
    \And Nadav Dym
    \\ Faculty of Mathematics
    \\ Technion --- Institute of Technology
    \\ \texttt{nadavdym@technion.ac.il}
    }
\date{17 May 2022}

\mynewbool{useTables}{true}
\mynewbool{useGraphics}{true}
\mynewbool{useTikz}{false}
\mynewbool{useAlgorithms}{true}

\mynewbool{usePublicationList}{false}
\usepackage{tikz}
\usepackage{pgfplots}
\usepackage{pgfplotstable}

\usetikzlibrary{plotmarks}
\usetikzlibrary{matrix}
\usepgfplotslibrary{units}

\pgfplotsset{compat=newest,table/search path={conent/figures}} 

\tikzset{every mark/.append style={solid}}

\usepackage{packages/ta-lists}
\usepackage{packages/ta-tables}
\usepackage{packages/ta-algorithms}
\usepackage{packages/color_names}
\usepackage[thmbysec]{packages/math_environments}
\usepackage{packages/cross_references}
\usepackage{packages/math_notation}
\usepackage{import}

\def\@testdef #1#2#3{%
  \def\reserved@a{#3}\expandafter \ifx \csname #1@#2\endcsname
 \reserved@a  \else
\typeout{^^Jlabel #2 changed:^^J%
\meaning\reserved@a^^J%
\expandafter\meaning\csname #1@#2\endcsname^^J}%
\@tempswatrue \fi}

\usepackage[disable,textsize=scriptsize]{todonotes}

\definecolor{col_red}{RGB}{255,0,0}
\definecolor{col_blue}{RGB}{0,0,180}

\newcounter{todocounter}

\newcounter{comment}
\newcommand{\comment}[2][]
{
    \refstepcounter{comment}
    \todo[color={red!100!green!33}]
    {
        \textbf{\thecomment~[{#1}]:}
        {#2}
        }
    }

\newcommand{\Econv}{\tilde{\textup{E}}}
\newcommand{\Ecov}{\textup{E}_{\mathrm{cov}}}

\NewDocumentCommand{\p}{O{i}O{}}{{p_{#2}^{\br{#1}}}}
\NewDocumentCommand{\q}{O{i}O{}}{{q_{#2}^{\br{#1}}}}
\NewDocumentCommand{\pt}{O{i}O{}}{{\tilde{p}_{#2}^{\br{#1}}}}
\NewDocumentCommand{\qt}{O{i}O{}}{{\tilde{q}_{#2}^{\br{#1}}}}
\NewDocumentCommand{\uu}{O{i}O{}}{{u_{#2}^{\br{#1}}}}
\NewDocumentCommand{\vv}{O{i}O{}}{{v_{#2}^{\br{#1}}}}
\NewDocumentCommand{\x}{O{i}O{}}{{x_{#2}^{\br{#1}}}}
\NewDocumentCommand{\y}{O{i}O{}}{{y_{#2}^{\br{#1}}}}

\newcommand{\I}{{I}} 

\def\Unif{\textup{Unif}}
\def\nunmapped{\tilde{n}}
\def\Punmapped{\tilde{P}}
\def\Qunmapped{\tilde{Q}}
\def\cov{\textup{cov}}

\definecolor{col_method_name}{RGB}{26, 12, 130}
\newcommand{\methodref}[1]{\texttt{\textcolor{col_method_name}{#1}}\xspace}

\def\SRP{\textup{SRP}\xspace}

\newcommand{\SRPtwo}{{$\textup{SRP}_2$}\xspace}
\newcommand{\SRPinf}{{$\textup{SRP}_{\infty}$}\xspace}
\newcommand{\SRPp}[1][p]{{$\textup{SRP}_{#1}$}\xspace}

\newcommand{\Eproc}{\E_{\textup{Proc}}}
\newcommand{\E}{\textup{E}}

\newcommand{\Ep}[1][p]{\textup{E}_{#1}}
\newcommand{\Epalt}[1][p]{\tilde{\textup{E}}_{#1}}

\newcommand{\Eopt}{\textup{E}^{*}}
\newcommand{\Epopt}{\textup{E}_p^{*}}

\newcommand{\Fp}[1][p]{\textup{F}_{#1}}
\newcommand{\Fpalt}[1][p]{\tilde{\textup{F}}_{#1}}
\newcommand{\G}{\textup{G}}

\DeclareDocumentCommand \Aopt{ o } { \IfNoValueTF {#1} {{A^{*}}} {{a^{*}_{#1}}}	}

\newcommand{\Rhat}{\hat{R}}
\newcommand{\Aproj}{\Pi\br{A}}
\newcommand{\Aoptproj}{\Pi\br{\Aopt}}
\newcommand{\orthproj}[1]{{\Pi\of{#1}}}

\newcommand{\that}{\hat{t}}
\DeclareDocumentCommand \topt{ o } { \IfNoValueTF {#1} {{t^{*}}} {{t^{*}_{#1}}}	}
\DeclareDocumentCommand \sopt{ o } { \IfNoValueTF {#1} {{s^{*}}} {{s^{*}_{#1}}}	}
\DeclareDocumentCommand \toptalt{ o } { \IfNoValueTF {#1} {{\tilde{t}^{*}}} {{\tilde{t}^{*}_{#1}}}	}
\DeclareDocumentCommand \soptalt{ o } { \IfNoValueTF {#1} {{\tilde{s}^{*}}} {{\tilde{s}^{*}_{#1}}}	}
\DeclareDocumentCommand \popt{ o } { \IfNoValueTF {#1} {{p^{*}}} {{p^{*}_{#1}}}	}
\DeclareDocumentCommand \qopt{ o } { \IfNoValueTF {#1} {{q^{*}}} {{q^{*}_{#1}}}	}
\DeclareDocumentCommand \uopt{ o } { \IfNoValueTF {#1} {{u^{*}}} {{u^{*}_{#1}}}	}
\DeclareDocumentCommand \vopt{ o } { \IfNoValueTF {#1} {{v^{*}}} {{v^{*}_{#1}}}	}
\DeclareDocumentCommand \xopt{ o } { \IfNoValueTF {#1} {{x^{*}}} {{x^{*}_{#1}}}	}
\DeclareDocumentCommand \yopt{ o } { \IfNoValueTF {#1} {{y^{*}}} {{y^{*}_{#1}}}	}

\newcommand{\avec}{{\bf a}}
\newcommand{\acoord}[1][k]{{a_{#1}}}

\newcommand{\onevec}{{\bf 1}}
\newcommand{\zerovec}{{\bf 0}}
\newcommand{\zeromat}[1][d]{{{\bf 0}_{#1}}}
\newcommand{\idmat}[1][d]{{{I}_{#1}}}

\newcommand{\hadprod}{\odot}

\usetikzlibrary{matrix}

\def\ourSubfigWidthFactor{0.48}
\def\ourTiksPictureScale{1}

\def\ourTikzPictureWidthFactor{0.95} 
\def\ourTikzPictureHeightFactor{0.7}
\def\ourLegendInnerSep{0.25ex}

\def \ourPlotTitleFontSize{\normalsize}
\def \ourAxisLabelFontSize {\small}
\def \ourTickLabelFontSize {\footnotesize}
\def \ourLegendFontSize {\footnotesize}
\def \ourInternalLegendFontSize {\large}

\newcommand{\setOurGridStyle}{x grid style={dashed,black!20}, y grid style={dashed,black!20}, grid=both, ymajorgrids = true, yminorgrids = true}
\newcommand{\setOurTickStyle}{xmajorticks=true, xminorticks=false}
\newcommand{\setOurLegendStyle}{legend cell align={left}, legend style={nodes={scale=0.6, transform shape}}}

\def \ourPlotLineWidth {0.8pt}

\newcommand\ourTikzLegend[1]{\matrix[draw, matrix of nodes, anchor=south east, nodes={font=\ourLegendFontSize,anchor=west},row sep=0, column sep=0.2ex, outer sep=0, inner sep=\ourLegendInnerSep]{#1}}

\definecolor{col_NonSym_Square}{RGB}{180,0,0}
\definecolor{col_Sym_Square}{RGB}{178, 168, 43}

\definecolor{col_NonSym}{RGB}{136,65,48}
\definecolor{col_Sym_one}{RGB}{1, 62, 226}
\definecolor{col_Sym_two}{RGB}{0, 140, 41}
\definecolor{col_Sym_inf}{RGB}{132, 25, 146}

\definecolor{col_NonSym_IRLS}{RGB}{102,49,36}
\definecolor{col_Sym_two_IRLS}{RGB}{0, 105, 30}
\definecolor{col_Sym_inf_IRLS}{RGB}{99, 19, 110}

\definecolor{col_GT}{RGB}{94, 34, 8}
\definecolor{col_Procrustes}{RGB}{35, 53, 186}
\definecolor{col_IRLS}{RGB}{242, 17, 17}
\definecolor{col_IRLS_rand}{RGB}{194, 196, 35}
\definecolor{col_Ibrahim}{RGB}{237, 116, 2}

\def \lineStyleSymSquare {densely dashed}

\def \lineStyleNonSym {solid}

\def \lineStyleSym {solid}
\def \lineStyleSymIRLS {densely dashed}

\def \lineStyleSymLowerBound {densely dashed}
\def \lineStyleSymUpperBound {densely dotted}

\def \lineStyleGT {densely dotted}
\def \lineStyleProcrustes {densely dotted}
\def \lineStyleIRLS {densely dashed}
\def \lineStyleIRLSrand {densely dotted}
\def \lineStyleIbrahim {solid}

\def \markNonSym {star} \def \markSizeNonSym {3pt}
 
\def \markSymTwo {square}  \def \markSizeSymTwo {2pt}
\def \markSymInf {diamond} \def \markSizeSymInf {3pt}

\def \markFuncMapsNonSym {*} \def \markSizeFuncMapsNonSym {0.4pt}
\def \markFuncMapsSymTwo {.} \def \markSizeFuncMapsSymTwo {0.4pt}

\def \markSymTwoIRLS {square}  \def \markSizeSymTwoIRLS {1.8pt}
\def \markSymInfIRLS {diamond}  \def \markSizeSymInfIRLS {2.3pt}

\def \markSymTwoLowerBound {triangle}  \def \markSizeSymTwoLowerBound {3pt}
\def \markSymInfLowerBound {triangle}  \def \markSizeSymInfLowerBound {3pt}

\def \markSymTwoUpperBound {triangle}  \def \markSizeSymTwoUpperBound {3pt}

\def \markSymSquare {triangle} \def \markSizeSymSquare {3pt}

\def \markGT {x} \def \markSizeGT {4pt}
\def \markProcrustes {x} \def \markSizeProcrustes {4pt}
\def \markIbrahim {asterisk} \def \markSizeIbrahim {4pt}

\def \markIRLS {o} \def \markSizeIRLS {2.7pt}
\def \markIRLSrand {diamond} \def \markSizeIRLSrand {4pt}

\def\legendLabelNonSym{NonSym}

\def\legendLabelSymTwo{$\textup{SRP}_2$}
\def\legendLabelSymInf{$\textup{SRP}_{\infty}$}

\def\legendLabelSymSquare{SRP square}

\def\legendLabelSymTwoUpperBound{$\textup{SRP}_2$ upper bound}

\def\legendLabelSymTwoLowerBound{$\textup{SRP}_2$ lower bound}
\def\legendLabelSymInfLowerBound{$\textup{SRP}_{\infty}$ lower bound}

\def\legendLabelSymTwoIRLS{$\textup{SRP}_2$+IRLS}
\def\legendLabelSymInfIRLS{$\textup{SRP}_{\infty}$+IRLS}

\def\legendLabelGT{Ground Truth}
\def\legendLabelProcrustes{Procrustes}
\def\legendLabelIbrahim{AA}
\def\legendLabelIRLS{IRLS}
\def\legendLabelIRLSrand{IRLS random}

\def\legendLabelFuncMapsNonSym{NonSym}
\def\legendLabelFuncMapsSymTwo{$\textup{SRP}_2$}

\newcommand{\addPlotNonSym}[1] { \addplot [col_NonSym, \lineStyleNonSym, mark = \markNonSym, mark size = \markSizeNonSym, line width = \ourPlotLineWidth] table [x index=0, y expr=\thisrow{NonSym}*#1]{\datatable}; }

\newcommand{\addPlotSymTwo}[1] { \addplot [col_Sym_two, \lineStyleSym, mark = \markSymTwo, mark size = \markSizeSymTwo, line width = \ourPlotLineWidth] table [x index=0, y expr=\thisrow{Sym-2}*#1]{\datatable}; }

\newcommand{\addPlotSymInf}[1] { \addplot [col_Sym_inf, \lineStyleSym, mark = \markSymInf, mark size = \markSizeSymInf, line width = \ourPlotLineWidth] table [x index=0, y expr=\thisrow{Sym-inf}*#1]{\datatable}; }

\newcommand{\addPlotSymSquare}[1] { \addplot [col_Sym_Square, \lineStyleSymSquare, mark = \markSymSquare, mark size = \markSizeSymSquare, line width = \ourPlotLineWidth] table [x index=0, y expr=\thisrow{Square}*#1]{\datatable}; }

\newcommand{\addPlotSymTwoUpperBound}[1] { \addplot [col_Sym_two, \lineStyleSymUpperBound, mark = \markSymTwoUpperBound, mark size = \markSizeSymTwoUpperBound, mark options={solid,rotate=180}, line width = \ourPlotLineWidth] table [x index=0, y expr=\thisrow{p=2 upper bound}*#1]{\datatable}; }

\newcommand{\addPlotSymTwoLowerBound}[1] { \addplot [col_Sym_two, \lineStyleSymLowerBound, mark = \markSymTwoLowerBound, mark size = \markSizeSymTwoLowerBound, line width = \ourPlotLineWidth] table [x index=0, y expr=\thisrow{p=2 lower bound}*#1]{\datatable}; }

\newcommand{\addPlotSymInfLowerBound}[1] { \addplot [col_Sym_inf, \lineStyleSymLowerBound, mark = \markSymInfLowerBound, mark size = \markSizeSymInfLowerBound, line width = \ourPlotLineWidth] table [x index=0, y expr=\thisrow{p=inf lower bound}*#1]{\datatable}; }

\newcommand{\addPlotSymTwoIRLS}[1] { \addplot [col_Sym_two_IRLS, \lineStyleSymIRLS, mark = \markSymTwoIRLS, mark size = \markSizeSymTwoIRLS, line width = \ourPlotLineWidth] table [x index=0, y expr=\thisrow{Sym-2+IRLS}*#1]{\datatable}; }

\newcommand{\addPlotSymInfIRLS}[1] { \addplot [col_Sym_inf_IRLS, \lineStyleSymIRLS, mark = \markSymInfIRLS, mark size = \markSizeSymInfIRLS, line width = \ourPlotLineWidth] table [x index=0, y expr=\thisrow{Sym-inf+IRLS}*#1]{\datatable}; }

\newcommand{\addPlotGT}[1] { \addplot [col_GT, \lineStyleGT, mark = \markGT, mark size = \markSizeGT, line width = \ourPlotLineWidth] table [x index=0, y expr=\thisrow{GT}*#1]{\datatable}; }

\newcommand{\addPlotProcrustes}[1] { \addplot [col_Procrustes, \lineStyleProcrustes, mark = \markProcrustes, mark size = \markSizeProcrustes, line width = \ourPlotLineWidth] table [x index=0, y expr=\thisrow{Procrustes}*#1]{\datatable}; }

\newcommand{\addPlotIbrahim}[1] { \addplot [col_Ibrahim, \lineStyleIbrahim, mark = \markIbrahim, mark size = \markSizeIbrahim, line width = \ourPlotLineWidth] table [x index=0, y expr=\thisrow{Ibrahim-tx2}*#1]{\datatable}; }

\newcommand{\addPlotIRLS}[1] { \addplot [col_IRLS, \lineStyleIRLS, mark = \markIRLS, mark size = \markSizeIRLS, line width = \ourPlotLineWidth] table [x index=0, y expr=\thisrow{IRLS}*#1]{\datatable}; }

\newcommand{\addPlotIRLSrand}[1] { \addplot [col_IRLS_rand, \lineStyleIRLSrand, mark = \markIRLSrand, mark size = \markSizeIRLSrand, line width = \ourPlotLineWidth] table [x index=0, y expr=\thisrow{IRLS-rand}*#1]{\datatable}; }

\newcommand{\addPlotFuncMapsNonSym}[1] { \addplot [col_NonSym, densely dotted, mark = \markFuncMapsNonSym, mark size = \markSizeFuncMapsNonSym, line width = \ourPlotLineWidth] table [x index=0, y expr=\thisrow{NonSym}*#1]{\datatable}; }

\newcommand{\addPlotFuncMapsSymTwo}[1] { \addplot [col_Sym_two, \lineStyleSym, mark = \markFuncMapsSymTwo, mark size = \markSizeFuncMapsSymTwo, line width = \ourPlotLineWidth] table [x index=0, y expr=\thisrow{Sym-2}*#1]{\datatable}; }

\begin{document}
\title{Symmetrized Robust Procrustes: Constant-Factor Approximation and Exact Recovery}
\maketitle


\def\textDir{content/text}
\def\proofDir{content/proofs}
\def\tableDir{content/tables}
\def\algDir{content/algorithms}
\def\figDir{content/figures}
\def\csvDir{content/csv}


\begin{abstract}
The classical {\em Procrustes} problem is to find a rigid motion (orthogonal transformation and translation) that best aligns two given point-sets in the least-squares sense. The {\em Robust Procrustes} problem is an important variant, in which a power-1 objective is used instead of least squares to improve robustness to outliers. While the optimal solution of the least-squares problem can be easily computed in closed form, dating back to Sch{\"o}nemann (1966), no such solution is known for the power-1 problem. 
In this paper we propose a novel convex relaxation for the Robust Procrustes problem. Our relaxation enjoys several theoretical and practical advantages: Theoretically, we prove that our method provides a $\sqrt{2}$-factor approximation to the Robust Procrustes problem, and that, under appropriate assumptions, it exactly recovers the true rigid motion from point correspondences contaminated by outliers. 
In practice, we find in numerical experiments on both synthetic and real robust Procrustes problems, that our method performs similarly to the standard Iteratively Reweighted Least Squares (IRLS). However the convexity of our algorithm allows incorporating additional convex penalties, which are not readily amenable to IRLS. This turns out to be a substantial advantage, leading to improved results in high-dimensional problems, including non-rigid shape alignment and semi-supervised interlingual word translation.
\end{abstract}

\section{Introduction}\label{sec_introduction}
The (Rigid) {\em Pocrustes} problem is the problem of finding a rigid motion that aligns two given point clouds as accurately as possible in the least squares sense. Formally, given  two ordered sets of $n$ points in $\R[d]$, denoted by  $P = \br{\p[1],\ldots,\p[n]}$ and $Q = \br{\q[1],\ldots,\q[n]}$, the Procrustes problem is the optimization problem
\begin{equation}\label{pr:procrustes_rigid}
	\min_{ R \in O\br{d}, t\in \R[d] }\ \Eproc\br{R,t} \coloneqq \sum_{i=1}^n \norm{R \p[i]-\q[i]+t}^2,
\end{equation}
with $O\br{d}$ denoting the set of $d \times d$ orthogonal matrices, and $\norm{\cdot}$ denoting the $\ell_2$-norm throughout the paper. In low dimension, $d=2,3$, the Procrustes problems and its generalizations (unknown correspondences, robustness to outliers) are  well-studied problems in computer vision, graphics, and robotics \cite{yang2020teaser,dym2019linearly,zhou2016fast}, with applications in scientific disciplines such as  morphology \cite{boyer2011algorithms} and chemistry \cite{andrade2004procrustes}. The high-dimensional case $d>3$ also has many applications, including  translation tasks in NLP \cite{conneau2018word,lubin2019aligning,artetxe2018robust} or non-rigid correspondence problems in computer vision and graphics \cite{maron2016point,dym2017exact,ovsjanikov2012functional}.

From an optimization viewpoint, the rigid Procrustes problem is convenient since it has a closed-form solution that is easy to compute, as shown by the author of \cite{schonemann1966generalized}. In essence, the solution is computed by applying a Singular Value Decomposition (SVD) to a matrix created from $P$ and $Q$. A key pitfall of this approach, however, is that the least-squares objective $\Eproc$ is sensitive to outliers. Due to known results on sparse signal recovery with the $\ell_1$ norm (\cite{tibshirani1996regression,donoho2006compressed,candes2006stable}), a natural way to address this is to consider the \emph{Robust Procrustes} problem, where the least-squares objective in \eqref{pr:procrustes_rigid} is replaced by a power-1 sum of $\ell_2(\R[d]) $ norms
\begin{equation}\label{pr:robust_rigid}
	\min_{ R \in O\br{d}, t\in \R[d] }\ \E\br{R,t} \coloneqq \sum_{i=1}^n \norm{R \p[i]-\q[i]+t}.
\end{equation}

To the best of our knowledge, no known algorithm to date is guaranteed to find the global minimum of the Robust Procrustes problem \cref{pr:robust_rigid}. Perhaps the most natural way to optimize this problem is by \emph{Iteratively Reweighted Least Squares (IRLS)}, whereby problem~\cref{pr:robust_rigid} is replaced by a sequence of weighted Procrustes problems~\cref{pr:procrustes_rigid} (see \cite{babin2019analysis,aftab2015convergence} and also \cite{zhou2014spatio,bouaziz2013sparse,groenen2005improved} ). While this method often works well in practice, it may fail find a global minimizer since (in contrast with more classical IRLS applications \cite{bissantz2009convergence}) the domain of \eqref{pr:robust_rigid} is non-convex.
In fact, it seems that there were no known algorithms with theoretical guarantees for this problem until the recent paper \cite{jubran2021provably}, where a RANSAC-like algorithm was proposed. Given enough running time, this algorithm was shown to find a solution whose energy is optimal up to a multiplicative constant $(1+\sqrt{2})^d$ which depends exponentially on $d$. While in practice this algorithm works well for low-dimensional problems, its performance deteriorates as the dimension increases, which is common to RANSAC-type algorithms. 

In this paper we propose a family of simple polynomial-time algorithms with strong theoretical guarantees for the Robust Procrustes problem, denoted by \SRPp[p], $1\leq p \leq \infty$, focusing on $p=2,\infty$. First, as we discuss in \cref{sec_theory}, our \SRPtwo algorithm is guaranteed to find a solution to \cref{pr:robust_rigid} that is optimal up to a multiplicative factor of $\sqrt{2}$. For \SRPinf we prove a weaker $2\sqrt{2}$ approximation factor, thought its performance is often slightly better in practice. We stress that these constants are universal, and in particular, independent of the number of points $n$ and their dimension $d$. 

In addition, if a subset of the points in $P$ and $Q$ (the {\em inliers}) are exactly related by a rigid motion $\br{R_0,t_0}$, and they dominate the remaining points (the {\em outliers}), then our \SRP algorithm is guaranteed to recover the true rigid motion $\br{R_0,t_0}$; see \cref{subsec_recovery_guarantees} for an exact formulation. Our algorithm and theoretical results are also applicable to the related {\em Orthogonal Robust Procrustes} problem
\begin{equation}\label{pr:robust_orth}
    \min_{ R \in O\br{d} }\ \E\br{R} \coloneqq \sum_{i=1}^n \norm{R \p[i]-\q[i]},
\end{equation}
where one seeks an orthogonal transformation without translation.

The \SRP algorithms we propose follow a `relax and project' scheme. In the `relax' step, a {\em symmetrized} relaxation of \eqref{pr:robust_rigid} is optimized over the space of affine transformations. This problem is convex and can be solved globally. Its minimal objective is a \emph{lower bound} on the true minimum of \eqref{pr:robust_rigid}, and the minimizers themselves need not be a rigid motion. To obtain a feasible solution to \eqref{pr:robust_rigid}, in the `project' step the solution is projected onto the set of rigid motions. The approximation results discussed above follow from a corresponding result on the ratio between the lower bound and the objective value of the projected solution. 


One advantage of algorithms that provide lower bounds is that they can be used to estimate the accuracy of other algorithms; specifically, our method provides a lower bound on the objective obtainable by {\em any} method for the Robust Procrustes problem \cref{pr:robust_rigid}. Moreover, this bound is tight up to a factor of $\frac{1}{\sqrt{2}}$. In practice we often find it to be significantly tighter. Additionally, lower bounds may be useful for Branch-and-Bound algorithms, which rely on the availability of such bounds.



Another advantage our `relax and project' approach is that it can easily accommodate additional convex energy terms and constraints; for example, if $\Econv(A)$ is some convex energy on $\R[d\times d]$, our algorithm can be readily extended to approximately solve 
\begin{equation}\label{pr:robust_orth_convex}
	\min_{ R \in O\br{d}}\  \sum_{i=1}^n \norm{R \p[i]-\q[i]}+\Econv(R).
\end{equation}
This type of energy is particularly useful in high dimension, where the Robust Procrustes problem may be underdetermined ($n<d$) or unstable without additional regularization. For example, such energy was used by the authors of \cite{ovsjanikov2012functional} to recover non-rigid transformations of surfaces. In \cref{subsec_covariance_energy} we discuss incorporating their energy into our objective to recover an orthogonal transformation in underdetermined settings in which an additional set of unordered, unmapped points is available. In \cref{sec_numerical_experiments} we apply this approach to semi-supervised learning of interlingual word translation.

Our numerical experiments, presented in \cref{sec_numerical_experiments}, show that our method for the robust Procrustes problem has practical advantages. In comparison to \cite{jubran2021provably}, the only method known to us with theoretical approximation guarantees, our method performs better in high dimensions. This is in accordance with the fact that our approximation factor is a constant and does not depend on $d$, whereas their factor grows exponentially with $d$. 
Although having no theoretical guarantees, an IRLS approach performs on par or better than both our method and that of Jubran et al. Typically initializing IRLS with the solution of our method leads to similar results, but leads to significantly improved results in certain settings; see \cref{fig_recovery_R}. 

Our experiments further demonstrate that our ability to  incorporate additional convex energies alongside the Robust Procrustes energy is a significant advantage. We show this advantage in \cref{sec_numerical_experiments} for synthetic experiments, non-rigid correspondence problems, and NLP. For example, for interlingual translation NLP tasks with small dictionaries, we obtain a $\SI{14}{\percent}$ improvement compared to both standard Procrustes optimization and IRLS.

To summarize, the main contributions of this paper are:
\begin{enumerate}
	\item We provide a polynomial-time algorithm that achieves a $\sqrt{2}$-factor approximation for the robust Procrustes problem.
	\item We prove that our algorithm, under certain assumptions, is able to exactly recover a rigid motion from noiseless correspondences with outliers.
	\item Our algorithm provides lower bounds that can be used to evaluate solutions of other methods, and for branch and bound algorithms. 
	\item Our algorithm can handle additional convex energies, and yields state-of-the-are performance for such problems.
\end{enumerate}

\section{Method}\label{sec_method}
%
%
%
%
We first describe our method for the simpler orthogonal problem \cref{pr:robust_orth} and then move to the rigid problem \cref{pr:robust_rigid}. To this end, we define the family of relaxation to \cref{pr:robust_orth}, parametrized by $p \in \brs{1,\infty}$, 
\begin{align}\label{pr:relax_orth}
	\min_{A \in \R[d \times d]}\ \Ep\br{A} & \coloneqq \sum_{i=1}^n \br{ \frac{ \norm{A \p[i]-\q[i]}^p + \norm{A^T \q[i]-\p[i]}^p } {2} }^{1/p}
		,  1 \leq p < \infty
		\\ 
		\min_{A \in \R[d \times d]} \Ep[\infty]\br{A}& \coloneqq \sum_{i=1}^n \max \brc{ \norm{A \p[i]-\q[i]}, \norm{A^T \q[i]-\p[i]} }	\nonumber
\end{align}

with the objective $\Ep$ defined for any matrix $A \in \R[d \times d]$. Note that if $A=R$ is orthogonal, then its relaxed objective $\Ep[p]\br{R}$ for any $p$ coincides with the original $\E\br{R}$ of \eqref{pr:relax_orth}. Therefore, the optimal objective of \cref{pr:relax_orth} is a lower bound on that of the original problem~\cref{pr:robust_orth}. Problem~\eqref{pr:relax_orth} is convex for any $p \in \brs{1,\infty}$. For $p \in \brc{1,2,\infty}$, it can be formulated as a second order conic program (SOCP) \cite{boyd2004convex} and solved (globally) by standard solvers such as MOSEK \cite{mosek} or GUROBI \cite{gurobi}. In practice we use our own faster implementation, as discussed in \cref{app_implementation}.

Since the optimal solution $\Aopt$ to \cref{pr:relax_orth} will generally not be an orthogonal matrix, we project it onto the nearest orthogonal matrix $\Aoptproj=UV^T$, where $U,V$ are taken from a singular value decomposition $\Aopt=U\Sigma V^T $ of $A$.
Our \SRPp method for the orthogonal problem is summarized in \cref{alg_orthogonal}. 
\begin{algorithm}
	\caption{Orthogonal \SRPp: Relax-and-project algorithm for  Orthogonal Robust Procrustes}\label{alg_orthogonal}
\begin{algorithmic}[1]
\Input $P,Q \in \R^{d \times n}$,\quad $p \in \brs{0,\infty}$
\Output Approximate solution $\Rhat$ of Problem~\cref{pr:robust_orth}
\State Find minimizer $\Aopt$ of Problem~\cref{pr:relax_orth}
\State\Return $\Rhat \eqdef \Aoptproj$ 
\end{algorithmic}
\end{algorithm}

In this work we focus mostly on \SRPtwo and \SRPinf. By the generalized mean inequality, $\Ep\br{A}$ is monotonically increasing with respect to $p$. Thus, $p=\infty$ has the advantage of incurring the highest penalty on non-orthogonal matrices. On the other hand, we derive better approximation bounds for $p=2$, as detailed in the next section.

For the rigid problem \cref{pr:robust_rigid} we redefine the objective $\Ep$ to accommodate translations, and consider the relaxation 
%
%
%
\begin{equation}\label{pr:relax_rigid}
    \min_{A \in \R[d \times d],\ t,s\in\R[d]}\ \Ep\br{A,t,s} \coloneqq \sum_{i=1}^n \br{ \frac{ \norm{A \p[i] - \q[i] + t }^p + \norm{A^T \q[i] - \p[i] + s }^p } {2} }^{1/p}.
\end{equation}
Note that for any orthogonal matrix $R$ and vector $t \in \R[d]$, 
\begin{equation}\label{eq:objectives_coincide_rigid}
    \Ep\of{R, t, -R^T t} = \E\of{R,t}, \quad
    p \in \brs{1, \infty}.
\end{equation}
Thus, similarly to the orthogonal case, the global minimum of problem~\cref{pr:relax_rigid} is a lower bound to that of the original problem~\cref{pr:robust_rigid}.

To approximately solve the rigid problem \cref{pr:robust_rigid}, we first find an optimal solution $\br{\Aopt,\topt,\sopt}$ to \cref{pr:relax_rigid} and take $\Rhat \eqdef \Aoptproj$ as an estimate for the orthogonal part. Then, as an estimate for the translation, we take $\that$ to be the minimizer of the Robust Procrustes problem \cref{pr:robust_rigid} when $R=\Rhat $ is fixed.

\begin{algorithm}
	\caption{Relax-and-project algorithm for the Rigid Robust Procrustes problem}\label{alg_rigid}
\begin{algorithmic}[1]
\Input $P,Q \in \R^{d \times n}$,\quad $p \in \brs{0,\infty}$
\Output Approximate solution $\br{\Rhat,\that}$ of Problem~\cref{pr:robust_rigid}
\State Find minimizer $\br{\Aopt,\topt,\sopt}$ of Problem~\cref{pr:relax_rigid}
\State Set $\Rhat \eqdef \Aoptproj$
\State Find minimizer $\that$ of Problem~\cref{pr:robust_rigid} with $R = \Rhat$ fixed.
\State\Return $\br{\Rhat, \that}$ 
\end{algorithmic}
\end{algorithm}



\subsection{Covariance energy for semi-supervised Procrustes}\label{subsec_covariance_energy}
%
As discussed in \cref{sec_introduction}, it is straightforward to generalize our algorithm to optimize robust Procrustes energies with an additional convex energy $\Econv$ as in \cref{pr:robust_orth_convex}. In the orthogonal case, this is achieved by minimizing the energy $E_p(A)+\Econv(A)$ over $\R[d\times d]$ and then projecting the solution to an orthogonal matrix as before. By a similar approach, not discussed in this paper, it is possible to add convex constraints and energy $\Econv\of{R,t}$ to the rigid problem \cref{pr:robust_rigid}. 

In this work we use a specific choice of $\Econv$ that is useful for semi-supervised Procrustes problems. This choice is inspired by the operator commutativity constraint used for functional maps \cite{ovsjanikov2012functional}. 
 Suppose we are given a point cloud $\Punmapped \in \R[d \times n_P]$, sampled uniformly from a surface $S \subset \R[d]$, and another point cloud $\Qunmapped \in \R[d \times n_Q]$ sampled from a transformed version of that surface $\tilde{S} = R \circ S$, related to $S$ by an orthogonal transformation $R$. Suppose that in addition we are given a small number $n \ll n_P,n_Q$ of points $P,Q \in \R[d\times n]$ that are known to be corresponding pairs $\q[i] = R \p[i]$, possibly with some outliers. 
Such settings arise in semi-supervised learning tasks, where two unlabelled sets are available and a small number of correspondences are found manually, possibly with some labelling errors. In this case the transformation $R$ that takes $S$ to $\tilde{S}$ satisfies
\begin{equation}\label{eq:covariance_equality}
    R \cdot \cov\of{p} = \cov\of{q} R,    
\end{equation}
where $\cov\of{p}, \cov\of{q}$ are the non-centered covariance matrices of $p \sim \Unif\of{S}$, $q \sim \Unif\of{\tilde{S}}$ respectively; see derivation in \cref{subapp_covariance_energy}. We take the empirical covariance matrices $\cov\of{\Punmapped} = \frac{1}{n_P}\Punmapped \Punmapped^T$, $\cov\of{\Qunmapped} = \frac{1}{n_Q}\Qunmapped \Qunmapped^T$ as estimates of $\cov\of{p}$, $\cov\of{q}$ and incorporate the following {\em covariance energy} into our objective,
\begin{equation}\label{pr:procrustes_with_covariance_term}
    \Ecov(R)=\norm{R\cdot\cov(\Punmapped)-\cov(\Qunmapped)R}_F,
\end{equation}
with $\norm{\cdot}_F$ being the Frobenius norm. In \cref{sec_numerical_experiments} we demonstrate the effectiveness of this approach in semi-supervised recovery of an orthogonal transformation from real and synthetic data.

\section{Theoretical results}\label{sec_theory}
\subsection{Approximation Guarantees}\label{subsec_approximation_guarantees}
For motivation, consider the following question: when does our relaxation \cref{pr:relax_orth} of the orthogonal robust Procrustes problem \cref{pr:robust_orth} achieve zero energy? What about the simpler, non-symmetrized relaxation
\begin{equation}\label{pr:nonsym_orth}
	\min_{A \in \R[d\times d]}\ \sum_{i=1}^n \norm{A \p[i]-\q[i]}?
\end{equation}
Clearly, the optimal energy of problem~\cref{pr:nonsym_orth} is zero if $P$ and $Q$ are related by a linear map $A$, even if it is not orthogonal. In contrast, it is straightforward to prove
\begin{proposition*}
Let $P,Q\in \R[d\times n]$ such that the columns of $P$ span $\R[d]$. Let $\Eopt$, $\Epopt$ be the minimal objectives of \cref{pr:robust_orth,pr:relax_orth} respectively, with $p \in \brs{1,\infty}$. Then $\Eopt=0$ if and only if $\Epopt=0$.
	\end{proposition*}
\begin{proof}
By definition $0\leq \Epopt \leq \Eopt$ and thus $\Eopt = 0 \Rightarrow \Epopt=0$. In the other direction, let $\Aopt$ be a minimizer of $\Ep\of{A}$. If $\Ep\of{\Aopt}=0$, then $\br{\Aopt}^T \Aopt \p[i] = \br{\Aopt}^T \q[i] = \p[i]$ for all $i=1,\ldots,n$. Since we assumed the columns of $P$ span $\R[d]$, it follows that $A^TA=I_d$ and so $\Eopt=0$. 	
\end{proof}

The following theorem shows that the assumption that the columns of $P$ span $\R[d]$ is not necessary. 
\begin{restatable}{theorem}{RestateApproximationOrth}\label{thm:approximation_orth}
For $P,Q\in \R[d\times n]$ and $p \in \brs{2,\infty}$, let $\Aopt$ be a minimizer of $\Ep\of{A}$, and let $\Rhat = \Aoptproj$. Let $\Eopt$ be the optimal objective of the orthogonal robust Procrustes problem~\cref{pr:robust_orth}. Then 
	\begin{equation}\label{eq:optimality_inequality_orth}
		\Eopt \leq \E\br{\Rhat} \leq 2 \Ep\of{\Aopt} \leq 2 \Eopt.
	\end{equation}
	Moreover, if $p=2$ then
	\begin{equation}\label{eq:optimality_inequality_orth_p2}
		\Eopt \leq \E\br{\Rhat} \leq \sqrt{2} \Ep\of{\Aopt} \leq \sqrt{2} \Eopt.
	\end{equation}
\end{restatable}
This theorem particularly implies that for $p\geq 2$, $\Eopt=0$ if and only if $\Epopt=0$. More importantly, it shows that the ratio between the objective $\E\of{\Rhat}$ obtained by \cref{alg_orthogonal} and the best possible objective $\Eopt$ is uniformly bounded by a constant ($2$ or $\sqrt{2}$). The ratio between $\Eopt$ and our lower bound $\Ep\of{\Aopt}$ is upper-bounded by this constant as well.

\cref{thm:approximation_orth} thus provides a $\sqrt{2}$-factor approximation result for the orthogonal robust Procrustes problem. This factor is in fact the best achieveable by {\em any} relax-and-project method; see discussion in \cref{subapp_additional_approximation_guarantees}.

The following \lcnamecref{thm:approximation_rigid} provides a similar result for the rigid problem.
\begin{restatable}{theorem}{RestateApproximationRigid}\label{thm:approximation_rigid}
	For $P,Q\in \R[d\times n]$ and $p \in \brs{2,\infty}$,
	let $\br{\Aopt, \topt, \sopt}$ be a minimizer of $\Ep\of{A,t,s}$. Let $\br{\Rhat,\that}$ be the output of \cref{alg_rigid} and let $\Eopt$ be the optimal objective of Problem~\cref{pr:procrustes_rigid}. Then
	\begin{equation}\label{eq:optimality_inequality_rigid}
			\Eopt \leq \E\br{\Rhat,\that} \leq 2\sqrt{2} \, \Ep\of{\Aopt,\topt,\sopt} \leq 2\sqrt{2} \Eopt.
	\end{equation}
	Moreover, if $p=2$ then
	\begin{equation}\label{eq:optimality_inequality_rigid_p2}
		\Eopt\leq \E\br{\Rhat,\that} \leq \sqrt{2} \, \Ep[2]\of{\Aopt,\topt,\sopt} \leq \sqrt{2} \Eopt.
\end{equation}
\end{restatable}

\subsection{Recovery guarnatees}\label{subsec_recovery_guarantees}
\comment[nd]{worth discussing: (a) for fixed inliers who span $R^d$ we can always add some outliers. (b) our results imply exactness of the original problem in this setting }
We now study the problem of recovery: Suppose that a subset of the points in $P$,$Q$, indexed by $\I \subseteq \brc{1,\ldots,n}$, are related by an orthogonal transformation $R_0$ or rigid motion $\br{R_0,t_0}$. Under what condition does the minimizer of the relaxed problems \cref{pr:relax_orth,pr:relax_rigid} yield a successful recovery of $R_0$ or $\br{R_0,t_0}$? \cref{thm:approximation_orth,thm:approximation_rigid} already guarantee exact recovery when $P$ and $Q$ are related by an orthogonal or rigid transformation and are not degenerate. We now show that this is true also when $P$, $Q$ contain some outliers, provided that the inliers are more `dominant', in the following sense:
\begin{definition*}
	We say that $P,Q\in \R[d\times n]$ satisfy the {\em linear dominance-of-inliers (DIP) property} with respect to $\br{R_0,\I}$ if $\q[i] = R_0 \p[i]$ for $i\in \I$, and for any unit vector $u \in \R[d]$,
	\begin{equation}\label{eq:linear_DIP}
		\sum_{i \in \I} \abs{\bra{u,\p[i]}} > \sum_{i \in \I^C} \abs{\bra{u,\p[i]}}
		\quad \textup{ and } \quad
		\sum_{i \in \I} \abs{\bra{u,\q[i]}} > \sum_{i \in \I^C} \abs{\bra{u,\q[i]}}.
	\end{equation}
\end{definition*}
Intuitively, the linear DIP requires that the inliers, indexed by $\I$, be more dominant than the outliers along every axis. By the following theorem, this condition guarantees exact recovery in the orthogonal problem.
\begin{restatable}{theorem}{RestateRecoveryOrth}\label{thm:recovery_orth}
Suppose that $P,Q\in \R[d\times n]$ satisfy the linear DIP with respect to $\br{R_0,\I}$. Then for any $p \in \brs{1,\infty}$, the unique global minimizer of $\Ep\of{A}$ is $R_0$.
\end{restatable}

We can achieve a similar result for the rigid problem by defining the following variant of DIP, which accommodates translations.
\begin{definition*}
	We say that $P$, $Q$ satisfy the {\em affine DIP} with respect to $\br{R_0,t_0,\I}$ if $\q[i] = R_0 \p[i] + t_0$ for $i\in \I$, and for any vector $u \in \R[d]$ and any $\alpha \in \R$ such that  $\norm{u} + \abs{\alpha} > 0$,
	\begin{equation}\label{eq:affine_DIP}
		\sum_{i \in \I} \abs{\bra{u,\p[i]} + \alpha} > \sum_{i \in \I^C} \abs{\bra{u,\p[i]} + \alpha}
		\quad \textup{ and } \quad
		\sum_{i \in \I} \abs{\bra{u,\q[i]} + \alpha} > \sum_{i \in \I^C} \abs{\bra{u,\q[i]} + \alpha}.
	\end{equation}
\end{definition*}
The affine DIP is a stronger condition than the linear DIP by definition. Intuitively, it requires the inliers to be more dominant than the outliers along every line, whereas the linear DIP only considers lines that go through the origin. Note that by setting $u=0$ and $\alpha=1$ in \cref{eq:affine_DIP}, it can be seen that the affine DIP requires the number of inliers to be greater than that of outliers.

The following theorem shows that the affine DIP guarantees successful recovery in the rigid problem.
\begin{restatable}{theorem}{RestateRecoveryRigid}\label{thm:recovery_rigid}
	Suppose that $P$,$Q$ satisfy the affine DIP  with respect to $\br{R_0,t_0,\I}$. Then for any $p \in \brs{1,\infty}$, the unique global minimizer of $\Ep\of{A,t,s}$ is $\br{R_0,t_0,-R_0^T t_0}$.
\end{restatable}

In addition to the results stated here, we have several other theoretical results, stated in \cref{app_theory}; these results show that the constants in our theorems are optimal, in an appropriate sense. Proofs are in \cref{app_proofs}.

\section{Numerical experiments}\label{sec_numerical_experiments}
In this section we provide empirical evaluation of our proposed method and theoretical results. First, in \cref{subsec_empirical_demonstration_of_theory} we visualise the theoretical guarantees discussed in \cref{sec_theory}, and juxtapose them with empirical evidence. Then, in \cref{subsection_synthetic_experiments} we demonstrate the competitive performance of our method in various synthetic settings. Lastly, in \cref{subsection_applications_experiments} we evaluate our method in two applications: (a) recovery of non-rigid shape transformations by Functional Maps \cite{ovsjanikov2012functional}, and (b) learning how to translate words between natural languages in a semi-supervised setting.

\subsection{Demonstration of theoretical results}
\label{subsec_empirical_demonstration_of_theory}
\paragraph{Approximation bounds}
\begin{figure}
    \caption{Approximation bound vs. empirical results}
    \label{fig_approximation}
    \centering
    \begin{subfigure}[b]{\ourSubfigWidthFactor\linewidth}
        \begin{tikzpicture}[scale=\ourTiksPictureScale]
    \pgfplotstableread[col sep=comma,]{\csvDir/test_approximation1_energy.csv}\datatable
    \begin{axis}[
    width = \ourTikzPictureWidthFactor\textwidth,
    height= \ourTikzPictureHeightFactor\textwidth,
    \setOurGridStyle,
    \setOurLegendStyle, 
    \setOurTickStyle, 
    title style = {font=\ourPlotTitleFontSize},
    label style={font=\ourAxisLabelFontSize},
    ticklabel style = {font=\ourTickLabelFontSize},
    legend style={font=\ourInternalLegendFontSize},
    %
    title={$d=3$},
    xlabel={Num. outliers},
    ylabel={Energy},
    legend style={at={(0.02, 0.97)},anchor=north west},
    %
    xtick = {0,400,800,1200,1600},
    ]    
    %
    \addPlotSymTwoUpperBound{1}\label{legendEntry_SymTwoUpperBound}
    \addPlotIbrahim{1}\label{legendEntry_Ibrahim}
    \addPlotProcrustes{1}\label{legendEntry_Procrustes}
    \addPlotNonSym{1}\label{legendEntry_NonSym}
    \addPlotSymTwo{1}\label{legendEntry_SymTwo}
    \addPlotSymInf{1}\label{legendEntry_SymInf}
    \addPlotGT{1}\label{legendEntry_GT}
    \addPlotIRLS{1}\label{legendEntry_IRLS}
    \addPlotSymInfLowerBound{1}\label{legendEntry_SymInfLowerBound}
    \addPlotSymTwoLowerBound{1}\label{legendEntry_SymTwoLowerBound}
\end{axis}
\end{tikzpicture}
    \end{subfigure}
    \hfill
    \begin{subfigure}[b]{\ourSubfigWidthFactor\linewidth}
        \begin{tikzpicture}[scale=\ourTiksPictureScale]
    \pgfplotstableread[col sep=comma,]{\csvDir/test_approximation2_energy.csv}\datatable
    \begin{axis}[
    width = \ourTikzPictureWidthFactor\textwidth,
    height= \ourTikzPictureHeightFactor\textwidth,
    \setOurGridStyle,
    \setOurLegendStyle, 
    \setOurTickStyle, 
    title style = {font=\ourPlotTitleFontSize},
    label style={font=\ourAxisLabelFontSize},
    ticklabel style = {font=\ourTickLabelFontSize},
    legend style={font=\ourInternalLegendFontSize},
    %
    title={$d=100$},
    xlabel={Num. outliers},
    ylabel={Energy},
    legend style={at={(0.02, 0.97)},anchor=north west},
    %
    xtick = {0,200,400,600,800,1000,1200,1400}
    ]    
    %
    \addPlotSymTwoUpperBound{1}
    \addPlotIbrahim{1}
    \addPlotProcrustes{1}
    \addPlotNonSym{1}
    \addPlotSymTwo{1}
    \addPlotSymInf{1}
    \addPlotGT{1}
    \addPlotIRLS{1}
    \addPlotSymInfLowerBound{1}
    \addPlotSymTwoLowerBound{1}
\end{axis}
\end{tikzpicture}
    \end{subfigure}
    \floatfoot{Lower and upper bounds obtained by our method, together with the energies obtained by different methods for the Robust Procrustes problem \cref{pr:robust_rigid} with 200 inliers and $\SI{2}{\percent}$ noise. Left panel: $d=3$. Right panel: $d=100$.} 
    \\
    \begin{tikzpicture}[ampersand replacement=\&]
        \ourTikzLegend {
        \ref{legendEntry_SymTwoUpperBound} \& \legendLabelSymTwoUpperBound \& 
        \ref{legendEntry_NonSym} \& \legendLabelNonSym \&
        \ref{legendEntry_GT} \& \legendLabelGT \& 
        \ref{legendEntry_SymTwoLowerBound} \& \legendLabelSymTwoLowerBound \\
        \ref{legendEntry_Ibrahim} \& \legendLabelIbrahim \&
        \ref{legendEntry_SymTwo} \& \legendLabelSymTwo \&
        \ref{legendEntry_IRLS} \& \legendLabelIRLS \&
        {} \& {} \\
        \ref{legendEntry_Procrustes} \& \legendLabelProcrustes \&
        \ref{legendEntry_SymInf} \& \legendLabelSymInf \&
        \ref{legendEntry_SymInfLowerBound} \& \legendLabelSymInfLowerBound \&
        {} \& {} \\
        };  
    \end{tikzpicture}
\end{figure}
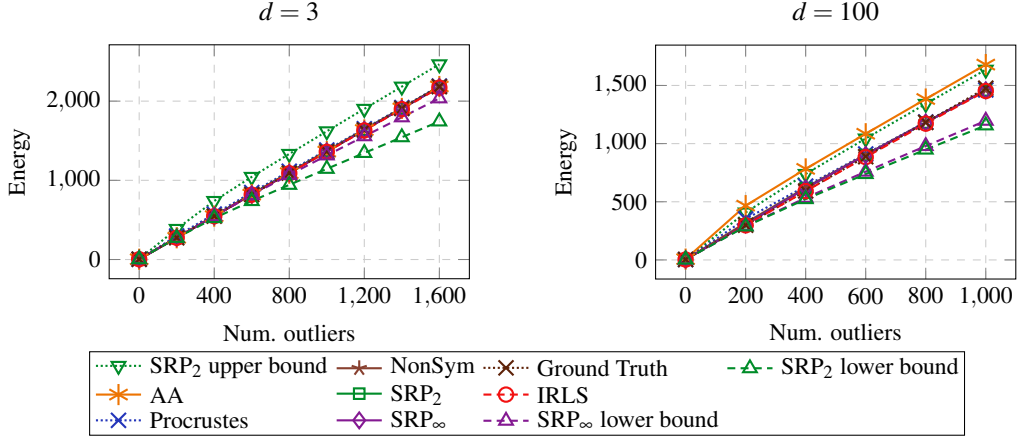

We start by illustrating the approximation bounds of \cref{subsec_approximation_guarantees}. Recall that the optimal objective value $\Ep[2](\Aopt)$ of the relaxation \cref{pr:relax_rigid} with $p=2$ provides lower bounds to the true minimum $\Eopt$ of the Robust Procrustes problem.
By \cref{thm:approximation_rigid}, the ratio between the obtained objective $\E\of{\Rhat,\that}$ and the aforementioned lower bound, is bounded by $\sqrt{2}$.
In \cref{fig_approximation} we see a numerical verification of this result. We further see that in practice most other methods also obtain energy which is lower than $\sqrt{2} \Ep[2](\Aopt)$, with the exception of the algorithm of  \cite{jubran2021provably} in high dimensions. 

Besides the \texttt{Approx-Alignment} (\methodref{\legendLabelIbrahim}) algorithm of \cite{jubran2021provably}, 
methods plotted in \cref{fig_approximation} are the standard Rigid Procrustes \cref{pr:procrustes_rigid} (\methodref{\legendLabelProcrustes}); the
non-symmetrized relaxation of \cref{pr:robust_rigid}, 
\begin{equation}\label{pr:nonsym_rigid}
    \min_{A \in \R[d\times d], t \in \R[d]}\ \sum_{i=1}^n \norm{A \p[i]+t-\q[i]},
\end{equation}
followed by orthogonal projection 
(\methodref{\legendLabelNonSym}); our proposed \methodref{\SRPtwo} and \methodref{\SRPinf}, outlined in \cref{alg_rigid}; and an \methodref{\legendLabelIRLS} scheme initialized with uniform weights. We also show the lower bounds obtained by our  relaxations \cref{pr:relax_rigid} with $p=2$ and $p=\infty$ (\methodref{\legendLabelSymTwoLowerBound}, \methodref{\legendLabelSymInfLowerBound}) and the energy of the rotation and translation used to generate the problem $\br{R_0,t_0}$ (\methodref{\legendLabelGT}). Details of the experimental setting are given in \cref{subapp_experimental_settings}. 

We conclude our discussion of \cref{pr:relax_rigid} by noting that the figure also verifies that the lower bound of \SRPinf is tighter than that of \SRPtwo, as discussed in \cref{sec_method}, and  that while most methods perform similarly in terms of their objective value; our next experiment will show that small relative difference in the energy often manifest as a significant difference in recovery performance.

\begin{figure}
    \caption{Exact recovery of $R$ in a noiseless setting}\label{fig_recovery_noiseless_R}
    \centering
    \begin{subfigure}[b]{\ourSubfigWidthFactor\linewidth}
        \begin{tikzpicture}[scale=\ourTiksPictureScale]
    \pgfplotstableread[col sep=comma,]{\csvDir/test_recovery_noiseless1_R_err_incl_adv.csv}\datatable
    \begin{axis}[
    width = \ourTikzPictureWidthFactor\textwidth,
    height= \ourTikzPictureHeightFactor\textwidth,
    \setOurGridStyle,
    \setOurLegendStyle, 
    \setOurTickStyle, 
    title style = {font=\ourPlotTitleFontSize},
    label style={font=\ourAxisLabelFontSize},
    ticklabel style = {font=\ourTickLabelFontSize},
    legend style={font=\ourInternalLegendFontSize},
    %
    title={$d=3$},
    xlabel={Num. outliers},
    ylabel={Recovery error},
    legend style={at={(0.02, 0.97)},anchor=north west},
    %
    xtick = {0, 400, 800, 1200, 1600},
    yticklabel={\pgfmathprintnumber{\tick}\si{\percent}},
    y unit={\si{\percent}},
    ymin = -4, ymax = 30,
    ]    
    %
    \addPlotProcrustes{100}
    \addPlotNonSym{100}
    \addPlotSymTwo{100}
    \addPlotSymInf{100}
    \addPlotIRLS{100}
    \addPlotIbrahim{100} 
    \addPlotSymTwoIRLS{100}
\end{axis}
\end{tikzpicture}
    \end{subfigure}
    \hfill
    \begin{subfigure}[b]{\ourSubfigWidthFactor\linewidth}
        \begin{tikzpicture}[scale=\ourTiksPictureScale]
    \pgfplotstableread[col sep=comma,]{\csvDir/test_recovery_noiseless2_R_err_incl_adv.csv}\datatable
    \begin{axis}[
    width = \ourTikzPictureWidthFactor\textwidth,
    height= \ourTikzPictureHeightFactor\textwidth,
    \setOurGridStyle,
    \setOurLegendStyle, 
    \setOurTickStyle, 
    title style = {font=\ourPlotTitleFontSize},
    label style={font=\ourAxisLabelFontSize},
    ticklabel style = {font=\ourTickLabelFontSize},
    legend style={font=\ourInternalLegendFontSize},
    %
    title={$d=100$},
    xlabel={Num. outliers},
    ylabel={Recovery error},
    legend style={at={(0.98, 0.03)},anchor=south east},
    %
    xtick = {0, 200, 400},
    yticklabel={\pgfmathprintnumber{\tick}\si{\percent}},
    y unit={\si{\percent}},
    ymin = -4, ymax = 30,
    ]    
    %
    \addPlotProcrustes{100}
    \addPlotNonSym{100}
    \addPlotSymTwo{100}
    \addPlotSymInf{100}
    \addPlotIRLS{100}
    \addPlotIbrahim{100} 
    \addPlotSymTwoIRLS{100}
\end{axis}
\end{tikzpicture}
    \end{subfigure}
    \floatfoot{Exact recovery of $R$ from 200 noiseless inliers contaminated by outliers. Left panel: $d=3$. Right panel: $d=100$.}
    \\
    \begin{tikzpicture}[ampersand replacement=\&]
        \ourTikzLegend {
        \ref{legendEntry_Ibrahim} \& \legendLabelIbrahim \&
        \ref{legendEntry_NonSym} \& \legendLabelNonSym \&
        \ref{legendEntry_SymInf} \& \legendLabelSymInf \&
        \ref{legendEntry_SymTwoIRLS} \& \legendLabelSymTwoIRLS \\
        \ref{legendEntry_Procrustes} \& \legendLabelProcrustes \& 
        \ref{legendEntry_SymTwo} \& \legendLabelSymTwo \&
        \ref{legendEntry_IRLS} \& \legendLabelIRLS \&
        {} \& {} \\
        };  
    \end{tikzpicture}    
\end{figure}
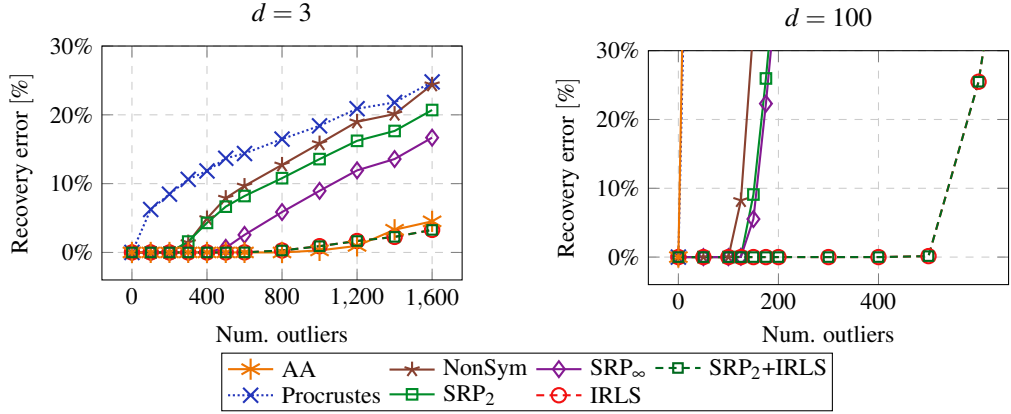

\paragraph{Exact recovery with outliers}
Next, we demonstrate the capability of our method to exactly recover a rigid motion from noiseless correspondences contaminated by outliers, as implied by \cref{thm:recovery_rigid}. In a setting similar to the above but with no noise, we evaluate several methods for their recovery of $\br{R_0,t_0}$. The error for $R$ is measured by $\norm{R-R_0}$, with $\norm{\cdot}$ denoting the spectral norm $\norm{A} = \max_{\norm{v} = 1} \norm{Av}$. The results appear in \cref{fig_recovery_noiseless_R}. Recovery errors for $t$ appear in \cref{fig_recovery_noiseless_t} in \cref{app_additional_numerical_results}.

It can be seen that all methods relying on a robust convex objective (\methodref{\legendLabelNonSym}, \methodref{\SRPtwo}, \methodref{\SRPinf}) can tolerate outliers up to a certain threshold and yield an exact recovery, with \methodref{\SRPinf} being the most resilient, followed by \methodref{\SRPtwo} -- indicating the advantage of a symmetrized objective. 
%
It is also evident that IRLS initialized by \methodref{\SRPtwo} (\methodref{\legendLabelSymTwoIRLS}), together with plain \methodref{IRLS}, perform similarly and yield superior recovery compared to all other methods.  In the next subsection we will see an example where IRLS initialized by \methodref{\SRPtwo} (\methodref{\legendLabelSymTwoIRLS}) outperforms plain \methodref{IRLS}.

Lastly, while \methodref{\legendLabelIbrahim} succeeds in finding an exact recovery in a low dimension, in a high dimension it breaks down even with a small number of outliers; this is since it depends on finding at least $d+1$ inliers at random -- an event whose probability decreases exponentially with $d$. 

\subsection{Synthetic exeriments}
\label{subsection_synthetic_experiments}
\paragraph{Recovery under noise}
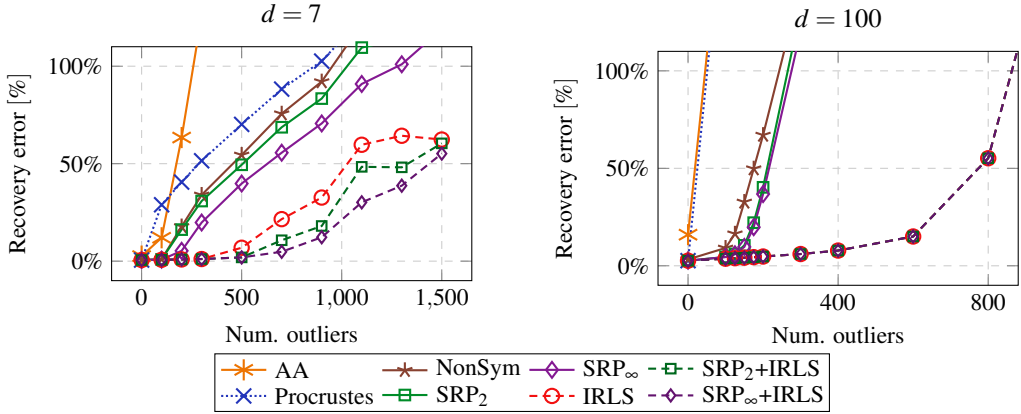
\begin{figure}
    \caption{Recovery of $R$ under noise}\label{fig_recovery_R}
    \centering
    \begin{subfigure}[b]{\ourSubfigWidthFactor\linewidth}
        \begin{tikzpicture}[scale=\ourTiksPictureScale]
    \pgfplotstableread[col sep=comma,]{\csvDir/test_recovery1_R_err_incl_adv.csv}\datatable
    \begin{axis}[
    width = \ourTikzPictureWidthFactor\textwidth,
    height= \ourTikzPictureHeightFactor\textwidth,
    \setOurGridStyle,
    \setOurLegendStyle, 
    \setOurTickStyle, 
    title style = {font=\ourPlotTitleFontSize},
    label style={font=\ourAxisLabelFontSize},
    ticklabel style = {font=\ourTickLabelFontSize},
    legend style={font=\ourInternalLegendFontSize},
    %
    title={$d=7$},
    xlabel={Num. outliers},
    ylabel={Recovery error},
    legend style={at={(0.02, 0.97)},anchor=north west},
    %
    yticklabel={\pgfmathprintnumber{\tick}\si{\percent}},
    y unit={\si{\percent}},
    ymin = -10, ymax = 110
    ]
    %
    \addPlotIbrahim{100} 
    \addPlotProcrustes{100}
    \addPlotNonSym{100}
    \addPlotSymTwo{100}
    \addPlotSymInf{100}
    \addPlotIRLS{100}
    \addPlotSymTwoIRLS{100}\label{legendEntry_SymTwoIRLS}
    \addPlotSymInfIRLS{100}\label{legendEntry_SymInfIRLS}
\end{axis}
\end{tikzpicture}
    \end{subfigure}
    \hfill
    \begin{subfigure}[b]{\ourSubfigWidthFactor\linewidth}
        \begin{tikzpicture}[scale=\ourTiksPictureScale]
    \pgfplotstableread[col sep=comma,]{\csvDir/test_recovery2_R_err_incl_adv.csv}\datatable
    \begin{axis}[
    width = \ourTikzPictureWidthFactor\textwidth,
    height= \ourTikzPictureHeightFactor\textwidth,
    \setOurGridStyle,
    \setOurLegendStyle, 
    \setOurTickStyle, 
    title style = {font=\ourPlotTitleFontSize},
    label style={font=\ourAxisLabelFontSize},
    ticklabel style = {font=\ourTickLabelFontSize},
    legend style={font=\ourInternalLegendFontSize},
    %
    title={$d=100$},
    xlabel={Num. outliers},
    ylabel={Recovery error},
    legend style={at={(0.02, 0.87)},anchor=north west},
    %
    xtick = {0, 400, 800, 1200, 1600},
    yticklabel={\pgfmathprintnumber{\tick}\si{\percent}},
    y unit={\si{\percent}},
    ymin = -10, ymax = 110
    ]
    %
    \addPlotIbrahim{100} 
    \addPlotProcrustes{100}
    \addPlotNonSym{100}
    \addPlotSymTwo{100}
    \addPlotSymInf{100}
    \addPlotIRLS{100}
    \addPlotSymTwoIRLS{100}
    \addPlotSymInfIRLS{100}
\end{axis}
\end{tikzpicture}
    \end{subfigure}
    \\
    \begin{tikzpicture}[ampersand replacement=\&]
        \ourTikzLegend {
        \ref{legendEntry_Ibrahim} \& \legendLabelIbrahim \&
        \ref{legendEntry_NonSym} \& \legendLabelNonSym \&
        \ref{legendEntry_SymInf} \& \legendLabelSymInf \&
        \ref{legendEntry_SymTwoIRLS} \& \legendLabelSymTwoIRLS \\
        \ref{legendEntry_Procrustes} \& \legendLabelProcrustes \& 
        \ref{legendEntry_SymTwo} \& \legendLabelSymTwo \&
        \ref{legendEntry_IRLS} \& \legendLabelIRLS \&
        \ref{legendEntry_SymInfIRLS} \& \legendLabelSymInfIRLS \\
        };  
    \end{tikzpicture}
    \floatfoot{Recovery of $R$ from 200 inliers with $\SI{2}{\percent}$ noise, contaminated by outliers. Left panel: $d=7$. Right panel: $d=100$.} 
\end{figure}

Here we demonstrate the ability of our method to recover a rigid motion from noisy point correspondences contaminated by outliers. Similarly to the above, we generate 200 inliers with $\SI{2}{\percent}$ noise and add outliers. We evaluate different methods for their recovery of $R$. Results for $t$ appear in \cref{fig_recovery_t} in \cref{app_additional_numerical_results}.

As in the noiseless case, the symmetrized methods \methodref{\SRPtwo}, \methodref{\SRPinf} have an overall advantage over the non-symmetrized \methodref{\legendLabelNonSym}, with a greater advantage to \methodref{\SRPinf}. Notably, \methodref{\legendLabelSymTwoIRLS} and \methodref{\legendLabelSymInfIRLS} outperform plain \methodref{IRLS} for $d=7$, whereas for $d=100$ the three methods perform similarly. Thus there exist settings where our refined method outperforms plain IRLS. However, we do find that typically IRLS yields similar results whether initialized by our methods or uniformly.

\paragraph{Semi-supervised Procrustes}
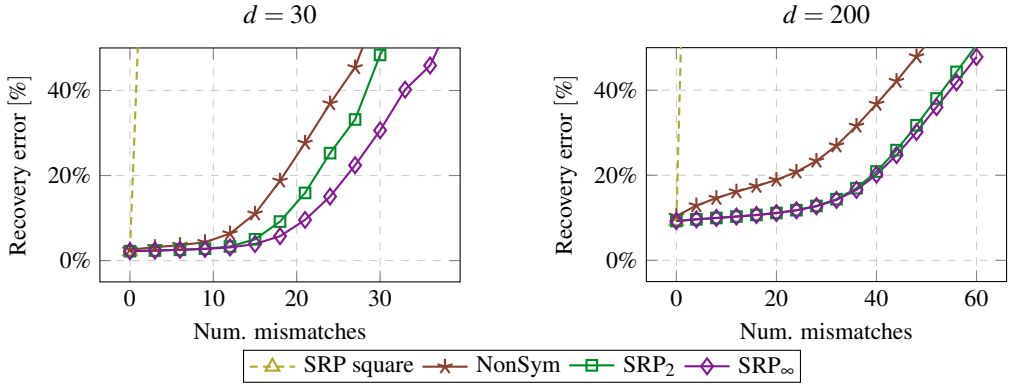
\begin{figure}
    \caption{Semi-supervised recovery}\label{fig_recovery_semisup_R}
    \centering
    \begin{subfigure}[b]{\ourSubfigWidthFactor\linewidth}
        \begin{tikzpicture}[scale=\ourTiksPictureScale]
    \pgfplotstableread[col sep=comma,]{\csvDir/test_semisup1_R_err_incl_adv.csv}\datatable
    \begin{axis}[
    width = \ourTikzPictureWidthFactor\textwidth,
    height= \ourTikzPictureHeightFactor\textwidth,
    \setOurGridStyle,
    \setOurLegendStyle, 
    \setOurTickStyle, 
    title style = {font=\ourPlotTitleFontSize},
    label style={font=\ourAxisLabelFontSize},
    ticklabel style = {font=\ourTickLabelFontSize},
    legend style={font=\ourInternalLegendFontSize},
    %
    title={$d=30$},
    xlabel={Num. mismatches},
    ylabel={Recovery error},
    legend style={at={(0.33, 0.97)},anchor=north west},
    %
    yticklabel={\pgfmathprintnumber{\tick}\si{\percent}},
    y unit={\si{\percent}},
    ymin = -5, ymax = 50,
    ]    
    %
    \addPlotSymSquare{100}\label{legendEntry_SymSquare}
    \addPlotNonSym{100}
    \addPlotSymTwo{100}
    \addPlotSymInf{100}
\end{axis}
\end{tikzpicture}
    \end{subfigure}
    \hfill
    \begin{subfigure}[b]{\ourSubfigWidthFactor\linewidth}
        \begin{tikzpicture}[scale=\ourTiksPictureScale]
    \pgfplotstableread[col sep=comma,]{\csvDir/test_semisup2_R_err_incl_adv.csv}\datatable
    \begin{axis}[
    width = \ourTikzPictureWidthFactor\textwidth,
    height= \ourTikzPictureHeightFactor\textwidth,
    \setOurGridStyle,
    \setOurLegendStyle, 
    \setOurTickStyle, 
    title style = {font=\ourPlotTitleFontSize},
    label style={font=\ourAxisLabelFontSize},
    ticklabel style = {font=\ourTickLabelFontSize},
    legend style={font=\ourInternalLegendFontSize},
    %
    title={$d=200$},
    xlabel={Num. mismatches},
    ylabel={Recovery error},
    legend style={at={(0.98, 0.03)},anchor=south east},
    %
    yticklabel={\pgfmathprintnumber{\tick}\si{\percent}},
    y unit={\si{\percent}},
    ymin = -5, ymax = 50,
    ]    
    %
    \addPlotSymSquare{100}
    \addPlotNonSym{100}
    \addPlotSymTwo{100}
    \addPlotSymInf{100}
\end{axis}
\end{tikzpicture}
    \end{subfigure}
    \\
    \begin{tikzpicture}[ampersand replacement=\&]
        \ourTikzLegend {
        \ref{legendEntry_SymSquare} \& \legendLabelSymSquare \&
        \ref{legendEntry_NonSym} \& \legendLabelNonSym \&
        \ref{legendEntry_SymTwo} \& \legendLabelSymTwo \&
        \ref{legendEntry_SymInf} \& \legendLabelSymInf \\
        };  
    \end{tikzpicture}
    \floatfoot{$R$ recovery error. Left panel: $d=30$, 16 inliers, 100 unmapped points, $\SI{1}{\percent}$ noise. Right panel: $d=200$, 40 inliers, 400 unmapped points, $\SI{2}{\percent}$ noise}
\end{figure}
The following experiment demonstrates how incorporating the covariance energy $\Ecov\of{R}$ described in \cref{subsec_covariance_energy} into our \SRP method enables the recovery of an orthogonal matrix from a small number of point correspondences $P$, $Q$, less than the dimension $d$, given larger, unordered sets of points $\Punmapped$, $\Qunmapped$ that are not matched in corresponding pairs. We evaluated our \methodref{\SRPtwo} and \methodref{\SRPinf}, adapted to minimize the robust Procrustes objective with the additional covariance term $\E\of{R}+ \lambda \Ecov\of{R}$ as in \cref{pr:procrustes_with_covariance_term}; a similar relax-and-project approach with the objective $\sum_{i=1}^n \norm{A \p[i]+t-\q[i]} + \lambda \Ecov\of{A}$, based on the non-symmetrized relaxation \cref{pr:nonsym_rigid} (\methodref{\legendLabelFuncMapsNonSym}); and a relax-and-project method based on the squared variant of $\Ep[2]\of{A}$,
\begin{equation}\label{eq:2reg}
    \min_{A \in \R[d\times d]}\ \sum_{i=1}^n \frac{1}{2}\Biggbr{ \norm{A\p[i]-\q[i]}^2 + \norm{A^T\q[i] - \p[i]}^2 } + \lambda \Ecov^2\of{A}.
\end{equation}
(\methodref{\legendLabelSymSquare}).
The results appear in \cref{fig_recovery_semisup_R}. Here the robust symmetrized methods \methodref{\SRPtwo}, \methodref{\SRPinf} have a clear advantage over \methodref{\legendLabelNonSym} in both $d=30$ and $200$. Moreover, for $d=30$, \methodref{\SRPinf} is superior to \methodref{\SRPtwo}, whereas both perform similarly at $d=200$. The squared-symmetrized method \methodref{\legendLabelSymSquare} breaks down even with a small number of outliers -- indicating the importance of using a robust objective in this setting. 

\subsection{Applications}
\label{subsection_applications_experiments}
\paragraph{Functional maps}
\begin{wrapfigure}[13]{R}{0.35\textwidth}
	\includegraphics[width=0.99\textwidth]{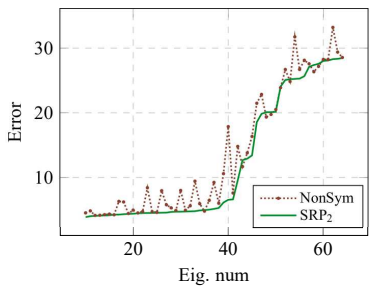}
	\caption{Functional Maps}\label{fig_functional_maps}
	
	\end{wrapfigure}


Functional Maps (FM) \cite{ovsjanikov2012functional} is a popular approach for computing non-rigid isometries between surfaces. This problem is visualized by the two surfaces (cats) in \cref{fig_functional_maps}: the ground-truth mapping between the surfaces in this problem preserves geodesic distances but not Euclidean distances. The FM approach reduces the non-rigid isometry problem in $\R[3]$ to a rigid-motion problem in high dimension, and then seeks a correspondence using the standard Procrustes energy with an additional covariance penalty term. 
 This formulation is optimized in \cite{ovsjanikov2012functional} by dropping the orthogonality constraint and solving the non-symmetrized relaxation of this problem. This solution is then refined using the ICP algorithm, which does enforce the orthogonality constraint. 

In \cref{fig_functional_maps} we show that replacing the non-symmetrized relaxation used in \cite{ovsjanikov2012functional} by our symmetrized \cref{eq:2reg} typically leads to improved results in terms of average geodesic error. Note that since this experiment does not include outliers, we only consider the squared objective as in \eqref{eq:2reg}. The setup for this experiment is described in \cref{subapp_experimental_settings}. 3D models were taken from TOSCA \cite{bronstein2008numerical}.

\paragraph{Semi-supervised translation}
\def\trigmv{\textup{trigmv}}
\def\trigexpmv{\textup{trigexpmv}}
\def\trigblock{\textup{trigblock}}
\def\expleja{\textup{expleja}}

\def\tol{\textup{tol}}
\def\tols{u_{\textup{single}}}
\def\told{u_{\textup{double}}}

\begin{table}
    \caption{Semi-supervised learning of word translation}\label{table_nlp1}
    \begingroup
    \renewcommand*{\arraystretch}{1.05}
    \begin{NiceTabular}{lcccccc}
        \toprule
        %
        \Block[c]{2-1}{Method}
    & \multicolumn{2}{c}{n = 250} & \multicolumn{2}{c}{n = 500} & \multicolumn{2}{c}{n = 1000}
    \\\cmidrule(lr){2-3}\cmidrule(lr){4-5}\cmidrule(lr){6-7}
                 & En-It  & En-De & En-It  & En-De & En-It  & En-De \\\midrule
    Procrustes   & $15.79 \pm 0.92$  & $22.90 \pm 1.23$ & $36.79 \pm 0.84$ & $40.93 \pm 0.90$ & $52.62 \pm 0.47$  & $52.91 \pm 0.54$ \\
    IRLS         & $16.60 \pm 0.94$  & $23.82 \pm 0.97$ & $38.65 \pm 0.93$ & $42.51 \pm 0.77$& $54.13 \pm 0.49$  & $\mathbf{53.71 \pm 0.53}$ \\
    NonSym       & $26.12 \pm 1.03$  & $28.91 \pm 0.64$ & $42.24 \pm 0.76$ & $41.62 \pm 0.75$ & $54.17 \pm 0.50$  & $51.77 \pm 0.51$ \\
    Sym-2        & $\mathbf{29.47 \pm 0.85}^{\dagger}$  & $\mathbf{31.94 \pm 0.59}^{\dagger}$ & $\mathbf{44.25 \pm 0.72}^{\dagger}$ & $\mathbf{43.48 \pm 0.70}$ & $\mathbf{55.08 \pm 0.50}$  & $53.24 \pm 0.55$
    \\\bottomrule
    \end{NiceTabular}
    \endgroup
    \floatfoot{\parbox{1\textwidth}{
        Translation accuracy in percents $\pm$ \SI{95}{\percent}-confidence radius. English-Italian (\texttt{En-It}) and English-German (\texttt{En-De}). The maximum in each is highlighted; significant differences ($p < 0.05$) are marked by $\dagger$. 
    }}
\end{table}
We consider the task of translating words between two natural languages. For each language we are given 200,000 words embedded in $\R[d], d=300$ by Word2Vec \cite{mikolov2013efficient}, and a dictionary with small number of correspondences between some of the words, varying between $n=200$ and $1000$\footnote{The data was downloaded from \cite{dinu2014improving,artetxe2018robust,mikolov2013efficient}.}. In \cite{xing2015normalized} this problem was approached by minimizing the Procrustes objective with $P,Q$ being the embedded corresponding words.  This approach is now quite popular, and is often used as a subroutine by unsupervised algorithms, which iteratively construct a dictionary and then solve a Procrustes problem \cite{artetxe2018robust,conneau2018word,hoshen2018non}. 

In this experiment we compare the performance of the standard Procrustes algorithm (\methodref{\legendLabelProcrustes}); the IRLS robust Procrustes algorithm (\methodref{\legendLabelIRLS}); and relax-and-project using a non-symmetrized (\methodref{\legendLabelNonSym}) and symmetrized (\methodref{\SRPtwo}) relaxation.
The results appear in \cref{table_nlp1}. We see that for small values of $n$, the methods that use the covariance energy (\methodref{\legendLabelNonSym}, \methodref{\SRPtwo}) have a significant advantage over the other methods, with our symmetrized method outperforming the non-symmetrized variant. As $n$ increases, this gap diminishes, with a slight advantage to \methodref{\legendLabelIRLS} at $n=1000$. Additional results appear in \cref{app_additional_numerical_results}, and a full discussion of the experimental setup is given in \cref{subapp_experimental_settings}.
 

\phantomsection 
\addcontentsline{toc}{section}{References} 
\bibliographystyle{unsrt}
\bibliography{main}

\newpage

\appendix
\section{Additional numerical results}\label{app_additional_numerical_results}
\begin{figure}
    \caption{Exact recovery of $t$ in a noiseless setting}\label{fig_recovery_noiseless_t}
    \centering
    \begin{subfigure}[b]{\ourSubfigWidthFactor\linewidth}
        \begin{tikzpicture}[scale=\ourTiksPictureScale]
    \pgfplotstableread[col sep=comma,]{\csvDir/test_recovery_noiseless1_t_err_incl_adv_rel.csv}\datatable
    \begin{axis}[
    width = \ourTikzPictureWidthFactor\textwidth,
    height= \ourTikzPictureHeightFactor\textwidth,
    \setOurGridStyle,
    \setOurLegendStyle, 
    \setOurTickStyle, 
    title style = {font=\ourPlotTitleFontSize},
    label style={font=\ourAxisLabelFontSize},
    ticklabel style = {font=\ourTickLabelFontSize},
    legend style={font=\ourInternalLegendFontSize},
    %
    title={$d=3$},
    xlabel={Num. outliers},
    ylabel={Recovery error},
    legend style={at={(0.02, 0.97)},anchor=north west},
    %
    xtick = {0, 400, 800, 1200, 1600},
    yticklabel={\pgfmathprintnumber{\tick}\si{\percent}},
    y unit={\si{\percent}},
    ymin = -2, ymax = 20
    ]    
    %
    \addPlotIbrahim{100} 
    \addPlotProcrustes{100}
    \addPlotNonSym{100}
    \addPlotSymTwo{100}
    \addPlotSymInf{100}
    \addPlotIRLS{100}
    \addPlotSymTwoIRLS{100}
\end{axis}
\end{tikzpicture}
    \end{subfigure}
    \hfill
    \begin{subfigure}[b]{\ourSubfigWidthFactor\linewidth}
        \begin{tikzpicture}[scale=\ourTiksPictureScale]
    \pgfplotstableread[col sep=comma,]{\csvDir/test_recovery_noiseless2_t_err_incl_adv_rel.csv}\datatable
    \begin{axis}[
    width = \ourTikzPictureWidthFactor\textwidth,
    height= \ourTikzPictureHeightFactor\textwidth,
    \setOurGridStyle,
    \setOurLegendStyle, 
    \setOurTickStyle, 
    title style = {font=\ourPlotTitleFontSize},
    label style={font=\ourAxisLabelFontSize},
    ticklabel style = {font=\ourTickLabelFontSize},
    legend style={font=\ourInternalLegendFontSize},
    %
    title={$d=100$},
    xlabel={Num. outliers},
    ylabel={Recovery error},
    legend style={at={(0.98, 0.03)},anchor=south east},
    %
    xtick = {0, 200, 400},
    yticklabel={\pgfmathprintnumber{\tick}\si{\percent}},
    y unit={\si{\percent}},
    ymin = -2, ymax = 20
    ]    
    %
    \addPlotIbrahim{100} 
    \addPlotProcrustes{100}
    \addPlotNonSym{100}
    \addPlotSymTwo{100}
    \addPlotSymInf{100}
    \addPlotIRLS{100}
    \addPlotSymTwoIRLS{100}
\end{axis}
\end{tikzpicture}
    \end{subfigure}
    \\
    \begin{tikzpicture}[ampersand replacement=\&]
        \ourTikzLegend {
        \ref{legendEntry_Ibrahim} \& \legendLabelIbrahim \&
        \ref{legendEntry_NonSym} \& \legendLabelNonSym \&
        \ref{legendEntry_SymInf} \& \legendLabelSymInf \&
        \ref{legendEntry_SymTwoIRLS} \& \legendLabelSymTwoIRLS \\
        \ref{legendEntry_Procrustes} \& \legendLabelProcrustes \& 
        \ref{legendEntry_SymTwo} \& \legendLabelSymTwo \&
        \ref{legendEntry_IRLS} \& \legendLabelIRLS \&
        {} \& {} \\
        };  
    \end{tikzpicture}    
    \floatfoot{Exact recovery of $t$ from 200 noiseless inliers contaminated by outliers. Left panel: $d=3$. Right panel: $d=100$.}
\end{figure}
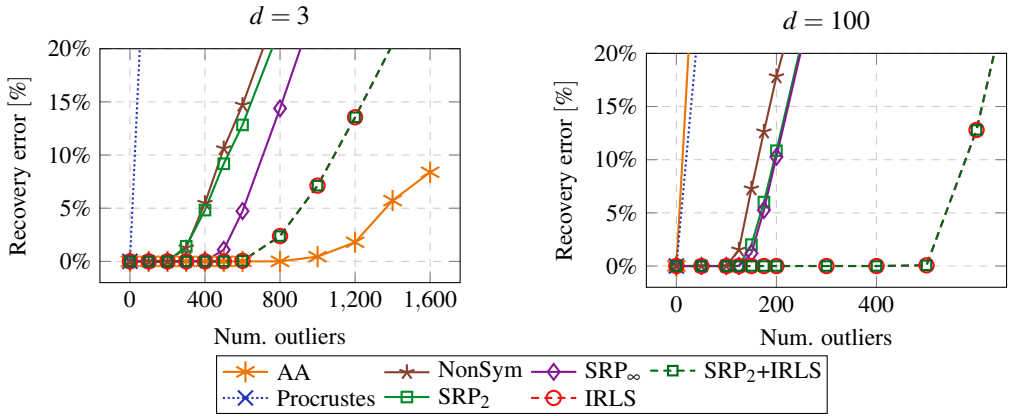

Here we present additional numerical results for the experiments described in \cref{sec_numerical_experiments}. \Cref{fig_recovery_noiseless_t} presents the recovery errors for $t$ in a noiseless setting. These results are qualitatively similar to those for $R$, shown in \cref{fig_recovery_noiseless_R}, with the exception that in low dimension ($d=3$) \methodref{\legendLabelIbrahim} significantly outperforms the remaining methods in recovering $t$, whereas it performs similarly to \methodref{\legendLabelIRLS} and \methodref{\legendLabelIRLSrand} in recovering $R$. Recovery errors are measured by $\norm{\that - t_0} / \norm{t_0}$.

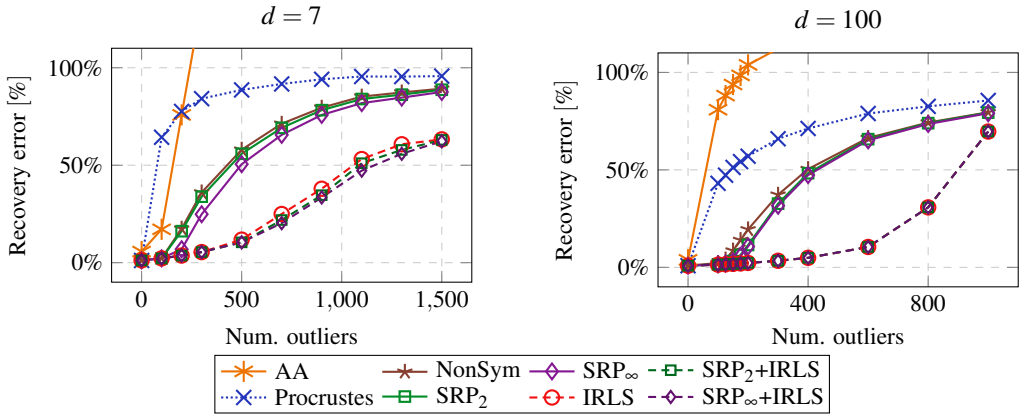
\begin{figure}
    \caption{Recovery of $t$ under noise}\label{fig_recovery_t}
    \centering
    \begin{subfigure}[b]{\ourSubfigWidthFactor\linewidth}
        \begin{tikzpicture}[scale=\ourTiksPictureScale]
    \pgfplotstableread[col sep=comma,]{\csvDir/test_recovery1_t_err_incl_adv_rel.csv}\datatable
    \begin{axis}[
    width = \ourTikzPictureWidthFactor\textwidth,
    height= \ourTikzPictureHeightFactor\textwidth,
    \setOurGridStyle,
    \setOurLegendStyle, 
    \setOurTickStyle, 
    title style = {font=\ourPlotTitleFontSize},
    label style={font=\ourAxisLabelFontSize},
    ticklabel style = {font=\ourTickLabelFontSize},
    legend style={font=\ourInternalLegendFontSize},
    %
    title={$d=7$},
    xlabel={Num. outliers},
    ylabel={Recovery error},
    legend style={at={(0.02, 0.97)},anchor=north west},
    %
    yticklabel={\pgfmathprintnumber{\tick}\si{\percent}},
    y unit={\si{\percent}},
    ymin = -10, ymax = 110,
    ]    
    %
    \addPlotIbrahim{100} 
    \addPlotProcrustes{100}
    \addPlotNonSym{100}
    \addPlotSymTwo{100}
    \addPlotSymInf{100}
    \addPlotIRLS{100}
    \addPlotSymTwoIRLS{100}
    \addPlotSymInfIRLS{100}
\end{axis}
\end{tikzpicture}
    \end{subfigure}
    \hfill
    \begin{subfigure}[b]{\ourSubfigWidthFactor\linewidth}
        \begin{tikzpicture}[scale=\ourTiksPictureScale]
    \pgfplotstableread[col sep=comma,]{\csvDir/test_recovery2_t_err_incl_adv_rel.csv}\datatable
    \begin{axis}[
    width = \ourTikzPictureWidthFactor\textwidth,
    height= \ourTikzPictureHeightFactor\textwidth,
    \setOurGridStyle,
    \setOurLegendStyle, 
    \setOurTickStyle, 
    title style = {font=\ourPlotTitleFontSize},
    label style={font=\ourAxisLabelFontSize},
    ticklabel style = {font=\ourTickLabelFontSize},
    legend style={font=\ourInternalLegendFontSize},
    %
    title={$d=100$},
    xlabel={Num. outliers},
    ylabel={Recovery error},
    legend style={at={(0.02, 0.87)},anchor=north west},
    %
    xtick = {0, 400, 800, 1200, 1600},
    yticklabel={\pgfmathprintnumber{\tick}\si{\percent}},
    y unit={\si{\percent}},
    ymin = -10, ymax = 110,
    ]    
    %
    \addPlotIbrahim{100} 
    \addPlotProcrustes{100}
    \addPlotNonSym{100}
    \addPlotSymTwo{100}
    \addPlotSymInf{100}
    \addPlotIRLS{100}
    \addPlotSymTwoIRLS{100}
    \addPlotSymInfIRLS{100}
\end{axis}
\end{tikzpicture}
    \end{subfigure}
    \\
    \begin{tikzpicture}[ampersand replacement=\&]
        \ourTikzLegend {
        \ref{legendEntry_Ibrahim} \& \legendLabelIbrahim \&
        \ref{legendEntry_NonSym} \& \legendLabelNonSym \&
        \ref{legendEntry_SymInf} \& \legendLabelSymInf \&
        \ref{legendEntry_SymTwoIRLS} \& \legendLabelSymTwoIRLS \\
        \ref{legendEntry_Procrustes} \& \legendLabelProcrustes \& 
        \ref{legendEntry_SymTwo} \& \legendLabelSymTwo \&
        \ref{legendEntry_IRLS} \& \legendLabelIRLS \&
        \ref{legendEntry_SymInfIRLS} \& \legendLabelSymInfIRLS \\
        };  
    \end{tikzpicture}
    \floatfoot{Recovery of $t$ from 200 inliers with $\SI{2}{\percent}$ noise, contaminated by outliers. Left panel: $d=7$. Right panel: $d=100$.}
\end{figure}

\Cref{fig_recovery_t} shows the recovery errors for $t$ under noise, complementary to those for $R$ shown in \cref{fig_recovery_R}. While some of the differences between the methods are smaller here compared to the results for $R$, possibly indicating that recovering $t$ is easier than  recovering $R$, these differences remain consistent.

\Cref{table_nlp2} shows translation results for English-Spanish and English-Finnish. Plots for additional values of $n$, in the settings of \cref{table_nlp1,table_nlp2}, appear in \cref{fig_nlp1,fig_nlp2} respectively. 

\begin{figure}
    \caption{Semi-supervised learning of word translations}\label{fig_nlp1}
    \centering
    \begin{subfigure}[b]{\ourSubfigWidthFactor\linewidth}
        \begin{tikzpicture}[scale=\ourTiksPictureScale]
    \pgfplotstableread[col sep=comma,]{\csvDir/test_nlp_en-it_accuracy.csv}\datatable
    \begin{axis}[
    width = \ourTikzPictureWidthFactor\textwidth,
    height= \ourTikzPictureHeightFactor\textwidth,
    \setOurGridStyle,
    \setOurLegendStyle, 
    \setOurTickStyle, 
    title style = {font=\ourPlotTitleFontSize},
    label style={font=\ourAxisLabelFontSize},
    ticklabel style = {font=\ourTickLabelFontSize},
    legend style={font=\ourInternalLegendFontSize},
    %
    title={English-Italian},
    xlabel={Num. training samples},
    ylabel={Accuracy},
    legend style={at={(0.98, 0.03)}, anchor=south east},
    %
    xtick = {200, 400, 600, 800, 1000},
    ytick = {10, 20, 30, 40, 50},
    yticklabel={\pgfmathprintnumber{\tick}\si{\percent}},
    y unit={\si{\percent}}
    ]    
    %
    \addPlotSymTwo{100}\addlegendentry{\legendLabelSymTwo}
    \addPlotNonSym{100}\addlegendentry{\legendLabelNonSym}
    \addPlotIRLS{100}\addlegendentry{\legendLabelIRLS}
    \addPlotProcrustes{100}\addlegendentry{\legendLabelProcrustes}
\end{axis}
\end{tikzpicture}
    \end{subfigure}
    \hfill
    \begin{subfigure}[b]{\ourSubfigWidthFactor\linewidth}
        \begin{tikzpicture}[scale=\ourTiksPictureScale]
    \pgfplotstableread[col sep=comma,]{\csvDir/test_nlp_en-de_accuracy.csv}\datatable
    \begin{axis}[
    width = \ourTikzPictureWidthFactor\textwidth,
    height= \ourTikzPictureHeightFactor\textwidth,
    \setOurGridStyle,
    \setOurLegendStyle, 
    \setOurTickStyle, 
    title style = {font=\ourPlotTitleFontSize},
    label style={font=\ourAxisLabelFontSize},
    ticklabel style = {font=\ourTickLabelFontSize},
    legend style={font=\ourInternalLegendFontSize},
    %
    title={English-German},
    xlabel={Num. training samples},
    ylabel={Accuracy},
    legend style={at={(0.98, 0.03)}, anchor=south east},
    %
    xtick = {200, 400, 600, 800, 1000},
    ytick = {20, 30, 40, 50},
    yticklabel={\pgfmathprintnumber{\tick}\si{\percent}},
    y unit={\si{\percent}}
    ]    
    %
    \addPlotSymTwo{100}\addlegendentry{\legendLabelSymTwo}
    \addPlotNonSym{100}\addlegendentry{\legendLabelNonSym}
    \addPlotIRLS{100}\addlegendentry{\legendLabelIRLS}
    \addPlotProcrustes{100}\addlegendentry{\legendLabelProcrustes}
\end{axis}
\end{tikzpicture}
    \end{subfigure}
    %
    \floatfoot{Learning of word translations.}
\end{figure}

\begin{figure}
    \caption{Semi-supervised learning of word translations}\label{fig_nlp2}
    \centering
    \begin{subfigure}[b]{\ourSubfigWidthFactor\linewidth}
        \begin{tikzpicture}[scale=\ourTiksPictureScale]
    \pgfplotstableread[col sep=comma,]{\csvDir/test_nlp_en-es_accuracy.csv}\datatable
    \begin{axis}[
    width = \ourTikzPictureWidthFactor\textwidth,
    height= \ourTikzPictureHeightFactor\textwidth,
    \setOurGridStyle,
    \setOurLegendStyle, 
    \setOurTickStyle, 
    title style = {font=\ourPlotTitleFontSize},
    label style={font=\ourAxisLabelFontSize},
    ticklabel style = {font=\ourTickLabelFontSize},
    legend style={font=\ourInternalLegendFontSize},
    %
    title={English-Spanish},
    xlabel={Num. training samples},
    ylabel={Accuracy},
    legend style={at={(0.98, 0.03)}, anchor=south east},
    %
    xtick = {200, 400, 600, 800, 1000},
    ytick = {10, 20, 30, 40, 50},
    yticklabel={\pgfmathprintnumber{\tick}\si{\percent}},
    y unit={\si{\percent}}
    ]    
    %
    \addPlotSymTwo{100}\addlegendentry{\legendLabelSymTwo}
    \addPlotNonSym{100}\addlegendentry{\legendLabelNonSym}
    \addPlotIRLS{100}\addlegendentry{\legendLabelIRLS}
    \addPlotProcrustes{100}\addlegendentry{\legendLabelProcrustes}
\end{axis}
\end{tikzpicture}
    \end{subfigure}
    \hfill
    \begin{subfigure}[b]{\ourSubfigWidthFactor\linewidth}
        \begin{tikzpicture}[scale=\ourTiksPictureScale]
    \pgfplotstableread[col sep=comma,]{\csvDir/test_nlp_en-fi_accuracy.csv}\datatable
    \begin{axis}[
    width = \ourTikzPictureWidthFactor\textwidth,
    height= \ourTikzPictureHeightFactor\textwidth,
    \setOurGridStyle,
    \setOurLegendStyle, 
    \setOurTickStyle, 
    title style = {font=\ourPlotTitleFontSize},
    label style={font=\ourAxisLabelFontSize},
    ticklabel style = {font=\ourTickLabelFontSize},
    legend style={font=\ourInternalLegendFontSize},
    %
    title={English-Finnish},
    xlabel={Num. training samples},
    ylabel={Accuracy},
    legend style={at={(0.98, 0.03)}, anchor=south east},
    %
    xtick = {200, 400, 600, 800, 1000},
    ytick = {10, 20, 30, 40, 50},
    yticklabel={\pgfmathprintnumber{\tick}\si{\percent}},
    y unit={\si{\percent}}
    ]    
    %
    \addPlotSymTwo{100}\addlegendentry{\legendLabelSymTwo}
    \addPlotNonSym{100}\addlegendentry{\legendLabelNonSym}
    \addPlotIRLS{100}\addlegendentry{\legendLabelIRLS}
    \addPlotProcrustes{100}\addlegendentry{\legendLabelProcrustes}
\end{axis}
\end{tikzpicture}
    \end{subfigure}
    %
    \floatfoot{Learning of word translations.}
\end{figure}

\def\trigmv{\textup{trigmv}}
\def\trigexpmv{\textup{trigexpmv}}
\def\trigblock{\textup{trigblock}}
\def\expleja{\textup{expleja}}

\def\tol{\textup{tol}}
\def\tols{u_{\textup{single}}}
\def\told{u_{\textup{double}}}

\begin{table}
    \caption{Semi-supervised learning of word translation}\label{table_nlp2}
    \begingroup
    \renewcommand*{\arraystretch}{1.05}
    \begin{NiceTabular}{lcccccc}
        \toprule
        %
        \Block[c]{2-1}{Method}
    & \multicolumn{2}{c}{n = 250} & \multicolumn{2}{c}{n = 500} & \multicolumn{2}{c}{n = 1000}
    \\\cmidrule(lr){2-3}\cmidrule(lr){4-5}\cmidrule(lr){6-7}
                 & En-Es  & En-Fi & En-Es & En-Fi & En-Es & En-Fi \\\midrule
    Procrustes   & $12.87 \pm 0.85$ & $12.21 \pm 0.86$ & $31.64 \pm 0.87$ & $25.52 \pm 0.93$ & $47.75 \pm 0.68$ & $35.46 \pm 0.39$ \\
    IRLS         & $13.62 \pm 0.65$ & $13.01 \pm 0.99$ & $33.17 \pm 0.82$ & $26.46 \pm 0.79$ & $\mathbf{49.45 \pm 0.60}$ & $\mathbf{36.54 \pm 0.48}^{\dagger}$ \\
    NonSym       & $15.08 \pm 0.99$ & $13.88 \pm 0.81$ & $31.21 \pm 0.77$ & $25.32 \pm 0.73$ & $46.20 \pm 0.69$ & $33.72 \pm 0.28$ \\
    Sym-2        & $\mathbf{19.14 \pm 1.00}^{\dagger}$ & $\mathbf{17.17 \pm 1.04}^{\dagger}$ & $\mathbf{34.44 \pm 0.82}$ & $\mathbf{27.20 \pm 0.95}$ & $48.32 \pm 0.69$ & $35.59 \pm 0.37$ 
    \\\bottomrule
    \end{NiceTabular}
    \endgroup
    \floatfoot{\parbox{1\textwidth}{
        Translation accuracy in percents $\pm$ \SI{95}{\percent}-confidence radius. English-Spanish (\texttt{En-Es}) and English-Finnish (\texttt{En-Fi}). The maximum in each is highlighted; significant differences ($p < 0.05$) are marked by $\dagger$. 
    }}
\end{table}
\section{Technical details}\label{app_technical_details}
\subsection{Experimental settings}\label{subapp_experimental_settings}

\paragraph{\Cref{fig_approximation,fig_recovery_noiseless_R,fig_recovery_R}}
We generated a random rotation matrix $R_0 \in O\of{d}$ and translation vector $t_0 \in \R[d]$. We then generated 200 random pairs of inliers $\p[i],\q[i] \in \R[d]$, related by $\q[i] = R_0 \p[i] + t_0$, to which noise was added as detailed below. We then add to $P$ and $Q$ a varying number of random outliers. All random vectors are drawn from scaled standard Gaussian distributions, with scaling parameters chosen such that for any index $i$,
\begin{equation}\label{eq_vector_norm_distribution_in_experiments}
	\mathbb{E}\brs{\norm{\p[i]}^2} = 1,            
	\quad
	\mathbb{E}\brs{\norm{t_0}^2} = \sigma_t^2,
	\quad
	\mathbb{E}\brs{\norm{R_0\p[i]+t-\q[i]}^2} = \sigma^2,
\end{equation} 
and $\mathbb{E}\brs{\norm{\q[i]}^2}$ is equal for all $i$, inlier or outlier. Thus, $\sigma$ denotes the relative noise strength and $\sigma_t$ denots the proportion of $t_0$ compared to the points in $P$. We used $\sigma_t = 0.3$ in all three figures. In the noiseless setting of \cref{fig_recovery_noiseless_R} we used $\sigma=0$, and in \cref{fig_approximation,fig_recovery_R} we used $\sigma=0.02$, making for $\SI{2}{\percent}$ noise. In each setting we tested 200 independent instances and took the average result.

Random rotation matrices were drawn uniformly over $O\of{d}$. Coordinates of random noise vectors $\xi$ were drawn i.i.d. $N\of{0,\frac{\sigma}{\sqrt{d}}}$. When a nonzero translation vector $t_0$ was used, its coordinates were drawn i.i.d. $N\of{0,\frac{\sigma_t}{\sqrt{d}}}$. The coordinates of inlier and outlier $\p[i]$ were drawn from $N\of{0,\frac{1}{\sqrt{d}}}$, and those of outlier $\q[i]$ are $N\of{0,\frac{\sqrt{1+\sigma_t^2+\sigma^2}}{\sqrt{d}}}$, thus satisfying \cref{eq_vector_norm_distribution_in_experiments}. This also implies $\mathbb{E}\brs{\norm{\q[i]}^2} = 1+\sigma_t^2+\sigma^2$ for any inlier or outlier $i$.

\paragraph{\Cref{fig_recovery_semisup_R}}
Here we simulated a setting whereby a few point correspondences are chosen manually, possibly with some labelling errors, from larger, unordered point sets that are known to correspond. First, in the same manner as described above, we generated a random rotation $R_0$ with zero translation ($\sigma_t=0$). We then generated two point-sets of noisy corresponding inliers $P,Q$ and $\Punmapped, \Qunmapped$, of sizes $n,\nunmapped$ respectively. Outliers were created to simulate `mismatches', caused by manual labelling; this was achieved by adding to $P,Q$ a varying number $n_{\textup{out}}$ of pairs $\tilde{p}^{\br{i}}, \tilde{q}^{\br{j}}$ chosen randomly from $\Punmapped,\Qunmapped$ respectively. We then reordered $\Qunmapped$ randomly so that $\Punmapped,\Qunmapped$ are no longer ordered in corresponding pairs. We gave $P,Q,\Punmapped,\Qunmapped$ as input to each of the evaluated methods. 

In the left panel of \cref{fig_recovery_semisup_R} we used $d=30$, $n=16$, $\nunmapped = 100$ and $\sigma = 0.01$. In the right panel we used $d=200$, $n = 40$, $\nunmapped = 400$ and $\sigma = 0.02$. 


The parameter $\lambda$ was set to $\lambda = \bar{\lambda} \alpha$, with $\bar{\lambda} = 0.2$, and $\alpha$ being a balancing factor, used to balance the two terms in the objective. For \methodref{\SRPtwo} we used $\alpha = \sum_{i=1}^n \frac{1}{\sqrt{2}}\sqrt{\norm{\p[i]}^2 + \norm{\q[i]}^2} / \max_{i,j \in \brs{d}} \abs{\sigma_i-\tau_j}$, with $\sigma,\tau$ denoting the singular values of $\cov\of{\Punmapped}$, $\cov\of{\Qunmapped}$ respectively. For \methodref{\SRPinf} we used 
$\alpha = \sum_{i=1}^n \max\brc{\norm{\p[i]},  \norm{\q[i]}} / \max_{i,j \in \brs{d}} \abs{\sigma_i-\tau_j}$, for \methodref{\legendLabelNonSym} 
$\alpha = \sum_{i=1}^n \norm{\p[i]} / \max_{i,j \in \brs{d}} \abs{\sigma_i-\tau_j}$
and for \methodref{\legendLabelSymSquare}
$\alpha = \sum_{i=1}^n \frac{1}{2}\br{\norm{\p[i]}^2 + \norm{\q[i]}^2} / \max_{i,j \in \brs{d}} \abs{\sigma_i-\tau_j}^2$.

\paragraph{Functional Maps} For this experiment we used the original Matlab code provided by \cite{ovsjanikov2012functional}, and solved problems in $d=50$ dimensions (corresponding to chosing the 50 smallest eigenvalues of the Laplace-Beltrami operator). We took eleven models of cats from the TOSCA dataset \cite{bronstein2008numerical} and solved the one-sided and two-sided relaxations on all 55 possible pairings of these models.  

\paragraph{Semi-supervised word translation}
 For each of the languages English, Italian, German, Spanish and Finnish, we used a dataset of 200,000 word embeddings into $\R[300]$ by Word2Vec \cite{mikolov2013efficient}, along with dictionaries from English to each of the other languages, downloaded from \cite{dinu2014improving,artetxe2018robust}. In each problem instance we randomly chose between 200 and 1000 translated words from a training dictionary of size 1500, from which the point correspondences $P$, $Q$ were created by taking the embedding of each word in the source language $\p[i]$ and its embedded translation in the target language $\q[i]$.
Together with these, we took the full set of 200,000 unmapped word embeddings from the source and target language, denoted $\Punmapped, \Qunmapped$ respectively. As a preprocessing step, $\brs{P,\Punmapped}$ and $\brs{Q, \Qunmapped}$ were centered, followed by point-normalization to the $\ell_2$ unit sphere. 

The task was to learn a rigid motion $\br{R,t}$ such that given a new embedded word $p$ in the source language, $Rp+t$ should approximate the embedded translation $q$ of that word in the target language. English was used as the source language in all tests. 

Evaluation was done on a separate test dictionary of 6000 words, using the Cross-domain Similarity Local Scaling (CSLS) measure of \cite{conneau2018word} with parameter $k=10$, as used in \cite{artetxe2018robust}. The average result over 10 independent instances was taken, together with $\SI{95}{\percent}$ confidence radii, estimated as two empirical standard errors of the mean: 
\begin{equation*}
	\textup{radius} = \frac{2}{\sqrt{10}} \sqrt{ \frac{1}{10-1} \sum_{t=1}^{10} \br{\textup{accuracy}_t - \tfrac{1}{10}\sum_{r=1}^{10}\textup{accuracy}_r}^2 }	
\end{equation*}


For \methodref{\SRPtwo}, \methodref{\legendLabelNonSym}, we set $\lambda = \bar{\lambda} \alpha$ as above, and chose $\tilde{lambda}$ for each method among $\brs{0, 0.025, 0.05, 0.1, 0.2, 0.3, 0.4, 0.6}$ using 10-fold cross validation. For each method, the validation error for each $\bar{\lambda}$ was measured by the average error $\norm{\Rhat p + \that - q}$ of the solution obtained for that $\bar{\lambda}$. The range $\brs{0.025, 0.6}$ was chosen based on a preliminary 7-fold, 2-instance cross-validation experiment using a wider range $\brs{0.01, 0.8}$, in which none of the methods chose values below $0.025$ or above $0.6$. 

We note that in the preliminary experiment mentioned above, the difference in performance between \methodref{\SRPinf} and \methodref{\SRPtwo} was negligible, hence we omitted the slower \methodref{\SRPinf} from this experiment to save running time.


%

\subsection{Implementation details}\label{app_implementation}
To minimize convex objectives $\Ep\of{A,t,s}$ for $p=2,\infty$, as well as the non-symmetrized relaxation of \cref{pr:nonsym_rigid}, we are using majorization-minimization schemes. Our code solves these problems significantly faster than CVXPY.

\subsection{Resources and running times}\label{subapp_running_times}
\paragraph*{Hardware used}
All our experiments were run on a \texttt{Lenovo Legion 7-16ACH 82N600CXIV} laptop computer with an \texttt{AMD Ryzen 9 5900HX} processor (8 cores, 3.30 to 4.60 GHz) and 32GB RAM. Our implemented solver for the robust Procrustes problem did not use a GPU.

\paragraph*{Software used}
All experiments were run on \texttt{Python 3.9.7}. 

\paragraph*{Running times}
\begin{table}
    \caption{Running times}\label{table_running_times}
    \begingroup
    \renewcommand*{\arraystretch}{1.05}
    \begin{NiceTabular}{lcccc}
        \toprule
    Figure & $d$ & \SRPtwo & \SRPinf & IRLS \\
    \midrule \\
    \cref{fig_approximation} left          &   3 & 6.9e-3 & 1.9e-1 & 4.7e-3 \\
    \cref{fig_approximation} right         & 100 & 2.1e-1 & 4.8 & 1.8e-1 \\
    \cref{fig_recovery_noiseless_R} left   &   3 & 1.5e-2 & 3.00e-1 & 1.0e-02 \\
    \cref{fig_recovery_noiseless_R} right  & 100 & 1.1 & 4.7 & 4.4e-1 \\
    \cref{fig_recovery_R} left             &   7 & 6.8e-3 & 1.8e-1 & 7.9e-3 \\
    \cref{fig_recovery_R} right            & 100 & 2.9e-1 & 5.1 & 2.6e-1 \\
    \cref{fig_recovery_semisup_R} left       &  30 & 3.9e-2 & 3.1e-1 & - \\
    \cref{fig_recovery_semisup_R} right      & 200 & 1.5 & 9.9 & - \\
    \bottomrule
    \end{NiceTabular}
    \endgroup
    \floatfoot{\parbox{1\textwidth}{
        Average running times in seconds. 
    }}
\end{table}

The running times of \SRPtwo are comparable to those of IRLS, while \SRPinf is typically slower by an order of magnitude. \Cref{table_running_times} table summarizes the average running times of our method together with those of IRLS for \cref{fig_approximation,fig_recovery_noiseless_R,fig_recovery_R,fig_recovery_semisup_R}. The times listed are the average times corresponding to the worst point of each graph. The running times for the functional maps test are around two seconds for our methods and less than a second for the non-symmetrized variant.

\section{Derivation of covariance energy}\label{subapp_covariance_energy}
Here we provide the derivation of the covariance energy discussed in \ref{subsec_covariance_energy}. 
%
Suppose that $S,\tilde{S}$ are two surfaces in $\R[d]$ such that $\tilde{S} = R \circ S$ for some orthogonal matrix $R$. Let $p,q$ be two random vectors uniformly distributed on $S$, $\tilde{S}$ respectively. Then,
\begin{equation*}
	R \cdot \mathbb{E}\brs{p p^T}
	= \mathbb{E}\brs{R p p^T R^T} R
	= \mathbb{E}\brs{\br{R p} \br{Rp}^T} R
	= \mathbb{E}\brs{q q^T} R,
\end{equation*}
with the last equality holding since $R$ is an orthogonal bijection from $S$ to $\tilde{S}$, and thus $Rp$ has the same distribution as $q$. Thus, under these assumptions, equation~\cref{eq:covariance_equality} is satisfied. Since in most applications the true matrices $\mathbb{E}\brs{p p^T}$, $\mathbb{E}\brs{q q^T}$ are not known, we use instead their empirical estimates $\cov\of{\Punmapped} = \frac{1}{n_P}\Punmapped \Punmapped^T$, $\cov\of{\Qunmapped} = \frac{1}{n_Q}\Qunmapped \Qunmapped^T$.

We note that if the points in $\tilde{P},\tilde{Q}$ are related by an orthogonal transformation $R$ followed by a permutation $\Pi$, then equation~\cref{eq:covariance_equality} holds also with the true covariances replaced by their empirical estimates, namely
\begin{equation}\label{eq:empirical_covariance_equality}
	R \cov\of{\tilde{P}} = \cov\of{\tilde{Q}} R.
\end{equation} 
To see this, suppose that $\tilde{P}, \tilde{Q} \in \R[d \times \tilde{n}]$ and that for all $i \in \brs{\tilde{n}}$, $\tilde{q}^{\br{i}} = R\tilde{p}^{\br{\Pi\of{i}}}$. Then, $\tilde{Q}=R\tilde{P}\Pi$ and
\begin{equation*}
	\frac{1}{\tilde{n}}\tilde{Q}\tilde{Q}^T R
	= \frac{1}{\tilde{n}} \br{R \tilde{P} \Pi} \br{R \tilde{P} \Pi}^T R
	= \frac{1}{\tilde{n}} \br{R \tilde{P} \Pi} \br{\Pi^T \tilde{P}^T R^T} R
	= R \br{ \frac{1}{\tilde{n}} \tilde{P} \tilde{P}^T },
\end{equation*}
and thus \cref{eq:empirical_covariance_equality} is satisfied.


\section{Additional theoretical results}\label{app_theory}
In this section we state theoretical results omitted from the main text.
\subsection{Approximation guarantees}\label{subapp_additional_approximation_guarantees}
The approximation guarantee for the orthogonal case, stated in \cref{thm:approximation_orth}, is based on bounding the ratio $\E\of{\Aoptproj} / \Ep\of{\Aopt}$ for $p \geq 2$ by $2$ or $\sqrt{2}$, assuming that $\Aopt$ is optimal for $\Ep\of{A}$. The following \lcnamecref{thm:projection_orth} shows that the optimality assumption is not necessary in order to bound this ratio by $2$. Proofs are in \cref{app_proofs}.
\begin{restatable}{lemma}{RestateProjectionOrth}\label{thm:projection_orth}
    For any $A \in \R[d\times d]$, $p \geq 2$, 
    \begin{equation}
        \E\of{\Aproj} \leq 2 \Ep\of{A}.
    \end{equation}
\end{restatable}

\paragraph{Optimality of approximation factors}
We now discuss the optimality of the approximation factors of \cref{subsec_approximation_guarantees}. First, the following \lcnamecref{thm:factors_are_optimal} shows that the factors of \cref{thm:approximation_orth,thm:projection_orth} cannot be improved.
\begin{restatable}{lemma}{RestateFactorsAreOptimal}\label{thm:factors_are_optimal}
    For any $d \geq 1$ and $p \in \brs{1,\infty}$,
    \begin{enumerate}
        \item There exist $P,Q \in \R[d \times n]$ and $A \in \R[d\times d]$ such that 
        \[ \E\br{\Aproj} = 2 \Ep\br{A} > 0. \]
        \item There exist $P,Q \in \R[d \times n]$ and $A \in \R[d\times d]$ such that $A$ is optimal for $\Ep$, and
        \[ \E\br{\Aproj} = \sqrt[\leftroot{-2}\uproot{2}p]{2}\Ep\br{A} > 0. \]
    \end{enumerate}
\end{restatable}
In particular, \cref{thm:factors_are_optimal} shows that the approximation factor of $\Ep$ with $p < 2$ is at best $2^{1/p}$, which is inferior to the factor $\sqrt{2}$ of $\Ep[2]$. We do not know whether a $\sqrt{2}$-factor approximation guarantee holds for $p > 2$. However, the factors observed in practice are often much smaller. 


We now show that for the orthogonal problem~\cref{pr:robust_orth}, $\sqrt{2}$-factor approximation in the best achieveable by any relax-and-project method; namely, any algorithm for \cref{pr:robust_orth} that returns the projected minimizer $\Aoptproj$ of a convex objective $\G\brmid{A}{P,Q}$ cannot achieve a better universal approximation factor than $\sqrt{2}$.
To this end, we make the following definition.
\begin{definition*}
    A {\em point mismatch function} is a real function $\G\brmid{A}{P,Q}$, defined for any $P,Q \in \R[d \times n]$, $n \geq 1$ and $A \in \R[d \times d]$, such that
    \begin{enumerate}
        \item $\G\brmid{A}{P,Q}$ is convex as a function of $A$.
        \item $\G$ is invariant to the order of the points. Namely, for any permutation matrix $\Pi \in \R[n \times n]$,
        \begin{equation*}
            \G\brmid{A}{P \Pi, Q \Pi} = \G\brmid{A}{P,Q}.
        \end{equation*}
        \item $\G$ is invariant to global reflections:
        \begin{equation*}
            \G\brmid{-A}{-P,Q} = \G\brmid{-A}{P,-Q} = \G\brmid{A}{P,Q}.  
        \end{equation*}
    \end{enumerate}
\end{definition*}
Note that assumption 3 is satisfied by any function that is based on metrics of the form
\begin{equation*}
    d\of{A \p[i],\q[i]}, \quad d\of{\p[i],A^T\q[i]}, \quad i=1,\ldots,d    
\end{equation*}
such that $d \of{-x,-y} = d\of{x,y}$.

By the following \lcnamecref{thm:factors_are_convex_optimal}, there exists a problem instance $P,Q$ at which any point mismatch function $\G\brmid{\argdot}{P,Q}$ has a minimizer $\Aopt$ whose best orthogonal approximation $\Aoptproj$ is $\sqrt{2}$-suboptimal in Problem~\cref{pr:robust_orth}.
\begin{restatable}{theorem}{RestateFactorsAreConvexOptimal}\label{thm:factors_are_convex_optimal}
    For $d \geq 2$ there exist points $P,Q \in \R[d \times n]$ such that for any point mismatch function $\G\brmid{A}{P,Q}$,
    \begin{enumerate}
        \item The zero matrix $\Aopt=\zeromat$ is a minimizer of $\G\brmid{\argdot}{P,Q}$.
        \item For $\Rhat = \Aoptproj$, 
            \begin{equation*}\E\of{\Rhat} = \sqrt{2} \min_{R \in O\of{d}}\ \E\of{R} > 0.\end{equation*}
    \end{enumerate}
\end{restatable}

\subsection{Recovery guarantees}\label{subapp_additional_recovery_guarantees}
As the following \lcnamecref{thm:DIP_is_optimal} shows, the linear DIP condition of \cref{thm:recovery_orth} is optimal, in the sense that it cannot be improved by a multiplicative constant and still guarantee exact, or even approximate, recovery of an orthogonal transformation.
\begin{restatable}{theorem}{RestateDIPisOptimal}\label{thm:DIP_is_optimal}
    For any $d \geq 1$ and $\varepsilon \in \br{0,1}$ there exist a rotation matrix $R_0 \in SO\br{d}$, points $P,Q \in \R[d \times n]$ and index set $\I \subseteq \brs{n}$ such that:
    \begin{enumerate}
        \item $\q[i] = R_0 \p[i]$ for $i \in \I$.
        \item For any unit vector $u \in \R[d]$,
             \begin{equation}\label{eq:almost_DIP}
                \sum_{i \in \I} \abs{\bra{u,\p[i]}} \geq \br{1-\varepsilon} \sum_{i \in \I^C} \abs{\bra{u,\p[i]}}
                \quad \textup{ and } \quad
                \sum_{i \in \I} \abs{\bra{u,\q[i]}} \geq \br{1-\varepsilon} \sum_{i \in \I^C} \abs{\bra{u,\q[i]}}.
                \end{equation}
        \item There exists a rotation matrix $R_1$ that is the unique global minimizer of $\Ep$ for any $p \geq 1$, and $\norm{R_1 - R_0}_F = \sqrt{2d}$, with $\norm{\argdot}_F$ denoting the Frobenius norm.
    \end{enumerate}
\end{restatable}
Note that \cref{thm:DIP_is_optimal} implies that even when optimizing over the set of orthogonal matrices or rotation matrices, a weaker condition than the linear DIP cannot guarantee successful recovery.

\section{Proofs}\label{app_proofs}
\subsection{Approximation guarantees}\label{subapp_proofs_approximation}
We start by proving \cref{thm:projection_orth}.
\RestateProjectionOrth*
\begin{proof}[Proof of \cref{thm:projection_orth}]
    It is enough to prove the \lcnamecref{thm:projection_orth}
    for $p=2$, since $\Ep \of A$ increases monotonically with $p$. Let $A = U \Sigma V^T$ be an SVD of $A$ such that $\Aproj = U V^T$. Let $\avec = \br{\acoord[k]}_{k=1}^d$ be the vector in $\R[d]$ consisting of the main diagonal entries of $\Sigma$, namely $\acoord[k] = \Sigma_{kk}$ for $k=1,\ldots,d$. 
    Denote 
    \begin{equation}\label{eq:def_uv}
        \uu[i] = V^T \p[i],
        \quad
        \vv[i] = U^T \q[i],
        \quad 
        i=1,\ldots,n.
    \end{equation}
    Then,
    \begin{equation}\label{pf:projection_orth:eq1}
        \begin{split}
        \Ep[2] \of A =& \frac{1}{\sqrt{2}} \sum_{i=1}^n \sqrt{ \norm{A \p[i] - \q[i]}^2 +  \norm{A^T \q[i] - \p[i]}^2}
        \\ =& \frac{1}{\sqrt{2}} \sum_{i=1}^n \sqrt{ \norm{U \Sigma V^T \p[i] - \q[i]}^2 +  \norm{V \Sigma U^T \q[i] - \p[i]}^2}
        \\ =& \frac{1}{\sqrt{2}} \sum_{i=1}^n \sqrt{ \norm{\Sigma \of{V^T \p[i]} - U^T \q[i]}^2 +  \norm{\Sigma \of{U^T \q[i]} - V^T \p[i]}^2} 
        \\ =& \frac{1}{\sqrt{2}} \sum_{i=1}^n \sqrt{ \norm{\Sigma \uu[i] - \vv[i]}^2 +  \norm{\Sigma \vv[i] - \uu[i]}^2} 
        \\ =& \frac{1}{\sqrt{2}} \sum_{i=1}^n \sqrt{ \norm{\avec \hadprod \uu[i] - \vv[i]}^2 +  \norm{\avec \hadprod \vv[i] - \uu[i]}^2}, 
        \end{split}
    \end{equation}
    where $\hadprod$ denotes the Hadamard (entrywise) product.
    By a similar derivation it can be shown that
    \begin{equation}\label{pf:projection_orth:eq2}
        \E\of{\Aproj} = \sum_{i=1}^n \norm{\uu[i]-\vv[i]}.
    \end{equation}
    By \cref{pf:projection_orth:eq1,pf:projection_orth:eq2} it is enough to show that
    \begin{equation}\label{pf:projection_orth:eq3}
        \frac{1}{\sqrt{2}} \sum_{i=1}^n \sqrt{ \norm{\avec \hadprod \uu[i] - \vv[i]}^2 +  \norm{\avec \hadprod \vv[i] - \uu[i]}^2}
        \geq \frac{1}{2} \sum_{i=1}^n \norm{\uu[i]-\vv[i]}.
    \end{equation}
    We shall show that the inequality \cref{pf:projection_orth:eq3} holds for each summand individually; namely, that
    \begin{equation*}
        \frac{1}{\sqrt{2}} \sqrt{ \norm{\avec \hadprod \uu[i] - \vv[i]}^2 +  \norm{\avec \hadprod \vv[i] - \uu[i]}^2}
        \geq \frac{1}{2} \norm{\uu[i]-\vv[i]}
    \end{equation*}
    for each $i \in \brs{n}$.
    Since the coordinates of $\avec$ are the singular values of $A$, they are nonnegative. Hence, it is enough to prove that for any $x \in \R[d]$ with nonnegative entries, and any $u,v \in \R[d]$,
    \begin{equation*}
        \frac{1}{\sqrt{2}} \sqrt{ \norm{x \hadprod u - v}^2 +  \norm{x \hadprod v - u}^2}
        \geq \frac{1}{2} \norm{u-v}.
    \end{equation*}
    Taking the square, we need to show that
    \begin{equation}\label{pf:projection_orth:eq4}
        2 \br{ \norm{x \hadprod u - v}^2 +  \norm{x \hadprod v - u}^2}
        \geq \norm{u-v}^2.
    \end{equation}
    Both sides in \cref{pf:projection_orth:eq4} are separable sums over the $d$ coordinates of $x,u,v$. Hence, it is enough to prove \cref{pf:projection_orth:eq4} for the case $d=1$. Let $u,v \in \R$ be two scalars. Let $g \of x: \R \rightarrow \R$ be given by
    \begin{equation}\label{pf:projection_orth_eq5}
        g \of x \eqdef 2 \br{ \br{ux-v}^2 + \br{vx-u}^2 } - \br{u-v}^2.
    \end{equation}
    We shall now show that $g \of x \geq 0$ for any $x \geq 0$. Rearranging terms in \cref{pf:projection_orth_eq5}, we get  
    \begin{equation}\label{pf:projection_orth_eq5b}
        \begin{split}
            g \of x = \br{u^2 + v^2}\br{1+2x^2} + \br{1-4x}2uv.
        \end{split}
    \end{equation}

    If $uv \geq 0$, then using the inequality $u^2 + v^2 \geq 2uv$ on \cref{pf:projection_orth_eq5b}, we get
    \begin{equation*}
        \begin{split}
            g \of x \geq& \br{2uv}\br{1+2x^2} + \br{1-4x}2uv
            \\=& 2uv\br{2x^2-4x+2} = 4uv\br{x-1}^2 \geq 0.
        \end{split}
    \end{equation*}
    On the other hand, if $uv < 0$, consider the following expression of $g \of x$, equivalent to \cref{pf:projection_orth_eq5b},
    \begin{equation*}
        \begin{split}
            g \of x = 2\br{u^2 + v^2}x^2 -\br{8uv}x + \br{u+v}^2.
        \end{split}
    \end{equation*}
    If $u^2+v^2=0$, then $g$ is identically zero. Otherwise, by Vieta's formula $g$ is minimized at $x^* = \tfrac{2uv}{u^2+v^2}$, which is negative by assumption. Thus, since $g$ is convex, it attains its minimum over $\left[0,\infty\right)$ at $x=0$. Therefore, for any $x \geq 0$,
    \begin{equation*}
        g\of x \geq g \of 0 = \br{u+v}^2 \geq 0,
    \end{equation*}
    which concludes the proof of the \lcnamecref{thm:projection_orth}.
\end{proof}

We now prove \cref{thm:approximation_orth}. \Cref{thm:projection_orth} will be used to prove the \lcnamecref{thm:approximation_orth} in the case $p > 2$. 
\RestateApproximationOrth*
\begin{proof}[Proof of \cref{thm:approximation_orth}]
    The inequality $\Eopt \leq \E\of{\Rhat}$ is by definition of $\Eopt$, and $\Ep\of{\Aopt}$ is a lower bound on $\Eopt$ as discussed in \cref{sec_method}. Since $\Rhat = \Aoptproj$, from \cref{thm:projection_orth} with $A = \Aopt$ we have $E\of{\Rhat} \leq 2 \Ep\of{\Aopt}$. Hence, the first part of the \lcnamecref{thm:approximation_orth} is proven.

    To prove the second part, we state the following \lcnamecref{thm:f_approximation}, to be proven below.
    \begin{lemma}\label{thm:f_approximation}
        Let $f : \R[d] \rightarrow \R$ be given by
        \begin{equation}\label{eq:def_f2}
            f \of x = \frac{1}{\sqrt{2}} \sum_{i=1}^n \sqrt{ \norm{x \hadprod \uu[i] - \vv[i]}^2 +  \norm{x \hadprod \vv[i] - \uu[i]}^2}.
        \end{equation}
        Suppose that all coordinates of $\avec$ are nonnegative, and that $\avec$ is a global minimizer of $f$. Let $\onevec \in \R[d]$ be the vector whose entries all equal 1. 
        Then \[ f \of \onevec \leq \sqrt{2} f \of \avec. \]
    \end{lemma}
    The second part of the \lcnamecref{thm:approximation_orth} follows from \cref{thm:f_approximation}. To see this, let $\uu[i],\vv[i]$ be as in \cref{eq:def_uv}. By the proof of \cref{thm:projection_orth}, specifically \cref{pf:projection_orth:eq1,pf:projection_orth:eq2}, $f \of \avec=\Ep[2] \of A$ and $f \of \onevec=\Ep[2]\of{\Aproj}$. Since $A$ is optimal for $\Ep[2]$, $\avec$ is optimal for $f$. Otherwise, suppose by contradiction there exists $\tilde{\avec} \in \R[d]$ such that $f\of{\tilde{\avec}} < f \of{\avec}$. Let $\tilde{A} = U \diag{\tilde{\avec}} V^T$. It can be verified in \cref{eq:def_f2,eq:def_uv} that $\Ep[2]\of{\tilde{A}} = f\of{\tilde{\avec}}$, and thus
    \begin{equation*}
        \Ep[2]\of{\tilde{A}} = f\of{\tilde{\avec}} < f \of \avec = \Ep[2] \of A,
    \end{equation*}
    contradicting the optimality of $A$. Thus, by \cref{thm:f_approximation},
    \begin{equation*}
        \Ep[2] \of A = f \of \avec \geq \tfrac{1}{\sqrt{2}} f \of \onevec = \tfrac{1}{\sqrt{2}} f\of{\Aproj}.
    \end{equation*}
    Let us now prove \cref{thm:f_approximation}.
    \begin{proof}
        To prove the \lcnamecref{thm:f_approximation}, we seek an upper bound on $f \of \onevec$. To this end, define for $i\in\brs{n}$, $x \in \R[d]$,
        \begin{equation}\label{eq:def_alpha}
            \alpha_i \of x = \norm{\begin{bmatrix}x \hadprod \uu[i] - \vv[i] \\ x \hadprod \vv[i] - \uu[i]\end{bmatrix}} = \sqrt{ \sum_{k=1}^d \br{x_k \uu[i][k] - \vv[i][k]}^2 + \br{x_k \vv[i][k] - \uu[i][k]}^2 }.
        \end{equation}
        Using this definition, we have
        \begin{equation}\label{eq:f_alpha_identity}
            f\of{x} = \frac{1}{\sqrt{2}} \sum_{i=1}^n \alpha_i\of{x}.
        \end{equation}
        The following \lcnamecref{thm:zero_terms_imply_a_equals_one} shall come in handly when dealing with terms $\alpha_i\of{\avec}$ that equal zero. 
    \begin{proposition}\label{thm:zero_terms_imply_a_equals_one}
        Let $\avec,u,v\in\R[d]$ such that
        \begin{equation}\label{pf:zero_terms_imply_a_equals_one:eq0}
            \norm{\avec \hadprod u - v}^2 + \norm{\avec \hadprod v - u}^2 = 0
        \end{equation}
        and suppose that all coordinates of $\avec$ are nonnegative.
        Then $u = v$. Moreover, for any $k \in \brs{d}$, if $u_k^2 + v_k^2 > 0$, then $\acoord[k] = 1$. 
    \end{proposition}
    \begin{proof}
        Let $k \in \brs{d}$. Then by \cref{pf:zero_terms_imply_a_equals_one:eq0},
        \begin{equation}\label{pf:zero_terms_imply_a_equals_one:eq1}
            \acoord[k] u_{k} - v_{k} = 0, \quad 
            \acoord[k] v_{k} - u_{k} = 0,
        \end{equation}
        and thus
        \begin{equation}\label{pf:zero_terms_imply_a_equals_one:eq2}
            \acoord[k] \br{ u_{k} - v_{k} } = v_{k} - u_{k}.
        \end{equation}
        Suppose by contradiction that $u_{k} \neq v_{k}$. Then, by \cref{pf:zero_terms_imply_a_equals_one:eq2}, $\acoord[k] = -1$, which is a contradiction. Therefore, $u = v$. Now, if $u_k^2 + v_k^2 > 0$, suppose W.L.O.G. that $u_k \neq 0$. Then by \cref{pf:zero_terms_imply_a_equals_one:eq1},
        \begin{equation*}
            \acoord[k] u_{k} = v_{k} = u_{k}
        \end{equation*}
        and thus $\acoord[k] = 1$.
    \end{proof}
    Define $\Lambda \subseteq \brs{n}$ to be the index set
    \begin{equation*}
        \Lambda = \set{i \in \brs{n}}{\alpha_i\of{\avec} \neq 0}.
    \end{equation*}
    It follows from \cref{thm:zero_terms_imply_a_equals_one} that if $i \in \brs{n} \setminus \Lambda$, then $\uu[i]=\vv[i]$. Therefore,
    \begin{equation}\label{pf:projection_orth_eq6}
        \begin{split}
            f \of \onevec 
            = &\sum_{i=1}^n \norm{\uu[i] - \vv[i]}
            = \sum_{i \in \Lambda} \norm{\uu[i] - \vv[i]}
            = \sum_{i \in \Lambda} \sqrt{ \alpha_i\of{\avec}^2 + \norm{\uu[i] - \vv[i]}^2 - \alpha_i\of{\avec}^2}
            \\ =& \sum_{i \in \Lambda} \alpha_i\of{\avec} \sqrt{ 1 + \frac{ \norm{\uu[i] - \vv[i]}^2 - \alpha_i\of{\avec}^2}{\alpha_i\of{\avec}^2}}.
        \end{split}
    \end{equation}
    According to Bernoulli's inequality, for any real number $x \geq -1$,
    \begin{equation}\label{eq:bernoulli}
        \sqrt{1+x} \leq 1 + \frac{1}{2}x.        
    \end{equation}
    Inserting \cref{eq:bernoulli} with $x = \frac{ \norm{\uu[i] - \vv[i]}^2 - \alpha_i\of{\avec}^2}{\alpha_i\of{\avec}^2}$ into \cref{pf:projection_orth_eq6} yields
    \begin{equation*}
        \begin{split}
        f \of \onevec
        \leq &\sum_{i \in \Lambda} \alpha_i\of{\avec} \br{ 1 + \frac{1}{2}\frac{ \norm{\uu[i] - \vv[i]}^2 - \alpha_i\of{\avec}^2}{\alpha_i\of{\avec}^2}}.
        \\ = & \sum_{i \in \Lambda} \alpha_i\of{\avec}
        + \frac{1}{2} \sum_{i \in \Lambda} \frac{ \norm{\uu[i] - \vv[i]}^2 - \alpha_i\of{\avec}^2}{\alpha_i\of{\avec}}.
        \\ =& \sqrt{2} f \of \avec + \frac{1}{2} \sum_{i \in \Lambda} \frac{ \norm{\uu[i] - \vv[i]}^2 - \alpha_i\of{\avec}^2}{\alpha_i\of{\avec}},
        \end{split}
    \end{equation*}
    where the last equality is by \cref{eq:f_alpha_identity}.
    Expanding the above by the definition of $\alpha_i\of{\avec}$ in \cref{eq:def_alpha}, we get
    \begin{equation*}
        f \br{\onevec} \leq \sqrt{2} f \of \avec + 
        \frac{1}{2} \sum_{i \in \Lambda} \sum_{k=1}^d \frac
        { -\acoord[k]^2 \br{ {\uu[i][k]}^2+{\vv[i][k]}^2 } + \br{4 \acoord[k] - 2} {\uu[i][k]}{\vv[i][k]} }
        {\alpha_i\of{\avec}}.
    \end{equation*}
    To prove the \lcnamecref{thm:f_approximation}, we shall show that
    \begin{equation}\label{pf:projection_orth_eq7}
        \sum_{i \in \Lambda} \sum_{k=1}^d \frac{ -\acoord[k]^2 \br{ {\uu[i][k]}^2+{\vv[i][k]}^2} + \br{4 \acoord[k] - 2} \uu[i][k]\vv[i][k] }{\alpha_i\of{\avec}} \leq 0.
    \end{equation}

    To prove \cref{pf:projection_orth_eq7}, we wish to express the directional derivatives of $f\of{x}$. Recall that for a function $g: \R[d] \rightarrow \R$, the directional derivative of $g \of x$ at $x=x_0$ in the direction $w \in \R[d]$ is defined as the limit
    \begin{equation*}
        \nabla_w g\br{x_0} =
        \lim_{t \rightarrow 0^+} \frac{g\br{x_0+tw} - g\br{x_0}}{t}. 
    \end{equation*}
    
    Using elementary calculus, it can be shown that for any $i$ with $\alpha_i\of{\avec} \neq 0$, and any vector $w \in \R[d]$, the directional derivative of $\alpha_i\of{\avec}$ in the direction $w$ is given by
    \begin{equation*}
        \begin{split}
            \nabla_w \alpha_i\of{\avec}
            =& \frac{ \begin{bmatrix} \avec \hadprod \uu[i] -\vv[i] \\ \avec \hadprod \vv[i] - \uu[i] \end{bmatrix}^T 
            \begin{bmatrix} \diag{\uu[i]} \\ \diag{\vv[i]} \end{bmatrix}  w}
            { \norm{ \begin{bmatrix} \avec \hadprod \uu[i] -\vv[i] \\ \avec \hadprod \vv[i] - \uu[i] \end{bmatrix} } }
            = \frac{ \begin{bmatrix} \avec \hadprod \uu[i] -\vv[i] \\ \avec \hadprod \vv[i] - \uu[i] \end{bmatrix}^T 
            \begin{bmatrix} \uu[i] \hadprod w \\ \vv[i] \hadprod w \end{bmatrix} }
            { \alpha_i\of{\avec} }
            \\ =& \sum_{k=1}^d \frac{ \acoord[k] w_k \br{{\uu[i][k]}^2 + {\vv[i][k]}^2} -2 w_k \uu[i][k] \vv[i][k] } {\alpha_i\of{\avec}}.
        \end{split}
    \end{equation*}
    To prove the \lcnamecref{thm:approximation_orth}, we consider the directional derivative of $f \of \avec$ in the direction $\avec - \onevec$. 
    Setting $w = \avec - \onevec$ in the above equation yields that
    \begin{equation}\label{pf:f_approximation:eq1}
        \nabla_{\avec-\onevec} \alpha_i\of{\avec} 
        = \sum_{k=1}^d \frac{ \br{\acoord[k]^2-\acoord[k]} \br{{\uu[i][k]}^2 + {\vv[i][k]}^2} + \br{2 - 2 \acoord[k]} \uu[i][k] \vv[i][k] } {\alpha_i\of{\avec}}, \quad i \in \Lambda.
    \end{equation} 
    Let us now calculate $\nabla_{\avec-\onevec} \alpha_i\of{\avec}$ for $i \in \brs{n} \setminus \Lambda$.
    Suppose that $\alpha_i\of{\avec} = 0$. Then
    \begin{equation}\label{pf:zero_terms_imply_a_equals_one:eq3}
        \begin{split}
            \nabla_{\avec-\onevec} \alpha_i\of{\avec}
            =& \lim_{t \to 0^+} \frac{\alpha_i\of{\avec + t \br{\avec-\onevec}} - \alpha_i\of{\avec} } {t}
            = \lim_{t \to 0^+} \frac{\alpha_i\of{\avec + t \br{\avec-\onevec}} } {t}
            \\ =& \lim_{t \to 0^+} \frac{ \norm{\begin{bmatrix}\avec \hadprod \uu[i] - \vv[i] + t \br{\avec-\onevec} \hadprod \uu[i] \\ \avec \hadprod \vv[i] - \uu[i] + t \br{\avec-\onevec} \hadprod \vv[i] \end{bmatrix}} } {t}
            = \lim_{t \to 0^+} \frac{ \norm{\begin{bmatrix}t \br{\avec-\onevec} \hadprod \uu[i] \\ t \br{\avec-\onevec} \hadprod \vv[i] \end{bmatrix}} } {t}
            \\ =& \norm{\begin{bmatrix} \br{\avec-\onevec} \hadprod \uu[i] \\ \br{\avec-\onevec} \hadprod \vv[i] \end{bmatrix}}.
        \end{split}
    \end{equation}
    By \cref{thm:zero_terms_imply_a_equals_one}, for any $i \in \brs{n}$ such that $\alpha_i\of{\avec} = 0$, and any $k \in \brs{d}$, $\acoord[k] = 1$ or $\uu[i][k] = \vv[i][k] = 0$. Therefore, by \cref{pf:zero_terms_imply_a_equals_one:eq3}
    \begin{equation}\label{pf:zero_terms_imply_a_equals_one:eq4}
        \nabla_{\avec-\onevec} \alpha_i\of{\avec} = 0, \quad i \in \brs{n} \setminus \Lambda.
    \end{equation}
    In conclusion, from \cref{pf:f_approximation:eq1,pf:zero_terms_imply_a_equals_one:eq4} it follows that
    \begin{equation}\label{pf:zero_terms_imply_a_equals_one:eq5}
        \begin{split}
            \nabla_{\avec-\onevec}f \of \avec 
            = &\frac{1}{\sqrt{2}} \sum_{i =1}^n \nabla_{\avec-\onevec} \alpha_i \of \avec 
            \\ = &\frac{1}{\sqrt{2}} \sum_{i \in \Lambda} \sum_{k=1}^d 
            \frac{ \br{\acoord[k]^2-\acoord[k]} \br{{\uu[i][k]}^2 + {\vv[i][k]}^2} + \br{2 - 2 \acoord[k]} \uu[i][k] \vv[i][k] } {\alpha_i\of{\avec}}.                
        \end{split}
    \end{equation}
    
    Recall that $\avec$ is optimal for $f$ by assumption. Therefore, all directional derivatives of $f \of x$ at $x=\avec$ are nonnegative. In particular,
    \begin{equation}\label{pf:zero_terms_imply_a_equals_one:eq6}
        \nabla_{\avec-\onevec}f \of \avec \geq 0.
    \end{equation}
    Inserting \cref{pf:zero_terms_imply_a_equals_one:eq6} into \cref{pf:zero_terms_imply_a_equals_one:eq5} and rearranging terms yields 
    \begin{equation*}
        \begin{split}            
        &\sum_{i \in \Lambda} \sum_{k=1}^d 
            \frac{ -\acoord[k]^2 \br{{\uu[i][k]}^2 + {\vv[i][k]}^2} + \br{4 \acoord[k]- 2} \uu[i][k] \vv[i][k] } {\alpha_i\of{\avec}}
        \leq \sum_{i \in \Lambda} \sum_{k=1}^d 
        \frac{ -\acoord[k] \br{{\uu[i][k]}^2 + {\vv[i][k]}^2} + 2 \acoord[k] \uu[i][k] \vv[i][k] } {\alpha_i\of{\avec}}
        \\ = & \sum_{i \in \Lambda} \sum_{k=1}^d \frac { -a_k \br{ \uu[i][k] - \vv[i][k] }^2 } { \alpha_i\of{\avec} }
        \leq 0,
        \end{split}
    \end{equation*}
    where the last inequality holds since $a_k \geq 0$ for $k \in \brs{d}$.
    Hence \cref{pf:projection_orth_eq7} holds, and \cref{thm:f_approximation} is proven.
    \end{proof}
    This concludes the proof of \cref{thm:approximation_orth}.
\end{proof}

We shall now prove \cref{thm:approximation_rigid}.
\RestateApproximationRigid*
\begin{proof}[Proof of \cref{thm:approximation_rigid}]
    We first prove \cref{eq:optimality_inequality_rigid} for a general $p \in \brs{2,\infty}$ and then prove the improved bound \cref{eq:optimality_inequality_rigid_p2} for $p=2$. 
    
    Define the translated objective $\Fp\of{A,\p[0],\q[0]}$ for $A \in \R[d\times d]$, $\p[0],\q[0] \in \R[d]$ by
    \begin{equation}\label{eq:def_F}
        \begin{gathered}
            \Fp\of{A,\p[0],\q[0]} = 
            \\\sum_{i=1}^n \br{ \frac{ \norm{A \br{\p[i] - \p[0]} - \br{\q[i]-\q[0]}}^p + \norm{A^T \br{\q[i]-\q[0]}- \br{\p[i]-\p[0]}}^p} {2} }^{\frac{1}{p}}.
        \end{gathered}
    \end{equation}
    Namely, $\Fp\of{A,\p[0],\q[0]}$ is similar to the objective $\Ep\br{A}$, with the points $P,Q$ translated by $-\p[0],-\q[0]$ respectively. Then for any $p \in \brs{1,\infty}$, 
    \begin{equation}\label{proof:approximation_rigid:eq01}
        \begin{split}
            \E\of{\Rhat,\that}
            \\\overeq{a}&\ \min_{t \in \R[d]}\ \E\of{\Rhat,t} 
            \\=&\ \min_{t \in \R[d]}\ \sum_{j=1}^n \norm{\Rhat \p[j] - \q[j] + t} 
            \\\overleq{b} &\ \min_{i \in \brs{n}}\ \sum_{j=1}^n \norm{\Rhat \p[j] - \q[j] - \br{\Rhat \p[i] - \q[i]}}
            \\= &\ \min_{i \in \brs{n}}\ \sum_{j=1}^n \norm{\Rhat \br{\p[j]-\p[i]} - \br{\q[j]-\q[i]}}
            \\= &\ \min_{i \in \brs{n}}\ \Fp\of{\Rhat,\p[i],\q[i]},
        \end{split}
    \end{equation}
    where (a) is by the definition of $\that$, and (b) can be seen by taking $t = -\br{\Rhat \p[i] - \q[i]}$.
    
    Recall that $\Rhat = \Aoptproj$. Invoking \cref{thm:approximation_orth} for each $i \in \brs{n}$ on the translated objective $\Fp\of{\argdot, \p[i], \q[i]}$ with $p=2$ yields
    \begin{equation}\label{proof:approximation_rigid:eq02}
        \Fp[2]\of{\Rhat,\p[i],\q[i]} \leq 2 \Fp[2]\of{\Aopt,\p[i],\q[i]}, \quad i \in \brs{d}.
    \end{equation}
    Therefore, from \cref{proof:approximation_rigid:eq01} with $p=2$ and \cref{proof:approximation_rigid:eq02}, we have
    \begin{equation}\label{proof:approximation_rigid:eq03}
        \begin{split}
            \E\of{\Rhat,\that}
            \leq&\ 2 \min_{i \in \brs{n}}\ \Fp[2]\of{\Aopt,\p[i],\q[i]}
            \\ \overeq{a}&\ 2 \min_{i \in \brs{n}}\ \sum_{j=1}^n \frac{1}{\sqrt{2}} \norm{\begin{bmatrix} \Aopt \br{\p[j]-\p[i]} - \br{\q[j]-\q[i]} \\ \Aopt^T \br{\q[j]-\q[i]} - \br{\p[j]-\p[i]} \end{bmatrix}}
            \\ =&\ 2 \min_{i \in \brs{n}}\ \sum_{j=1}^n \frac{1}{\sqrt{2}} \norm{
                \begin{bmatrix} \Aopt \p[j] - \q[j] \\ \Aopt^T \q[j] - \p[j] \end{bmatrix}
                - \begin{bmatrix} \Aopt \p[i] - \q[i] \\ \Aopt^T \q[i] - \p[i] \end{bmatrix}
                },
        \end{split}
    \end{equation}
    where (a) is by a reformulation of \cref{eq:def_F} with $p=2$.
    
    Let us now state a \lcnamecref{thm:translated_objective}, to be proven below.
    \begin{lemma}\label{thm:geomed}
        Let $\br{\x[i]}_{i=1}^n$ be points in $\R[d]$. Then
        \begin{equation}\label{eq:geomean}
            \min_{i \in \brs{n}}\ \sum_{j=1}^n \norm{\x[j]-\x[i]} 
            \leq \sqrt{2} \min_{v \in \R[d]}\ \sum_{j=1}^n \norm{\x[j]-v}.
        \end{equation}
    \end{lemma}
\comment[nd]{Perhaps mention comparison with Kimmel's factor 2 approximation}
    \noindent Using \cref{thm:geomed} with 
    $\x[i] = \brs{\begin{smallmatrix} \Aopt \p[i] - \q[i] \\ \Aopt^T \q[i] - \p[i] \end{smallmatrix}}$, $i \in \brs{n}$ and $v = -\begin{bmatrix} t \\ s \end{bmatrix}$ yields
    \begin{equation}\label{proof:approximation_rigid:eq04}
        \begin{split}
            &\min_{i \in \brs{n}}\ \sum_{j=1}^n \frac{1}{\sqrt{2}} \norm{
                \begin{bmatrix} \Aopt \p[j] - \q[j] \\ \Aopt^T \q[j] - \p[j] \end{bmatrix}
                - \begin{bmatrix} \Aopt \p[i] - \q[i] \\ \Aopt^T \q[i] - \p[i] \end{bmatrix}}
            \\ \leq\ &\sqrt{2} \min_{t,s \in \R[d]}\ \sum_{j=1}^n \frac{1}{\sqrt{2}} \norm{
               \begin{bmatrix} \Aopt \p[j] - \q[j] \\ \Aopt^T \q[j] - \p[j] \end{bmatrix}
             + \begin{bmatrix} t \\ s \end{bmatrix}    }
             \\ \overeq{a}\ &\sqrt{2} \min_{t,s\in \R[d]}\ \Ep[2]\of{\Aopt,t,s},
        \end{split}        
    \end{equation}
    where (a) is by a reformulation of \cref{pr:relax_rigid}. Hence, by \cref{proof:approximation_rigid:eq03,proof:approximation_rigid:eq04},
    \begin{equation}
        \begin{split}
            \E\of{\Rhat,\that}
            \leq\ &2 \sqrt{2} \min_{t,s\in \R[d]}\ \Ep[2]\of{\Aopt,t,s}                
            \\ \overleq{a}\ &2 \sqrt{2} \min_{t,s\in \R[d]}\ \Ep[p]\of{\Aopt,t,s}                
            \\ \overeq{b}\ &2 \sqrt{2} \Ep[p]\of{\Aopt,\topt,\sopt}
            \\ \overeq{c}\ &2 \sqrt{2} \min_{\substack{A \in \R[d\times d] \\ t,s\in \R[d]}}\ \Ep[p]\of{A,t,s}                
            \\ \leq\ &2 \sqrt{2} \min_{\substack{R \in O\of{d} \\ t\in \R[d]}}\ \Ep[p]\of{R,t,-R^T t}                
            \\ \overeq{d}\ &2 \sqrt{2} \min_{\substack{R \in O\of{d} \\ t\in \R[d]}}\ \E\of{R,t}                
            = 2 \sqrt{2} \Eopt,
        \end{split}
    \end{equation}
    where (a) is since $p \geq 2$; (b), (c) are by the definition of $\br{\Aopt,\topt,\sopt}$; and (d) is by the identity \cref{eq:objectives_coincide_rigid}.
    This concludes the proof of \cref{eq:optimality_inequality_rigid} for $p \in \brs{2, \infty}$. 
    
    We shall now prove the improved bound \cref{eq:optimality_inequality_rigid_p2} for $p=2$. The following \lcnamecref{thm:translated_objective} shows that if $\br{\Aopt, \topt, \sopt}$ is a minimizer of $\Ep[2]$, then $\Aopt$ can be completed to a minimizer $\br{\Aopt, \popt, \qopt}$ of $\Fp[2]$.
    \begin{lemma}\label{thm:translated_objective}
        Let $\br{\Aopt, \topt, \sopt}$ be a minimizer of $\Ep[2]\of{A,t,s}$. Then there exist $\popt,\qopt \in \R[d]$ such that
        \begin{equation*}
            \begin{split}                
                \Ep[2]\of{\Aopt, \topt, \sopt} = \Fp[2]\of{\Aopt, \popt, \qopt} = \min_{ \substack{A \in \R[d \times d] \\ \p[0],\q[0] \in \R[d]}}\ \Fp[2]\of{A, \p[0], \q[0]}.
            \end{split}
        \end{equation*}
    \end{lemma}
    \noindent The \lcnamecref{thm:translated_objective} is proven below. 
    We continue the proof of the \lcnamecref{thm:approximation_rigid}. 
    
    Let $\br{\Aopt, \topt, \sopt}$ be a minimizer of $\Ep[2]\br{A,t,s}$, and let $\popt, \qopt$ be as in \cref{thm:translated_objective}. Then
    \begin{equation}\label{proof:approximation_rigid:eq1}
        \begin{split}
            \Eopt = &\min_{\substack{R \in O\br{d} \\ t \in \R[d]}} \E\of{R,t} 
            = \min_{\substack{R \in O\br{d} \\ t \in \R[d]}} \Ep[2]\of{R,t,-R^T t} 
            \geq \min_{\substack{A \in \R[d \times d] \\ t,s \in \R[d]}} \Ep[2]\of{A,t,s}
            \\ = &\Ep[2]\of{\Aopt,\topt,\sopt} 
             = \Fp[2]\of{\Aopt,\popt,\qopt}.
        \end{split}
    \end{equation}
    Since $\Aopt$ is optimal for the translated objective $\Fp[2]\of{\argdot,\popt,\qopt}$,  \cref{thm:projection_orth} implies that
    \begin{equation*}
        \Fp[2]\of{\Rhat, \popt, \qopt} \leq \sqrt{2} \Fp[2]\of{\Aopt, \popt, \qopt}.
    \end{equation*}
    Inserting the above inequality to \cref{proof:approximation_rigid:eq1} yields
    \begin{equation}\label{proof:approximation_rigid:eq2}
        \Eopt \geq \frac{1}{\sqrt{2}} \Fp[2]\of{\Rhat, \popt, \qopt}.
    \end{equation}
    Note that
    \begin{equation}\label{proof:approximation_rigid:eq3}
        \begin{split}
            \Fp[2]\of{\Rhat, \popt, \qopt}
            = &\sum_{i=1}^n \norm{\Rhat \br{\p[i] - \popt} - \br{\q[i]-\qopt} }
            \\= &\sum_{i=1}^n \norm{\Rhat \p[i]-\q[i] - \br{\Rhat \popt-\qopt}}
            \\\geq &\min_{t \in \R[d]}\ \sum_{i=1}^n \norm{\Rhat \p[i]-\q[i] + t}
            \\= &\sum_{i=1}^n \norm{\Rhat \p[i]-\q[i] + \that}
            \\= &\E\of{\Rhat,\that}.
        \end{split}
    \end{equation}
    Combining \cref{proof:approximation_rigid:eq2,proof:approximation_rigid:eq3} yields
    \begin{equation*}
        \E\of{\Rhat,\that} \leq \sqrt{2} \Eopt,
    \end{equation*}
    which concludes the proof of \cref{eq:optimality_inequality_rigid_p2} for $p=2$.

    To complete the proof of the \lcnamecref{thm:approximation_rigid}, we shall now prove \cref{thm:geomed,thm:translated_objective}.
    \begin{proof}[Proof of \cref{thm:geomed}]
    Let $\vopt \in \R[d]$ be a minimizer of 
    \begin{equation}\label{thm:geomed:eq1}
        g \of v = \sum_{j=1}^n \norm{\x[j]-v}.
    \end{equation} 
    If there exist $i \in \brs{n}$ for which $\vopt = \x[i]$, then the claim of the \lcnamecref{thm:geomed} clearly holds. Otherwise, $g\of{v}$ is differentiable at $\vopt$, and since $\vopt$ is a minimizer of $g$, we have
    \begin{equation}\label{thm:geomed:eq2}
        \nabla g\br{\vopt} = -\sum_{j=1}^n \frac{\x[j]-\vopt}{\norm{\x[j]-\vopt}} = \zerovec.
    \end{equation}
    
    Let 
    \begin{equation*}
        i = \argmin_{j \in \brs{n}} \norm{\x[j] - \vopt}.            
    \end{equation*}
    Then for any $j \in \brs{n}$,
    \begin{equation*}
        \begin{split}
            \norm{\x[j]-\x[i]}
            = &\norm{\x[j] - \vopt - \br{ \x[i] - \vopt}}
            \\= &\sqrt{\norm{\x[j] - \vopt}^2 + \norm{\x[i] - \vopt}^2 -2\bra{\x[j] - \vopt,\x[i] - \vopt}} 
            \\\leq &\sqrt{2 \norm{\x[j] - \vopt}^2 - 2\bra{\x[j] - \vopt,\x[i] - \vopt}} 
            \\= &\sqrt{2} \norm{\x[j] - \vopt} \sqrt{1 - \Bigbra{\frac{\x[j] - \vopt}{\norm{\x[j] - \vopt}^2}, \x[i] - \vopt } }.
        \end{split}
    \end{equation*}
    Using Bernoulli's inequality \cref{eq:bernoulli} with $x = -\Bigbra{\frac{\x[j] - \vopt}{\norm{\x[j] - \vopt}^2}, \x[i] - \vopt } $, we have
    \begin{equation*}
        \begin{split}
            \norm{\x[j]-\x[i]} 
            \leq &\sqrt{2} \norm{\x[j] - \vopt} \br{1 - \frac{1}{2}\Bigbra{\frac{\x[j] - \vopt}{\norm{\x[j] - \vopt}^2}, \x[i] - \vopt } }.
            \\= &\sqrt{2} \norm{\x[j] - \vopt} - \frac{1}{\sqrt{2}}\Bigbra{\frac{\x[j] - \vopt}{\norm{\x[j] - \vopt}}, \x[i] - \vopt }.
        \end{split}
    \end{equation*}
    for $j \in \brs{d}$. Therefore,
    \begin{equation*}
        \begin{split}
            \sum_{j=1}^n \norm{\x[j]-\x[i]} 
            \leq &\sqrt{2} \sum_{j=1}^n \norm{\x[j] - \vopt} 
            - \frac{1}{\sqrt{2}} \Biggbra{\br{ \sum_{j=1}^n{ \frac{\x[j] - \vopt}{\norm{\x[j] - \vopt}} }}, \x[i] - \vopt }
            \\ \overeq{a}& \sqrt{2} \sum_{j=1}^n \norm{\x[j] - \vopt} - \frac{1}{\sqrt{2}} \bra{\zerovec, \x[i] - \vopt },
            \\ =& \sqrt{2} \sum_{j=1}^n \norm{\x[j] - \vopt},
        \end{split}
    \end{equation*}
    where (a) is by \cref{thm:geomed:eq2}. Thus, the \lcnamecref{thm:geomed} holds.
\end{proof}

    \begin{proof}[Proof of \cref{thm:translated_objective}]
    Let $\br{\Aopt,\topt,\sopt}$ be a minimizer of $\Ep[2]\of{A,t,s}$. Then
    \begin{equation}\label{pf:translated_objective:eq0}
        \begin{split}
            &\min_{\substack{A \in \R[d \times d]\\ \p[0],\q[0] \in \R[d]}} \ \Fp[2]\of{A,\p[0],\q[0]}
            \\ \overeq{a} &\min_{\substack{A \in \R[d \times d]\\ \p[0],\q[0] \in \R[d]}}\ \sum_{i=1}^n \sqrt{ \frac{ \norm{A \p[i]-\q[i] - \br{A \p[0]-\q[0]}}^2 + \norm{A^T \q[i]-\p[i] - \br{A^T \q[0]-\p[0]}}^2} {2} }
            \\  \overgeq{b} &\min_{\substack{A \in \R[d \times d] \\ t,s \in \R[d]}} \ \sum_{i=1}^n \sqrt{ \frac{ \norm{A \p[i]-\q[i] + t}^2 + \norm{A^T \q[i]-\p[i] + s}^2} {2} }
            \\  \overeq{c} &\min_{\substack{A \in \R[d \times d] \\ t,s \in \R[d]}} \ \Ep[2]\of{A,t,s}
            = \Ep[2]\of{\Aopt,\topt,\sopt},
        \end{split}
    \end{equation}
    where (a), (c) are by the definitions of $\Fp[2]$, $\Ep[2]$ respectively, and (b) can be shown by taking $t = -\br{A \p[0]-\q[0]}$, $s = -\br{A^T \q[0] - \p[0]}$. If there exist $\popt,\qopt \in \R[d]$ such that 
    \begin{equation}\label{pf:translated_objective:eq1}
        \Fp[2]\of{\Aopt,\popt,\qopt} = \Ep[2]\of{\Aopt,\topt,\sopt},
    \end{equation} 
     then by \cref{pf:translated_objective:eq0},
    \begin{equation*}
        \begin{split}
            &\min_{\substack{A \in \R[d \times d]\\ \p[0],\q[0] \in \R[d]}}\ \Fp[2]\of{A,\p[0],\q[0]}
            \leq \Fp[2]\of{\Aopt,\popt,\qopt}
            = \Ep[2]\of{\Aopt,\topt,\sopt}
            \leq \min_{\substack{A \in \R[d \times d]\\ \p[0],\q[0] \in \R[d]}} \ \Fp[2]\of{A,\p[0],\q[0]}
        \end{split}
    \end{equation*}
    and thus $\br{\Aopt,\popt,\qopt}$ is a minimizer of $\Fp[2]$. Therefore, to prove the \lcnamecref{thm:translated_objective}, it is enough to show that there exist $\popt,\qopt \in \R[d]$ that satisfy \cref{pf:translated_objective:eq1}.

    Now, suppose there exist $\popt, \qopt \in \R[d]$ that satisfy the matrix equation
    \begin{equation}\label{pf:translated_objective:eq3}
        {\begin{bmatrix}-\Aopt & \idmat \\ \idmat & -\Aopt^T\end{bmatrix}} 
        \begin{bmatrix}\popt \\ \qopt\end{bmatrix}
        = \begin{bmatrix}\topt \\ \sopt \end{bmatrix},
    \end{equation}
    where $I$ is the $d \times d$ identity matrix. Equation~\cref{pf:translated_objective:eq3} implies that for any $i \in \brs{n}$, 
    \begin{equation}\label{pf:translated_objective:eq4}
        \begin{split}
            \Aopt \p[i] - \q[i] + \topt =\ &\Aopt \br{\p[i]-\popt} - \br{\q[i]-\qopt},
            \\ \Aopt^T \q[i] - \p[i] + \sopt =\ &\Aopt^T \br{\q[i]-\qopt} - \br{\p[i]-\popt}.
        \end{split}
    \end{equation}
    In turn, equation~\cref{pf:translated_objective:eq4} implies that \cref{pf:translated_objective:eq1} holds, as can be verified in the definition of $\Ep[2]$ and $\Fp[2]$. Therefore it is enough to prove that there exist $\popt,\qopt \in \R[d]$ that satisfy \cref{pf:translated_objective:eq3}.
    
   We first prove the existence of such $\popt$,$\qopt$ under two assumptions: (i) $\Aopt$ is diagonal with nonnegative entries, and (ii) $\topt[k]=-\sopt[k]$ for all $k \in \brs{d}$ such that $\Aopt[kk] = 1$. We then release these assumptions in two steps.
    
    First, suppose that assumptions (i) and (ii) hold. \comment[nd]{This might be easier to understand and shorter by stating explicitly that the equation decomposes to seperable equations in $p_k,q_k$, and then showing what the equations look like without stating the solution, just showing it exits} Let $\popt,\qopt \in \R[d]$ be given by 
    \begin{equation}\label{pf:translated_objective:eq5}
        \begin{bmatrix}\popt[k] \\ \qopt[k]\end{bmatrix}
        =
        \begin{cases}
            {\begin{bmatrix}-\Aopt[kk] & 1 \\ 1 & -\Aopt[kk]\end{bmatrix}}^{-1}
            \begin{bmatrix}\topt[k] \\ \sopt[k]\end{bmatrix}
        & \Aopt[kk] \neq 1
        \\ -\frac{1}{2} \begin{bmatrix}\topt[k] \\ \sopt[k]\end{bmatrix}
        & \Aopt[kk]=1,
        \end{cases}
        \qquad k \in \brs{d}.
    \end{equation}
    Note that $\det\of{\brs{\begin{smallmatrix}-\Aopt[kk] & 1 \\ 1 & -\Aopt[kk]\end{smallmatrix}}} = \Aopt[kk]^2 - 1$. Since $\Aopt[kk] \geq 0$ by assumption (i), if $\Aopt[kk] \neq 1$ then the matrix $\brs{\begin{smallmatrix}-\Aopt[kk] & 1 \\ 1 & -\Aopt[kk]\end{smallmatrix}}$ is invertible. Thus, $\popt$ and $\qopt$ of \cref{pf:translated_objective:eq5} are well defined. Note that if $\Aopt[kk] = 1$ then 
    \begin{equation*}
        \begin{split}
        &{\begin{bmatrix}-\Aopt[kk] & 1 \\ 1 & -\Aopt[kk]\end{bmatrix}} 
        \begin{bmatrix}\popt[k] \\ \qopt[k]\end{bmatrix}
        = {\begin{bmatrix}-1 & 1 \\ 1 & -1\end{bmatrix}} 
        \begin{bmatrix}\popt[k] \\ \qopt[k]\end{bmatrix}
        = \begin{bmatrix}\qopt[k] - \popt[k] \\ \popt[k] - \qopt[k]\end{bmatrix}
        \\ = &\tfrac{1}{2} \begin{bmatrix}\topt[k] - \sopt[k] \\ \sopt[k] - \topt[k]\end{bmatrix}
        \overeq{a} \tfrac{1}{2} \begin{bmatrix}2\topt[k] \\ 2 \sopt[k] \end{bmatrix}
        = \begin{bmatrix}\topt[k] \\ \sopt[k] \end{bmatrix},
        \end{split}
    \end{equation*}
    where (a) is by assumption (ii).
    Thus, for any $k \in \brs{d}$,
    \begin{equation}\label{pf:translated_objective:eq6}
        {\begin{bmatrix}-\Aopt[kk] & 1 \\ 1 & -\Aopt[kk]\end{bmatrix}} 
        \begin{bmatrix}\popt[k] \\ \qopt[k]\end{bmatrix}
        = \begin{bmatrix}\topt[k] \\ \sopt[k] \end{bmatrix}.
    \end{equation}
    The combination of \cref{pf:translated_objective:eq6} for all $k \in \brs{d}$, in conjunction with the fact that $\Aopt$ is diagonal, implies that \cref{pf:translated_objective:eq3} holds. Therefore the \lcnamecref{thm:translated_objective} holds under assumptions (i) and (ii).

    Second, to release assumption (ii), suppose that assumption (i) holds. Let us define the vectors $\toptalt, \soptalt \in \R[d]$ by
    \begin{equation}\label{pf:translated_objective:eq7}
        \begin{split}
            \begin{bmatrix} \toptalt[k] \\ \soptalt[k] \end{bmatrix} =
            \begin{cases}
                \begin{bmatrix} \topt[k] \\ \sopt[k] \end{bmatrix} & \Aopt[kk] \neq 1
                \\ \tfrac{1}{2}\begin{bmatrix} \topt[k]-\sopt[k] \\ \sopt[k]-\topt[k] \end{bmatrix} & \Aopt[kk] = 1,
            \end{cases}
            \qquad k \in \brs{d}.
        \end{split}
    \end{equation}
    We shall now show that $\br{\Aopt, \toptalt, \soptalt}$ is a minimizer of $\Ep[2]$ that satisfies both assumptions (i), (ii). Assumption (i) is satisfied since $\Aopt$ is unmodified. Asumption (ii) is satisfied by definition in \cref{pf:translated_objective:eq7}. It is left to show that $\br{\Aopt, \toptalt, \soptalt}$ is indeed a minimizer of $\Ep[2]$. For this it is enough to show that
    \begin{equation}\label{pf:translated_objective:eq7b}
        \Ep[2]\of{\Aopt, \toptalt, \soptalt} \leq \Ep[2]\of{\Aopt, \topt, \sopt}.
    \end{equation}

    It can be shown by a convexity argument that for any three real numbers $a,x,y$,
    \begin{equation}\label{pf:translated_objective:eq8}
        \br{a+x}^2 + \br{a-y}^2 \geq 2 \br{a+\frac{x-y}{2}}^2.
    \end{equation}
    Let $k \in \brs{d}$ such that $\Aopt[kk] = 1$. Then for any $i \in \brs{n}$,
    \begin{equation}\label{pf:translated_objective:eq9}
        \begin{split}
            &\br{\Aopt[kk] \p[i][k] - \q[i][k] + \topt[k] }^2 + \br{\Aopt[kk] \q[i][k] - \p[i][k] + \sopt[k] }^2
            \\ = &\br{\p[i][k] - \q[i][k] + \topt[k] }^2 + \br{\q[i][k] - \p[i][k] + \sopt[k] }^2
            \\ = &\br{\p[i][k] - \q[i][k] + \topt[k] }^2 + \br{\p[i][k] - \q[i][k] - \sopt[k] }^2
            \\ \overgeq{a} &2 \br{\p[i][k] - \q[i][k] + \frac{\topt[k]-\sopt[k]}{2} }^2
             \overeq{b} 2 \br{\p[i][k] - \q[i][k] + \toptalt[k] }^2
            \\ \overeq{c} & \br{\p[i][k] - \q[i][k] + \toptalt[k] }^2 + \br{\p[i][k] - \q[i][k] - \soptalt[k] }^2
            \\ = & \br{\p[i][k] - \q[i][k] + \toptalt[k] }^2 + \br{\q[i][k] - \p[i][k] + \soptalt[k] }^2
            \\ = & \br{\Aopt[kk]\p[i][k] - \q[i][k] + \toptalt[k] }^2 + \br{\Aopt[kk]\q[i][k] - \p[i][k] + \soptalt[k] }^2,
        \end{split}
    \end{equation}
    where (a) follows from \cref{pf:translated_objective:eq8} with $a = \p[i][k] - \q[i][k]$, $x = \topt[k]$ and $y = \sopt[k]$; and (b),(c) follow from the definition of $\toptalt$, $\soptalt$ in \cref{pf:translated_objective:eq7}. Recall that for any $k \in \brs{d}$ such that $\Aopt[kk] \neq 1$, $\toptalt[k]=\topt[k]$ and $\soptalt[k] = \sopt[k]$. This, together with \cref{pf:translated_objective:eq9}, imply that for any $k \in \brs{d}$,
    \begin{equation*}
        \begin{split}
            &\br{\Aopt[kk] \p[i][k] - \q[i][k] + \topt[k] }^2 + \br{\Aopt[kk] \q[i][k] - \p[i][k] + \sopt[k] }^2
            \\ \geq & \br{\Aopt[kk]\p[i][k] - \q[i][k] + \toptalt[k] }^2 + \br{\Aopt[kk]\q[i][k] - \p[i][k] + \soptalt[k] }^2.
        \end{split}
    \end{equation*}
    Therefore, \comment[nd]{suggest delete equations (56)-(57). From the equations you state below it is clear that if we defined $\hat t, \hat s$ so that $\hat t_k=t_k$ if $A_{kk}\neq 1$ and $\hat t_k=-s_k$ otherwise, and defined $\hat s$ accordingly, then  $E_2(A^*,\hat t,\hat s)=E_2(A^*,t^*,s^*) $. By convexity of $E_2$ (or any $E_p$) this equality also holds for the average of $(\hat t,\hat s)$ and $(t^*,s^*)$ which is $(\tilde t,\tilde s) $ }
    \begin{equation*}
        \begin{split}
            \Ep[2]\of{\Aopt,\topt,\sopt}
            =\ &\frac{1}{\sqrt{2}}\sum_{i=1}^n \sqrt{ \norm{\Aopt\p[i]-\q[i]+\topt}^2 + \norm{\Aopt^T\q[i]-\p[i]+\sopt}^2 }
            \\ \overeq{a}\ &\frac{1}{\sqrt{2}}\sum_{i=1}^n \sqrt{ \sum_{k=1}^d \br{\Aopt[kk] \p[i][k] - \q[i][k] + \topt[k] }^2 + \br{\Aopt[kk] \q[i][k] - \p[i][k] + \sopt[k] }^2 }
            \\ \geq\ & \frac{1}{\sqrt{2}} \sum_{i=1}^n \sqrt{ \sum_{k=1}^d \br{\Aopt[kk] \p[i][k] - \q[i][k] + \toptalt[k] }^2 + \br{\Aopt[kk] \q[i][k] - \p[i][k] + \soptalt[k] }^2 }
            \\ =\ &\Ep[2]\of{\Aopt,\toptalt,\soptalt},
        \end{split}
    \end{equation*}
    where (a) is by the diagonal structure of $\Aopt$.
    Hence, \cref{pf:translated_objective:eq7b} holds and thus $\br{\Aopt,\toptalt,\soptalt}$ is a minimizer of $\Ep[2]$. By the previous part of the proof, there exist $\popt,\qopt \in \R[d]$ with $\Fp[2]\of{\Aopt,\popt,\qopt} = \Ep[2]\of{\Aopt,\toptalt,\soptalt}$. Since $\br{\Aopt,\topt,\sopt}$, $\br{\Aopt,\toptalt,\soptalt}$ are both minimizers of $\Fp[2]$,
    \begin{equation*}
        \Fp[2]\of{\Aopt,\popt,\qopt}
        = \Fp[2]\of{\Aopt,\toptalt,\soptalt}
        = \Ep[2]\of{\Aopt,\topt,\sopt}.
    \end{equation*}
    Therefore $\popt,\qopt$ satisfy \cref{pf:translated_objective:eq1}, and thus the \lcnamecref{thm:translated_objective} holds under assumption (i).

    Finally, to release assumption (i), let $\br{\Aopt,\topt,\sopt}$ be a minimizer of $\Ep[2]$. Let $\Aopt = U \Sigma V^T$ be an SVD of $\Aopt$, and let
    \begin{equation*}
        \uu[i] = V^T \p[i],
        \quad
        \vv[i] = U^T \q[i],
        \quad 
        i=1,\ldots,n.
    \end{equation*}
    Define the modified objectives  $\Epalt\of{B,x,y}$ and $\Fpalt\of{B,\uu[0],\vv[0]}$ by
    \begin{equation}\label{pf:translated_objective:eq11}
        \begin{split}
            \Epalt\of{B,x,y}
            = &\sum_{i=1}^n \br{\frac{ \norm{B \uu[i] - \vv[i] + x}^p + \norm{B^T \vv[i] - \uu[i] + y}^p} {2} }^{\frac{1}{p}}
            \\
            \Fpalt\of{B,\uu[0],\vv[0]}
            = &\sum_{i=1}^n \br{\frac{\norm{B \br{\uu[i]-\uu[0]} - \br{\vv[i]-\vv[0]}}^p + \norm{B^T \br{\vv[i]-\vv[0]} - \br{\uu[i]-\uu[0]}}^p} {2}}^{\frac{1}{p}}
        \end{split}
    \end{equation}
    for $B \in \R[d \times d]$ and $x,y,\uu[0],\vv[0] \in \R[d]$.
    By a similar derivation to \cref{pf:projection_orth:eq1} it can be shown that for any matrix $B \in \R[d \times d]$ and vectors $x,y,\uu[0],\vv[0] \in \R[d]$,
    \begin{equation}\label{pf:translated_objective:eq12}
        \begin{split}            
            \Epalt\of{B,x,y} =&\ \Ep\of{U B V^T, U x, V y},
            \\ \Fpalt\of{B,\uu[0],\vv[0]} =&\ \Fp\of{U B V^T, V\uu[0], U\vv[0]}.
        \end{split}
    \end{equation}
    Let
    \begin{equation*}
        \xopt = U^T \topt,
        \quad
        \yopt = V^T \sopt.
    \end{equation*}
    Since $\br{\Aopt,\topt,\sopt}$ is a minimizer of $\Ep[2]$, then by \cref{pf:translated_objective:eq12}, $\br{\Sigma, \xopt, \yopt}$ is a minimizer of $\Epalt[2]$.
    Since $\Sigma$ is diagonal with nonnegative entries, then $\br{\Sigma, \xopt, \yopt}$ satisfies assumption (i). Therefore, by the previous part of the proof, there exist $\uopt, \vopt \in \R[d]$ such that
    \begin{equation}\label{pf:translated_objective:eq13}
        \Fpalt[2]\of{\Sigma,\uopt,\vopt} = \Epalt[2]\of{\Aopt, \xopt, \yopt}.
    \end{equation}
    Let
    \begin{equation*}
        \popt = V \uopt,
        \quad
        \qopt = U \vopt.
    \end{equation*}
    Then 
    \begin{equation*}
        \begin{split}
            \Fp[2]\of{\Aopt,\popt,\qopt}
            =&\ \Fp[2]\of{\Aopt, V \uopt, U \vopt}
            \\ \overeq{a}&\ \Fpalt[2]\of{U^T \Aopt V, V^T V \uopt, U^T U \vopt}
            \\ =&\ \Fpalt[2]\of{\Sigma, \uopt, \vopt}
            \\ \overeq{b}&\ \Epalt[2]\of{\Sigma, \xopt, \yopt}
            \\ =&\ \Epalt[2]\of{\Sigma, U^T \topt,  V^T \sopt}
            \\ \overeq{c}&\ \Ep[2]\of{U \Sigma V^T, U U^T \topt,  V V^T \sopt}
            \\ =&\ \Ep[2]\of{\Aopt, \topt,  \sopt},
        \end{split}
    \end{equation*}
    where (a),(c) are by \cref{pf:translated_objective:eq12} and (b) is by \cref{pf:translated_objective:eq13}. Therefore $\popt,\qopt$ satisfy \cref{pf:translated_objective:eq1}, and thus the \lcnamecref{thm:translated_objective} is proven.
\end{proof}
    This concludes the proof of \cref{thm:approximation_rigid}.
\end{proof}

\RestateFactorsAreOptimal*
\begin{proof}[Proof of \cref{thm:factors_are_optimal}]
Set $n=1$ and let $\p[1] \in \R[d]$ be an arbitrary nonzero vector. Let $A = \zeromat$ be the zero matrix. For part 1, set $\q[1] = -\Aproj\p[1]$. 
Then for any $p \in \brs{1,\infty}$, 
\begin{equation*}
    \begin{split}
        \Ep\br{A} 
        = \Ep\br{\zeromat} 
        =& \br{ \frac{\norm{\q[1]}^p + \norm{\p[1]}^p}{2} }^{\frac{1}{p}} 
        = \norm{\p[1]}. 
    \end{split}
\end{equation*}
However,
\begin{equation*}
    \begin{split}
        \Ep\br{\Aproj} 
        =& \br{ \frac{\norm{\Aproj\p[1]-\q[1]}^p + \norm{\Aproj^T\q[1]-\p[1]}^p}{2} }^{\frac{1}{p}} 
        \\ =& \br{ \frac{\norm{2\Aproj\p[1]}^p + \norm{-2\p[1]}^p}{2} }^{\frac{1}{p}} 
        = 2\norm{\p[1]}
        = 2 \Ep \of A.
    \end{split}
\end{equation*}
For part 2, set $\q[1] = \zerovec$. Then
\begin{equation*}
    \begin{split}
        \Ep\br{A} 
        =& \br{ \frac{\norm{A \p[1]}^p + \norm{\p[1]}^p}{2} }^{\frac{1}{p}} 
        \geq \frac{1}{2^{\frac{1}{p}}}\norm{\p[1]}
        = \Ep\br{\zeromat}.
    \end{split}
\end{equation*}
Therefore, $\zeromat$ is a global minimizer of $\Ep$, with an objective value of $\frac{1}{2^{\frac{1}{p}}}\norm{\p[1]}$. However, for any orthogonal matrix $R$,
\begin{equation*}
    \begin{split}
        \Ep\br{R} 
        =& \br{ \frac{\norm{R \p[1]}^p + \norm{\p[1]}^p}{2} }^{\frac{1}{p}} 
        = \norm{\p[1]}.
    \end{split}
\end{equation*}
Thus, for $A = \zeromat$,
\begin{equation*}
    \Ep\br{\Aproj} = \norm{\p[1]} = 2^{\frac{1}{p}} \Ep\br{A}.
\end{equation*}
\end{proof}

\RestateFactorsAreConvexOptimal*
\begin{proof}[Proof of \cref{thm:factors_are_convex_optimal}]
    Let $u,v \in \R[d]$ be two vectors such that $u \perp v$ and $\norm{u} = \norm{v} = 1$. Let $R_0 \in O\of{d}$ such that $R_0 u = v$. Let $\Rhat = \orthproj{\zeromat}$. Define $P,Q \in \R[d \times n]$ for $n=2$ by 
    \begin{equation*}
        \begin{split}
            \p[1]=\ &\p[2] = \Rhat^T u,
            \\ \q[1]=\ &v,\ \q[2] = -v.
        \end{split}
    \end{equation*}
    Let $\Pi \in \R[2 \times 2]$ be the permutation matrix
    \begin{equation*}
        \Pi = 
        \begin{bmatrix} 0 & 1 \\ 1 & 0 \end{bmatrix}
    \end{equation*}
    and note that $Q \Pi = -Q$ and $P \Pi = P$. Let $A \in \R[d \times d]$ be an arbitrary matrix. Then
    \begin{equation}\label{pr:factors_are_convex_optimal:eq1}
        \begin{split}
            \G\brmid{-A}{P,Q} 
            =\ &\G\brmid{-A}{P \Pi,-Q \Pi} 
            \\ \overeq{a}\ &\G\brmid{A}{P \Pi, Q \Pi} 
            \overeq{b} \G\brmid{A}{P, Q},
        \end{split}
    \end{equation}
    where (a), (b) are by Assumptions 3 and 2 respectively. Combining \cref{pr:factors_are_convex_optimal:eq1} with the assumption that $\G\brmid{\argdot}{P,Q}$ is convex, we have
    \begin{equation}\label{pr:factors_are_convex_optimal:eq2}
        \begin{split}
            \G\brmid{\zeromat}{P,Q}
            =\ \G\brmid{\frac{A + \br{-A}}{2}}{P,Q}
            \leq\ &\frac{1}{2}\G\brmid{A}{P,Q} + \frac{1}{2}\G\brmid{-A}{P,Q}
            \\=\ &\frac{1}{2}\G\brmid{A}{P,Q} + \frac{1}{2}\G\brmid{A}{P,Q}
            = \G\brmid{A}{P,Q}.
        \end{split}
    \end{equation}
    Since $A$ is arbitrary, \cref{pr:factors_are_convex_optimal:eq2} implies that $\Aopt = \zeromat$ is a minimizer of $\G\brmid{\argdot}{P,Q}$. Therefore, 
    \begin{equation*}
        \Rhat = \orthproj{\zeromat} = \Aoptproj.
    \end{equation*}
    Since $u$ and $v$ are perpendicular unit vectors,
    \begin{equation*}
        \norm{u - v} = \norm{u + v} = \sqrt{\norm{u}^2 + \norm{v}^2} = \sqrt{2}
    \end{equation*}
    and thus
    \begin{equation*}
        \begin{split}
            \E\of{\Rhat} 
            =\ &\norm{\Rhat \p[1] - \q[1]} + \norm{\Rhat \p[2] - \q[2]}
            \\=\ &\norm{\Rhat \br{\Rhat^T u} - v} + \norm{\Rhat \br{\Rhat^T u} + v}
            \\=\ &\norm{u - v} + \norm{u + v}
             =  2 \sqrt{2}.
            \end{split}
    \end{equation*}
    However,
    \begin{equation*}
        \begin{split}
            \E\of{R_0 \Rhat} 
            =\ &\norm{R_0 \Rhat \p[1]-\q[1]} + \norm{R_0 \Rhat \p[2]-\q[2]} 
            \\=\ &\norm{R_0 \Rhat \br{\Rhat^T u} - v} + \norm{R_0 \Rhat \br{\Rhat^T u} + v} 
            \\=\ &\norm{R_0 u - v} + \norm{R_0 u + v} 
            \\ =\ &\norm{v - v} + \norm{v + v} 
             = 2 \norm{v} = 2.
            \end{split}
    \end{equation*}
    Also note that for any $R \in O\of{d}$,
    \begin{equation*}
        \begin{split}
            \E\of{R} 
            =\ &\norm{R \br{\Rhat^T u} - v} + \norm{R \br{\Rhat^T u} + v}
            \\\overgeq{a}\ &\sqrt{ \norm{R \br{\Rhat^T u} - v}^2 + \norm{R \br{\Rhat^T u} + v}^2 }
            \\\overeq{b}\ &\sqrt{2\norm{R \br{\Rhat^T u}}^2 + 2\norm{v}^2} = 2.
            \end{split}
    \end{equation*}
    where (a) is by the $\ell_1-\ell_2$ norm inequality and (b) is by the parallelogram law.
    Therefore,
    \begin{equation*}
        \E\of{\Rhat} = \sqrt{2} \E\of{R_0 \Rhat} = \sqrt{2} \min_{R \in O\of{d}}\ \E\of{R} > 0.
    \end{equation*}
\end{proof}
\subsection{Recovery guarantees}\label{subapp_proofs_recovery}
\RestateRecoveryOrth*
\begin{proof}[Proof of \cref{thm:recovery_orth}]
    Define the {\em advantage function} $\phi : \R[d \times d] \rightarrow \R$,
    \begin{equation}\label{eq:def_phi_orth}
        \phi \br{A} = \frac{1}{2}\sum_{i\in\I} \Bigbr{ \norm{A \p[i]} + \norm{A^T \q[i]} }  
        -  \frac{1}{2}\sum_{i\in\I^C}  \Bigbr{ \norm{A \p[i]} + \norm{A^T \q[i]} }. 
    \end{equation}    
    Using $\phi\br{A}$, the following \lcnamecref{thm:advantage_inequality_orth} provides a lower bound on the advantage of the true $R_0$ as a solution of \cref{pr:relax_orth} over other potential solutions. A proof appears below.
    \begin{lemma}\label{thm:advantage_inequality_orth}
        For any $A \in \R[d \times d]$ and any $p \in \brs{1,\infty}$,
        \begin{equation}\label{eq:advantage_orth}
            \Ep\br{R_0+A} - \Ep\br{R_0}
            \geq \phi \br{A}.
        \end{equation}
    \end{lemma}
    It follows from \cref{thm:advantage_inequality_orth} that if $\phi\br{A} > 0$ for all $A \neq 0$, then $R_0$ is the unique global minimizer of $\Ep$. The following \lcnamecref{thm:DIP_implies_advantage_orth}, proven below, shows that if $P$,$Q$ satisfy the linear DIP, then $\phi\br{A}$ is indeed positive for any $A \neq 0$.
    \begin{lemma}\label{thm:DIP_implies_advantage_orth}
        Suppose that $P$,$Q$ satisfy the linear DIP. Then for any matrix $A \neq 0$, $\phi\br{A} > 0$.
    \end{lemma}
    We shall now prove \cref{thm:advantage_inequality_orth,thm:DIP_implies_advantage_orth}.
    \begin{proof}[Proof of \cref{thm:advantage_inequality_orth}]
        Let $A \in \R[d \times d]$. Since $\Ep\br{A} \geq \Ep[1]\br{A}$ and $\Ep\br{R_0} = E\br{R_0} = \Ep[1]\br{R_0}$, we have
        \begin{equation*}
            \begin{split}
                &\Ep\br{R_0 + A} - \Ep\br{R_0} =  \Ep\br{R_0 + A} - \E\br{R_0} 
                \\ \geq &\Ep[1]\br{R_0 + A} - \E\br{R_0}
                = \Ep[1]\br{R_0 + A} - \Ep[1]\br{R_0}.
            \end{split}
        \end{equation*}
        Therefore it is enough to prove \cref{eq:advantage_orth} for $p=1$. Expanding the left-hand side of \cref{eq:advantage_orth}, we get
        \begin{equation}\label{pf:advantage_inequality_orth:eq1}
            \begin{split}
            2 \br{ \Ep[1]\br{R_0+A} - \Ep[1]\br{R_0} }
            =& \sum_{i=1}^n { \norm{\br{R_0+A}\p[i] - \q[i]} + \norm{\br{R_0+A}^T \q[i] - \p[i]} }
            \\-& \sum_{i=1}^n { \norm{R_0 \p[i] - \q[i]} + \norm{R_0^T \q[i] - \p[i]} }.
            \end{split}
        \end{equation}
        Let us split the sums in \cref{pf:advantage_inequality_orth:eq1} to inlier and outlier terms. First consider the inliers. Since $\q[i] = R_0 \p[i]$ for $i \in \I$, we have
        \begin{equation}\label{pf:advantage_inequality_orth:eq2}
            \begin{split}
            &\sum_{i\in\I} { \norm{\br{R_0+A}\p[i] - \q[i]} + \norm{\br{R_0+A}^T \q[i] - \p[i]} }
            - \sum_{i\in\I} { \norm{R_0 \p[i] - \q[i]} + \norm{R_0^T \q[i] - \p[i]} }
            \\ = &\sum_{i\in\I} { \norm{\br{R_0+A}\p[i] - \q[i]} + \norm{\br{R_0+A}^T \q[i] - \p[i]} }
            \\ = &\sum_{i\in\I} { \norm{A \p[i]} + \norm{A^T \q[i]} }.
            \end{split}
        \end{equation}
        Second, consider the outliers. By the triangle inequality, for each $i$,
        \begin{equation*}
            \begin{split}
            { \norm{\br{R_0+A}\p[i] - \q[i]} + \norm{\br{R_0+A}^T \q[i] - \p[i]} }
            + { \norm{A \p[i]} + \norm{A^T \q[i]} }
            \geq { \norm{R_0 \p[i] - \q[i]} + \norm{R_0^T \q[i] - \p[i]} }.
            \end{split}
        \end{equation*}
        Therefore,
        \begin{equation}\label{pf:advantage_inequality_orth:eq3}
            \begin{split}
            &\sum_{i\in\I^C} { \norm{\br{R_0+A}\p[i] - \q[i]} + \norm{\br{R_0+A}^T \q[i] - \p[i]} }
            - \sum_{i\in\I^C} { \norm{R_0 \p[i] - \q[i]} + \norm{R_0^T \q[i] - \p[i]} }
            \\ &\geq - \sum_{i\in\I^C} { \norm{A \p[i]} + \norm{A^T \q[i]} }.
            \end{split}
        \end{equation}
        Inserting inequalities~\cref{pf:advantage_inequality_orth:eq2,pf:advantage_inequality_orth:eq3} into \cref{pf:advantage_inequality_orth:eq1} yields
        \begin{equation*}
            \begin{split}
            2 \br{ \Ep[1]\br{R_0+A} - \Ep[1]\br{R_0} }
            \geq& \sum_{i\in\I} { \norm{A \p[i]} + \norm{A^T \q[i]} } - \sum_{i\in\I^C} { \norm{A \p[i]} + \norm{A^T \q[i]} }
            \\ =& 2 \phi\br{A},
            \end{split}
        \end{equation*}
        which proves the \lcnamecref{thm:advantage_inequality_orth}.
    \end{proof}
    Let us now prove \cref{thm:DIP_implies_advantage_orth}.
    \begin{proof}[Proof of \cref{thm:DIP_implies_advantage_orth}]
        We first make the following \lcnamecref{thm:integration}.
        \begin{claim}\label{thm:integration}
            For any $x \in \R[d]$,
            \begin{equation}\label{eq:integration}
                \int_{S^{d-1}} \abs{\bra{v,x}}dv = C \norm{x},
            \end{equation}
            where $S^{d-1}$ is the unit sphere in $\R[d]$, and $C$ is a positive constant that depends only on the dimension $d$.
        \end{claim}
        \cref{thm:integration} essentially states that the expected magnitude of a random projection of $x$ on a line that goes through the origin is proportional to $\norm{x}$. This \lcnamecref{thm:integration} can be easily proven by symmetry considerations. 

        Now suppose that $P$, $Q$ satisfy the linear DIP. Let $A \in \R[d]$ such that $A \neq 0$. Since the nullspace of $A^T$ is at most $\br{d-1}$-dimensional, it is of measure zero in $\R[d]$. It follows that for almost any $v \in S^{d-1}$ we have that $A^Tv \neq 0$. For such $v$ we can invoke the linear DIP \cref{eq:linear_DIP} with $u=A^Tv$ and get
        \begin{equation*}
            \sum_{i \in \I} \abs{\bra{A^Tv,\p[i]}} > \sum_{i \in \I^C} \abs{\bra{A^Tv,\p[i]}},
        \end{equation*} 
        or equivalently 
        \begin{equation*}
            \sum_{i \in \I} \abs{\bra{v,A \p[i]}} - \sum_{i \in \I^C} \abs{\bra{v,A \p[i]}} > 0.
        \end{equation*} 
        By the above discussion, the left-hand side of the above inequality is positive for almost any $v \in S^{d-1}$. Therefore its integral is also positive:
        \begin{equation*}
            \int_{S^{d-1}} \br{ \sum_{i \in \I} \abs{\bra{v,A \p[i]}} - \sum_{i \in \I^C} \abs{\bra{v,A \p[i]}} } dv > 0.
        \end{equation*} 
        Applying \cref{thm:integration} to the above yields
        \begin{equation*}
            \sum_{i \in \I} \norm{A \p[i]} - \sum_{i \in \I^C} \norm{A \p[i]} > 0.
        \end{equation*}
        Using a similar argument on the right-hand side of \cref{eq:linear_DIP} with $u = Av$, it can be shown that the linear DIP implis that
        \begin{equation*}
            \sum_{i \in \I} \norm{A^T \q[i]} - \sum_{i \in \I^C} \norm{A^T \q[i]} > 0.
        \end{equation*}
        The two above inequalities combined imply the claim of the \lcnamecref{thm:DIP_implies_advantage_orth}. 
    \end{proof}

    This concludes the proof of \cref{thm:recovery_orth}.
\end{proof}

\RestateRecoveryRigid*
\begin{proof}[Proof of \cref{thm:recovery_rigid}]
    The proof is similar to that of \cref{thm:recovery_orth}. We first redefine the advantage function $\phi$ of \cref{eq:def_phi_orth} to accomodate translations:
    \begin{equation}\label{eq:def_phi_rigid}
        \phi \br{A,t,s} = \frac{1}{2}\sum_{i\in\I} \Bigbr{ \norm{A \p[i] + t} + \norm{A^T \q[i] + s} }  
        -  \frac{1}{2}\sum_{i\in\I^C}  \Bigbr{ \norm{A \p[i] + t} + \norm{A^T \q[i] + s} }. 
    \end{equation}
    Using $\phi\br{A,t,s}$, the following \lcnamecref{thm:advantage_inequality_rigid} lower-bounds the advantage of $\br{R_0,t_0,-R_0^T t_0}$ over other solutions of \cref{pr:relax_rigid}.
    \begin{lemma}\label{thm:advantage_inequality_rigid}
        For any $p \in \brs{1,\infty}$, any $A \in \R[d \times d]$ and any $t,s \in \R[d]$,
        \begin{equation}\label{eq:advantage_rigid}
            \Ep\br{R_0+A, t_0+t, -R_0^T t_0 + s} - \Ep\br{R_0, t_0, -R_0^T t_0}
            \geq \phi \br{A,t,s}.
        \end{equation}
    \end{lemma}
    The following \lcnamecref{thm:DIP_implies_advantage_rigid} shows that if $P$,$Q$ satisfy the affine DIP, then $\phi\br{A,t,s}$ is positive for any $A,t,s$ that are not all equal to zero. 
    \begin{lemma}\label{thm:DIP_implies_advantage_rigid}
        Suppose that $P$,$Q$ satisfy the affine DIP. Then for any $A$,$t$,$s$ such that $\norm{A}+\norm{t}+\norm{s} > 0$, $\phi\br{A,t,s} > 0$.
    \end{lemma}
    From \cref{thm:advantage_inequality_rigid,thm:DIP_implies_advantage_rigid} it follows that
    $\br{R_0, t_0, -R_0^T t_0}$ is the unique global minimizer of $\Ep\br{R,t,s}$. To complete the proof of the \lcnamecref{thm:recovery_rigid}, let us now prove these lemmas.
        \begin{proof}[Proof of \cref{thm:advantage_inequality_rigid}]
            Let $A \in \R[d \times d]$ and $t,s \in \R[d]$. Since $\Ep$ is monotone-increasing with respect to $p$, and $\Ep\br{R_0,t_0,-R_0^T t_0} = \E\br{R_0,t_0}$ for all $p \in \brs{1,\infty}$, it is enough to prove \cref{eq:advantage_rigid} for $p=1$. Expanding the left-hand side of \cref{eq:advantage_rigid}, we get
            \begin{equation}\label{pf:advantage_inequality_rigid:eq1}
                \begin{split}
                &2 \br{ \Ep[1]\br{R_0+A, t_0+t, -R_0^T t_0 + s} - \Ep[1]\br{R_0, t_0, -R_0^T t_0} }
                \\=& \sum_{i=1}^n { \norm{\br{R_0+A}\p[i] - \q[i] + t_0 + t} + \norm{\br{R_0+A}^T \q[i] - \p[i] -R_0^T t_0 + s} }
                \\-& \sum_{i=1}^n { \norm{R_0 \p[i] - \q[i] + t_0} + \norm{R_0^T \q[i] - \p[i] - R_0^T t_0} }.
                \end{split}
            \end{equation}
            Let us split \cref{pf:advantage_inequality_rigid:eq1} to inlier and outlier terms. For $i \in \I$, $\q[i] = R_0 \p[i] + t_0$, and thus
            \begin{equation}\label{pf:advantage_inequality_rigid:eq2}
                \begin{split}
                \\& \sum_{i \in \I} { \norm{\br{R_0+A}\p[i] - \q[i] + t_0 + t} + \norm{\br{R_0+A}^T \q[i] - \p[i] -R_0^T t_0 + s} }
                \\-& \sum_{i \in \I} { \norm{R_0 \p[i] - \q[i] + t_0} + \norm{R_0^T \br{\q[i] - t_0} - \p[i]} }.
                \\=& \sum_{i \in \I} { \norm{\br{R_0+A}\p[i] - \q[i] + t_0 + t} + \norm{\br{R_0+A}^T \q[i] - \p[i] -R_0^T t_0 + s} }
                \\ = &\sum_{i\in\I} { \norm{A \p[i] + t} + \norm{A^T \q[i] + s} }.
                \end{split}
            \end{equation}
            Second, consider the outliers. By the triangle inequality, for each $i \in \brs{n}$,
            \begin{equation*}
                \begin{split}
                    &{ \norm{\br{R_0+A}\p[i] - \q[i] + t_0 + t} + \norm{\br{R_0+A}^T \q[i] - \p[i] -R_0^T t_0 + s} }
                + { \norm{A \p[i] + t} + \norm{A^T \q[i] + s} }
                \\ &\geq { \norm{R_0\p[i] - \q[i] + t_0 } + \norm{R_0^T \q[i] - \p[i] -R_0^T t_0} }.
                \end{split}
            \end{equation*}
            Therefore,
            \begin{equation}\label{pf:advantage_inequality_rigid:eq3}
                \begin{split}
                &\sum_{i\in\I^C} { \norm{\br{R_0+A}\p[i] - \q[i] + t_0 + t} + \norm{\br{R_0+A}^T \q[i] - \p[i] -R_0^T t_0 + s} }
                \\-&\sum_{i\in\I^C} { \norm{R_0\p[i] - \q[i] + t_0 } + \norm{R_0^T \q[i] - \p[i] -R_0^T t_0} }
                \\ \geq &- \sum_{i\in\I^C} { \norm{A \p[i] + t} + \norm{A^T \q[i] + s} }.
                \end{split}
            \end{equation}
            Inserting inequalities~\cref{pf:advantage_inequality_rigid:eq2,pf:advantage_inequality_rigid:eq3} into \cref{pf:advantage_inequality_rigid:eq1} yields
            \begin{equation*}
                \begin{split}
                &2 \br{ \Ep[1]\br{R_0+A, t_0+t, -R_0^T t_0 + s} - \Ep[1]\br{R_0, t_0, -R_0^T t_0} }
                \\\geq& \sum_{i\in\I} { \norm{A \p[i] + t } + \norm{A^T \q[i] + s} } - \sum_{i\in\I^C} { \norm{A \p[i] + t } + \norm{A^T \q[i] + s} }
                 = 2 \phi\br{A,t,s},
                \end{split}
            \end{equation*}
            which proves the \lcnamecref{thm:advantage_inequality_rigid}.
        \end{proof}
        Let us now prove \cref{thm:DIP_implies_advantage_rigid}.
        \begin{proof}[Proof of \cref{thm:DIP_implies_advantage_rigid}]
            Let $A \in \R[d \times d]$ and $t \in \R[d]$ that are not both equal to zero. Let $v \in \R[d]$. Denote
            \begin{equation*}
                u\br{v} = A^Tv, \qquad \alpha\br{v} = \bra{v, t}.
            \end{equation*}
            For any $p \in \R[d]$,
            \begin{equation*}
                \bra{v, Ap+t} = \bra{u\br{v}, p} + \alpha\br{v}. 
            \end{equation*}
            Therefore,
            \begin{equation}\label{pf:DIP_implies_advantage_rigid:eq1}
                \begin{split}
                &\sum_{i \in \I} \abs{ \bra{v, A\p[i]+t} } = \sum_{i \in \I} \abs{ \bra{u\br{v}, \p[i]} + \alpha\br{v} },
                \\ &\sum_{i \in \I^C} \abs{ \bra{v, A\p[i]+t} } = \sum_{i \in \I^C} \abs{ \bra{u\br{v}, \p[i]} + \alpha\br{v} }.
                \end{split}
            \end{equation}
            If $\norm{u\br{v}} + \abs{\alpha\br{v}} > 0$, then by the affine DIP \cref{eq:affine_DIP,pf:DIP_implies_advantage_rigid:eq1},
            \begin{equation}\label{pf:DIP_implies_advantage_rigid:eq2}
                \sum_{i \in \I} \abs{ \bra{v, A\p[i]+t} } > \sum_{i \in \I^C} \abs{ \bra{v, A\p[i]+t} }.
            \end{equation}
            Since $A$ and $t$ are not both equal to zero, the set $\brc{v \in S^{d-1}\ :\ u\br{v}=0 \wedge \alpha\br{v}=0}$ is of measure zero in the unit sphere $S^{d-1}$. Therefore, the inequality \cref{pf:DIP_implies_advantage_rigid:eq2} is satisfied for almost any $v \in S^{d-1}$. This implies that 
            \begin{equation*}
                \int_{S^{d-1}}
                \sum_{i \in \I} \abs{ \bra{v, A\p[i]+t} } dv > \int_{S^{d-1}} \sum_{i \in \I^C} \abs{ \bra{v, A\p[i]+t} } dv.
            \end{equation*}
            Applying \cref{thm:integration} to the above inequality yields
            \begin{equation}\label{pf:DIP_implies_advantage_rigid:eq3}
                \sum_{i \in \I} \norm{ A\p[i]+t } > \sum_{i \in \I^C} \norm{ A\p[i]+t }.
            \end{equation}
            By a similar argument it can be shown that for any $A \in \R[d\times d]$, $s \in \R[d]$ that are not both equal zero, the affine DIP implies that
            \begin{equation}\label{pf:DIP_implies_advantage_rigid:eq4}
                \sum_{i \in \I} \norm{ A^T \q[i]+s } > \sum_{i \in \I^C} \norm{ A^T \q[i]+s }.
            \end{equation}
    
            To finalize the proof, let $A \in \R[d \times d]$ and $t,s \in \R[d]$ such that not all three equal zero. We need to show that $\phi\br{A,t,s} > 0$. If $A \neq 0$, or if $A=0$ and $t,s \neq 0$, then we can use \cref{pf:DIP_implies_advantage_rigid:eq3,pf:DIP_implies_advantage_rigid:eq4} and get
            \begin{equation*}
                \sum_{i \in \I} \norm{ A\p[i]+t } + \norm{ A^T\q[i]+s } > 
                \sum_{i \in \I^C} \norm{ A\p[i]+t } + \norm{ A^T\q[i]+s }.
            \end{equation*}
            If $A=0$ and exactly one of $t$,$s$ equals zero, suppose W.L.O.G that $s=0$. Then $t \neq 0$, and by \cref{pf:DIP_implies_advantage_rigid:eq3},
            \begin{equation*}
                \sum_{i \in \I} \norm{ t } 
                = \sum_{i \in \I} \norm{ A\p[i]+t }
                > \sum_{i \in \I^C} \norm{ A\p[i]+t }
                = \sum_{i \in \I^C} \norm{ t }.
            \end{equation*}
            Therefore,
            \begin{equation*}
                \sum_{i \in \I} \norm{ A\p[i]+t } + \norm{ A^T\q[i]+s }
                = \sum_{i \in \I} \norm{ t }
                > \sum_{i \in \I^C} \norm{ t }
                = \sum_{i \in \I^C} \norm{ A\p[i]+t } + \norm{ A^T\q[i]+s }.
            \end{equation*}
            The case $A,t=0$, $s \neq 0$ is handled similarly.
            Thus the \lcnamecref{thm:DIP_implies_advantage_rigid} is proven.
        \end{proof}
        This concludes the proof of \cref{thm:recovery_rigid}.
    \end{proof}

\RestateDIPisOptimal*
\begin{proof}[Proof of \cref{thm:DIP_is_optimal}]
    Let $d \geq 1$ and $\varepsilon \in \br{0,1}$. Let $\brc{e_i}_{i=1}^d$ be the standard unit vectors in $\R[d]$. Let $R_0 = I_d$ be the $d \times d$ unit matrix, and let $R_1 \in \R[d \times d]$ be defined by
    \begin{equation*}
        R_1 e_i = e_{i+1},\ i=1, \ldots, d-1,\quad R_1 e_d = \sigma e_1, 
    \end{equation*}
    where $\sigma \in \brc{1,-1}$ is chosen such that $\det\br{R_1} = 1$. It is easy to check that $\norm{R_0-R_1}^2_F = 2d$.

    Define the points $P = \br{\p[i]}_{i=1}^n$, $Q = \br{\q[i]}_{i=1}^n$ for $n=2d$ by
    \begin{equation}\label{pf:DIP_is_optimal_eq1}
        \begin{split}
            &\p[i] = \br{1-\varepsilon} e_i, \quad \q[i] = R_0 \p[i], \quad i=1,\ldots,d,
            \\&\p[i] = e_{i-d}, \quad \q[i] = R_1 \p[i], \quad i=d+1,\ldots,2d.
        \end{split}
    \end{equation}
    Set $\I = \brc{1,\ldots,d}$, then $P$, $Q$ satisfy condition no. 1 of the \lcnamecref{thm:DIP_is_optimal}. It follows from \cref{pf:DIP_is_optimal_eq1} that $R_0$ and $R_1$ act as permutations on the points $P$, up to a possibe multiplication by $-1$ (in the case $\q[2d] = \sigma e_1$). Hence, the following equalities of unordered sets hold:
    \begin{equation}\label{pf:DIP_is_optimal_eq3}
        \begin{split}
            &\brc{\p[i]}_{i \in \I} = \brc{\q[i]}_{i \in \I} = \brc{\br{1-\varepsilon}e_i}_{i=1,\ldots,d}, \quad
            \\&\brc{\p[i]}_{i \in \I^C} = \brc{\sigma_i \q[i]}_{i \in \I^C} = \brc{e_i}_{i=1,\ldots,d},
        \end{split}        
    \end{equation}
    where $\sigma_{i}=\sigma$ for $i \neq 2d$ and $\sigma_{2d} = -1$.

    Let $u \in \R[d]$ be an arbitrary vector. Then by \cref{pf:DIP_is_optimal_eq3},
    \begin{equation}\label{pf:DIP_is_optimal_eq2}
        \begin{split}
            &\sum_{i \in \I} \abs{\bra{u,\p[i]}} = \br{1-\varepsilon} \sum_{i \in \I^C} \abs{\bra{u,\p[i]}},
            \\&\sum_{i \in \I} \abs{\bra{u,\q[i]}} = \br{1-\varepsilon} \sum_{i \in \I^C} \abs{\bra{u,\q[i]}}.
        \end{split}
    \end{equation}
    and thus $P$ and $Q$ satisfy condition no. 2 of the \lcnamecref{thm:DIP_is_optimal}. 
    However, \cref{pf:DIP_is_optimal_eq2} also implies that $P$, $Q$ satisfy the linear DIP with respect to $\br{R_1,\I^C}$. Therefore, by \cref{thm:recovery_orth}, the unique global minimizer of $\Ep$ is $R_1$ for any $p \in \brs{1,\infty}$. 
\end{proof}

\listoftodos

\end{document}